\documentclass{article}
\usepackage[main,preprint]{public}
\usepackage{times}
\usepackage{amsfonts}
\usepackage{flafter}
\usepackage{float}
\usepackage{placeins}
\usepackage{url}
\usepackage{xspace}
\usepackage{amsmath,amsthm}
\usepackage{booktabs}
\usepackage{longtable}
\usepackage{graphicx}
\usepackage[dvipsnames]{xcolor}
\usepackage{amssymb}
\usepackage{array,tabularx}
\usepackage[shortlabels,inline]{enumitem}
\usepackage{caption}
\ExplSyntaxOn
\NewDocumentCommand{\wrapname}{m}{\group_begin:\cs_set_eq:NN\xspace\prg_do_nothing:\tl_set:Nx\l_tmpa_tl{#1}\tl_replace_all:Nnn\l_tmpa_tl{~}{\space}\tl_map_inline:Nn\l_tmpa_tl{##1\allowbreak}\group_end:}
\ExplSyntaxOff
\usepackage[strings]{underscore}
\setcitestyle{numbers,square,comma}
\newcommand{\iv}[1]{\textsubscript{\fontsize{7}{8}\selectfont #1}}
\newcommand{\ckpt}[1]{\texttt{\small #1}}
\newcommand{\refusedmark}{\textsuperscript{\fontsize{7}{8}\selectfont\dag}}
\newcommand{\aftertest}{\textsuperscript{\ddag}\xspace}
\xspaceaddexceptions{\aftertest}
\newcommand{\INnine}{ImageNet-9\xspace}
\newcommand{\INonek}{ImageNet 1k\xspace}
\newcommand{\CCBYtwo}{Creative Commons Attribution 2.0\xspace}
\newcommand{\CCBYfour}{Creative Commons Attribution 4.0\xspace}
\newcommand{\Apachetwo}{Apache 2.0\xspace}
\newcommand{\WordNetRel}{WordNet~3.0~\cite{miller1995wordnet}\xspace}

\usepackage{colortbl}
\definecolor{MethodHue}{HTML}{D55E00}
\colorlet{MethodTint}{MethodHue!8}
\newcommand{\BestA}[1]{\ifmmode\mathbf{#1}\else\textbf{#1}\fi}
\newcommand{\BestB}[1]{\ifmmode\mathbf{#1}\else\textbf{#1}\fi}
\newcommand{\BestC}[1]{\ifmmode\mathbf{#1}\else\textbf{#1}\fi}
\newcommand{\BestD}[1]{#1}
\newcommand{\BestE}[1]{#1}
\newcommand{\BestF}[1]{#1}
\newcommand{\BestG}[1]{#1}
\newcommand{\BestH}[1]{#1}
\newcommand{\BestI}[1]{#1}
\newcommand{\BestJ}[1]{\ifmmode\mathbf{#1}\else\textbf{#1}\fi}
\newcommand{\BestK}[1]{\ifmmode\mathbf{#1}\else\textbf{#1}\fi}
\newcommand{\BestL}[1]{#1}
\newcommand{\BestM}[1]{#1}
\newcommand{\BestN}[1]{#1}
\newcommand{\BestO}[1]{#1}
\newcommand{\BestP}[1]{#1}
\newcommand{\BestQ}[1]{#1}
\newcommand{\BestR}[1]{#1}
\newcommand{\BestS}[1]{#1}
\newcommand{\BestT}[1]{#1}
\newcommand{\BestU}[1]{#1}
\newcommand{\BestV}[1]{#1}
\newcommand{\BestW}[1]{#1}
\newcommand{\BestX}[1]{#1}
\newcommand{\BestY}[1]{#1}
\newcommand{\BestZ}[1]{#1}
\newcommand{\BestAA}[1]{#1}
\newcommand{\BestAB}[1]{#1}
\newcommand{\BestAC}[1]{#1}
\newcommand{\BestAD}[1]{#1}
\newcommand{\BestAE}[1]{#1}
\newcommand{\BestAF}[1]{#1}
\newcommand{\BestAG}[1]{#1}
\newcommand{\BestAH}[1]{#1}
\newcommand{\BestAI}[1]{#1}
\newcommand{\BestAJ}[1]{#1}
\newcommand{\BestAK}[1]{#1}
\newcommand{\BestAL}[1]{#1}
\newcommand{\BestAM}[1]{#1}
\newcommand{\BestAN}[1]{#1}
\newcommand{\BestAO}[1]{#1}
\newcommand{\BestAP}[1]{#1}
\newcommand{\BestAQ}[1]{#1}
\newcommand{\BestAR}[1]{#1}
\newcommand{\BestAS}[1]{#1}
\newcommand{\BestAT}[1]{#1}
\newcommand{\BestAU}[1]{#1}
\newcommand{\BestAV}[1]{#1}
\newcommand{\BestAW}[1]{#1}
\newcommand{\BestAX}[1]{#1}
\newcommand{\BestAY}[1]{#1}
\newcommand{\BestAZ}[1]{#1}
\newcommand{\BestBA}[1]{#1}
\newcommand{\BestBB}[1]{#1}
\newcommand{\BestBC}[1]{#1}
\newcommand{\BestBD}[1]{#1}
\newcommand{\BestBE}[1]{#1}
\newcommand{\BestBF}[1]{#1}
\newcommand{\BestBG}[1]{#1}
\newcommand{\BestBH}[1]{#1}
\newcommand{\BestBI}[1]{#1}
\newcommand{\BestBJ}[1]{#1}
\newcommand{\BestBK}[1]{#1}
\newcommand{\BestBL}[1]{#1}
\newcommand{\BestBM}[1]{#1}
\newcommand{\BestBN}[1]{#1}
\newcommand{\BestBO}[1]{#1}
\newcommand{\BestBP}[1]{#1}
\newcommand{\BestBQ}[1]{#1}
\newcommand{\BestBR}[1]{#1}
\newcommand{\BestBS}[1]{\ifmmode\mathbf{#1}\else\textbf{#1}\fi}
\newcommand{\BestBT}[1]{\ifmmode\mathbf{#1}\else\textbf{#1}\fi}
\newcommand{\BestBU}[1]{\ifmmode\mathbf{#1}\else\textbf{#1}\fi}
\newcommand{\BestBV}[1]{#1}
\newcommand{\BestBW}[1]{#1}
\newcommand{\BestBX}[1]{#1}
\newcommand{\BestBY}[1]{#1}
\newcommand{\BestBZ}[1]{#1}
\newcommand{\BestCA}[1]{#1}
\newcommand{\BestCB}[1]{#1}
\newcommand{\BestCC}[1]{#1}
\newcommand{\BestCD}[1]{#1}
\newcommand{\BestCE}[1]{#1}
\newcommand{\BestCF}[1]{#1}
\newcommand{\BestCG}[1]{#1}
\newcommand{\BestCH}[1]{#1}
\newcommand{\BestCI}[1]{#1}
\newcommand{\BestCJ}[1]{#1}
\newcommand{\BestCK}[1]{#1}
\newcommand{\BestCL}[1]{#1}
\newcommand{\BestCM}[1]{#1}
\newcommand{\BestCN}[1]{#1}
\newcommand{\BestCO}[1]{#1}
\newcommand{\BestCP}[1]{#1}
\newcommand{\BestCQ}[1]{#1}
\newcommand{\BestCR}[1]{#1}
\newcommand{\BestCS}[1]{#1}
\newcommand{\BestCT}[1]{#1}
\newcommand{\BestCU}[1]{#1}
\newcommand{\BestCV}[1]{#1}
\newcommand{\BestCW}[1]{#1}
\newcommand{\BestCX}[1]{#1}
\newcommand{\BestCY}[1]{#1}
\newcommand{\BestCZ}[1]{#1}
\newcommand{\BestDA}[1]{#1}
\newcommand{\BestDB}[1]{#1}
\newcommand{\BestDC}[1]{#1}
\newcommand{\BestDD}[1]{#1}
\newcommand{\BestDE}[1]{#1}
\newcommand{\BestDF}[1]{#1}
\newcommand{\BestDG}[1]{#1}
\newcommand{\BestDH}[1]{#1}
\newcommand{\BestDI}[1]{#1}
\newcommand{\BestDJ}[1]{#1}
\newcommand{\BestDK}[1]{#1}
\newcommand{\BestDL}[1]{#1}
\newcommand{\BestDM}[1]{#1}
\newcommand{\BestDN}[1]{#1}
\newcommand{\BestDO}[1]{#1}
\newcommand{\BestDP}[1]{#1}
\newcommand{\BestDQ}[1]{\ifmmode\mathbf{#1}\else\textbf{#1}\fi}
\newcommand{\BestDR}[1]{#1}
\newcommand{\BestDS}[1]{#1}
\newcommand{\BestDT}[1]{#1}
\newcommand{\BestDU}[1]{#1}
\newcommand{\BestDV}[1]{#1}
\newcommand{\BestDW}[1]{#1}
\newcommand{\BestDX}[1]{#1}
\newcommand{\BestDY}[1]{\ifmmode\mathbf{#1}\else\textbf{#1}\fi}
\newcommand{\BestDZ}[1]{#1}
\newcommand{\BestEA}[1]{\ifmmode\mathbf{#1}\else\textbf{#1}\fi}
\newcommand{\BestEB}[1]{#1}
\newcommand{\BestEC}[1]{#1}
\newcommand{\BestED}[1]{#1}
\newcommand{\BestEE}[1]{\ifmmode\mathbf{#1}\else\textbf{#1}\fi}
\newcommand{\BestEF}[1]{#1}
\newcommand{\BestEG}[1]{#1}
\newcommand{\BestEH}[1]{#1}
\newcommand{\BestEI}[1]{#1}
\newcommand{\BestEJ}[1]{#1}
\newcommand{\BestEK}[1]{#1}
\newcommand{\BestEL}[1]{#1}
\newcommand{\BestEM}[1]{#1}
\newcommand{\BestEN}[1]{#1}
\newcommand{\BestEO}[1]{#1}
\newcommand{\BestEP}[1]{#1}
\newcommand{\BestEQ}[1]{#1}
\newcommand{\BestER}[1]{\ifmmode\mathbf{#1}\else\textbf{#1}\fi}
\newcommand{\BestES}[1]{#1}
\newcommand{\BestET}[1]{#1}
\newcommand{\BestEU}[1]{#1}
\newcommand{\BestEV}[1]{#1}
\newcommand{\BestEW}[1]{#1}
\newcommand{\BestEX}[1]{\ifmmode\mathbf{#1}\else\textbf{#1}\fi}
\newcommand{\BestEY}[1]{\ifmmode\mathbf{#1}\else\textbf{#1}\fi}
\newcommand{\BestEZ}[1]{#1}
\newcommand{\BestFA}[1]{#1}
\newcommand{\BestFB}[1]{#1}
\newcommand{\BestFC}[1]{#1}
\newcommand{\BestFD}[1]{#1}
\newcommand{\BestFE}[1]{#1}
\newcommand{\BestFF}[1]{#1}
\newcommand{\BestFG}[1]{#1}
\newcommand{\BestFH}[1]{#1}
\newcommand{\BestFI}[1]{#1}
\newcommand{\BestFJ}[1]{#1}
\newcommand{\BestFK}[1]{#1}
\newcommand{\BestFL}[1]{\ifmmode\mathbf{#1}\else\textbf{#1}\fi}
\newcommand{\BestFM}[1]{#1}
\newcommand{\BestFN}[1]{#1}
\newcommand{\BestFO}[1]{#1}
\newcommand{\BestFP}[1]{#1}
\newcommand{\BestFQ}[1]{#1}
\newcommand{\BestFR}[1]{\ifmmode\mathbf{#1}\else\textbf{#1}\fi}
\newcommand{\BestFS}[1]{#1}
\newcommand{\BestFT}[1]{#1}
\newcommand{\BestFU}[1]{#1}
\newcommand{\BestFV}[1]{#1}
\newcommand{\BestFW}[1]{#1}
\newcommand{\BestFX}[1]{#1}
\newcommand{\BestFY}[1]{\ifmmode\mathbf{#1}\else\textbf{#1}\fi}
\newcommand{\BestFZ}[1]{#1}
\newcommand{\BestGA}[1]{#1}
\newcommand{\BestGB}[1]{#1}
\newcommand{\BestGC}[1]{\ifmmode\mathbf{#1}\else\textbf{#1}\fi}
\newcommand{\BestGD}[1]{#1}
\newcommand{\BestGE}[1]{#1}
\newcommand{\BestGF}[1]{#1}
\newcommand{\BestGG}[1]{#1}
\newcommand{\BestGH}[1]{#1}
\newcommand{\BestGI}[1]{#1}
\newcommand{\BestGJ}[1]{#1}
\newcommand{\BestGK}[1]{#1}
\newcommand{\BestGL}[1]{#1}
\newcommand{\BestGM}[1]{#1}
\newcommand{\BestGN}[1]{#1}
\newcommand{\BestGO}[1]{#1}
\newcommand{\BestGP}[1]{#1}
\newcommand{\BestGQ}[1]{#1}
\newcommand{\BestGR}[1]{#1}
\newcommand{\BestGS}[1]{#1}
\newcommand{\BestGT}[1]{#1}
\newcommand{\BestGU}[1]{#1}
\newcommand{\BestGV}[1]{#1}
\newcommand{\BestGW}[1]{#1}
\newcommand{\BestGX}[1]{#1}
\newcommand{\BestGY}[1]{#1}
\newcommand{\BestGZ}[1]{#1}
\newcommand{\BestHA}[1]{#1}
\newcommand{\BestHB}[1]{#1}
\newcommand{\BestHC}[1]{#1}
\newcommand{\BestHD}[1]{#1}
\newcommand{\BestHE}[1]{#1}
\newcommand{\BestHF}[1]{#1}
\newcommand{\BestHG}[1]{#1}
\newcommand{\BestHH}[1]{#1}
\newcommand{\BestHI}[1]{#1}
\newcommand{\BestHJ}[1]{#1}
\newcommand{\BestHK}[1]{#1}
\newcommand{\BestHL}[1]{#1}
\newcommand{\BestHM}[1]{#1}
\newcommand{\BestHN}[1]{#1}
\newcommand{\BestHO}[1]{#1}
\newcommand{\BestHP}[1]{#1}
\newcommand{\BestHQ}[1]{#1}
\newcommand{\BestHR}[1]{#1}
\newcommand{\BestHS}[1]{#1}
\newcommand{\BestHT}[1]{#1}
\newcommand{\BestHU}[1]{#1}
\newcommand{\BestHV}[1]{#1}
\newcommand{\BestHW}[1]{#1}
\newcommand{\BestHX}[1]{#1}
\newcommand{\BestHY}[1]{#1}
\newcommand{\BestHZ}[1]{#1}
\newcommand{\BestIA}[1]{#1}
\newcommand{\BestIB}[1]{#1}
\newcommand{\BestIC}[1]{#1}
\newcommand{\BestID}[1]{#1}
\newcommand{\BestIE}[1]{\ifmmode\mathbf{#1}\else\textbf{#1}\fi}
\newcommand{\BestIF}[1]{#1}
\newcommand{\BestIG}[1]{#1}
\newcommand{\BestIH}[1]{#1}
\newcommand{\BestII}[1]{\ifmmode\mathbf{#1}\else\textbf{#1}\fi}
\newcommand{\BestIJ}[1]{#1}
\newcommand{\BestIK}[1]{#1}
\newcommand{\BestIL}[1]{#1}
\newcommand{\BestIM}[1]{#1}
\newcommand{\BestIN}[1]{#1}
\newcommand{\BestIO}[1]{#1}
\newcommand{\BestIP}[1]{\ifmmode\mathbf{#1}\else\textbf{#1}\fi}
\newcommand{\BestIQ}[1]{#1}
\newcommand{\BestIR}[1]{#1}
\newcommand{\BestIS}[1]{\ifmmode\mathbf{#1}\else\textbf{#1}\fi}
\newcommand{\BestIT}[1]{#1}
\newcommand{\BestIU}[1]{#1}
\newcommand{\BestIV}[1]{#1}
\newcommand{\BestIW}[1]{#1}
\newcommand{\BestIX}[1]{#1}
\newcommand{\BestIY}[1]{#1}
\newcommand{\BestIZ}[1]{#1}
\newcommand{\BestJA}[1]{#1}
\newcommand{\BestJB}[1]{#1}
\newcommand{\BestJC}[1]{#1}
\newcommand{\BestJD}[1]{#1}
\newcommand{\BestJE}[1]{#1}
\newcommand{\BestJF}[1]{#1}
\newcommand{\BestJG}[1]{#1}
\newcommand{\BestJH}[1]{#1}
\newcommand{\BestJI}[1]{#1}
\newcommand{\BestJJ}[1]{#1}
\newcommand{\BestJK}[1]{#1}
\newcommand{\BestJL}[1]{#1}
\newcommand{\BestJM}[1]{#1}
\newcommand{\BestJN}[1]{#1}
\newcommand{\BestJO}[1]{#1}
\newcommand{\BestJP}[1]{#1}
\newcommand{\BestJQ}[1]{#1}
\newcommand{\BestJR}[1]{#1}
\newcommand{\BestJS}[1]{#1}
\newcommand{\BestJT}[1]{#1}
\newcommand{\BestJU}[1]{#1}
\newcommand{\BestJV}[1]{#1}
\newcommand{\BestJW}[1]{#1}
\newcommand{\BestJX}[1]{#1}
\newcommand{\BestJY}[1]{#1}
\newcommand{\BestJZ}[1]{#1}
\newcommand{\BestKA}[1]{\ifmmode\mathbf{#1}\else\textbf{#1}\fi}
\newcommand{\BestKB}[1]{#1}
\newcommand{\BestKC}[1]{\ifmmode\mathbf{#1}\else\textbf{#1}\fi}
\newcommand{\BestKD}[1]{#1}
\newcommand{\BestKE}[1]{#1}
\newcommand{\BestKF}[1]{#1}
\newcommand{\BestKG}[1]{#1}
\newcommand{\BestKH}[1]{#1}
\newcommand{\BestKI}[1]{#1}
\newcommand{\BestKJ}[1]{#1}
\newcommand{\BestKK}[1]{#1}
\newcommand{\BestKL}[1]{#1}
\newcommand{\BestKM}[1]{#1}
\newcommand{\BestKN}[1]{#1}
\newcommand{\BestKO}[1]{#1}
\newcommand{\BestKP}[1]{#1}
\newcommand{\BestKQ}[1]{#1}
\newcommand{\BestKR}[1]{#1}
\newcommand{\BestKS}[1]{#1}
\newcommand{\BestKT}[1]{\ifmmode\mathbf{#1}\else\textbf{#1}\fi}
\newcommand{\BestKU}[1]{#1}
\newcommand{\BestKV}[1]{\ifmmode\mathbf{#1}\else\textbf{#1}\fi}
\newcommand{\BestKW}[1]{#1}
\newcommand{\BestKX}[1]{#1}
\newcommand{\BestKY}[1]{#1}
\newcommand{\BestKZ}[1]{#1}
\newcommand{\BestLA}[1]{\ifmmode\mathbf{#1}\else\textbf{#1}\fi}
\newcommand{\BestLB}[1]{#1}
\newcommand{\BestLC}[1]{#1}
\newcommand{\BestLD}[1]{#1}
\newcommand{\BestLE}[1]{#1}
\newcommand{\BestLF}[1]{#1}
\newcommand{\BestLG}[1]{#1}
\newcommand{\BestLH}[1]{\ifmmode\mathbf{#1}\else\textbf{#1}\fi}
\newcommand{\BestLI}[1]{#1}
\newcommand{\BestLJ}[1]{#1}
\newcommand{\BestLK}[1]{#1}
\newcommand{\BestLL}[1]{\ifmmode\mathbf{#1}\else\textbf{#1}\fi}
\newcommand{\BestLM}[1]{#1}
\newcommand{\BestLN}[1]{#1}
\newcommand{\BestLO}[1]{#1}
\newcommand{\BestLP}[1]{#1}
\newcommand{\BestLQ}[1]{#1}
\newcommand{\BestLR}[1]{#1}
\newcommand{\BestLS}[1]{\ifmmode\mathbf{#1}\else\textbf{#1}\fi}
\newcommand{\BestLT}[1]{#1}
\newcommand{\BestLU}[1]{#1}
\newcommand{\BestLV}[1]{#1}
\newcommand{\BestLW}[1]{#1}
\newcommand{\BestLX}[1]{#1}
\newcommand{\BestLY}[1]{#1}
\newcommand{\BestLZ}[1]{#1}
\newcommand{\BestMA}[1]{#1}
\newcommand{\BestMB}[1]{#1}
\newcommand{\BestMC}[1]{#1}
\newcommand{\BestMD}[1]{#1}
\newcommand{\BestME}[1]{#1}
\newcommand{\BestMF}[1]{#1}
\newcommand{\BestMG}[1]{#1}
\newcommand{\BestMH}[1]{#1}
\newcommand{\BestMI}[1]{#1}
\newcommand{\BestMJ}[1]{#1}
\newcommand{\BestMK}[1]{#1}
\newcommand{\BestML}[1]{#1}
\newcommand{\BestMM}[1]{#1}
\newcommand{\BestMN}[1]{#1}
\newcommand{\BestMO}[1]{#1}
\newcommand{\BestMP}[1]{#1}
\newcommand{\BestMQ}[1]{#1}
\newcommand{\BestMR}[1]{#1}
\newcommand{\BestMS}[1]{#1}
\newcommand{\BestMT}[1]{#1}
\newcommand{\BestMU}[1]{#1}
\newcommand{\BestMV}[1]{#1}
\newcommand{\BestMW}[1]{#1}
\newcommand{\BestMX}[1]{#1}
\newcommand{\BestMY}[1]{#1}
\newcommand{\BestMZ}[1]{#1}
\newcommand{\BestNA}[1]{\ifmmode\mathbf{#1}\else\textbf{#1}\fi}
\newcommand{\BestNB}[1]{\ifmmode\mathbf{#1}\else\textbf{#1}\fi}
\newcommand{\BestNC}[1]{\ifmmode\mathbf{#1}\else\textbf{#1}\fi}
\newcommand{\BestND}[1]{\ifmmode\mathbf{#1}\else\textbf{#1}\fi}
\newcommand{\BestNE}[1]{\ifmmode\mathbf{#1}\else\textbf{#1}\fi}
\newcommand{\BestNF}[1]{\ifmmode\mathbf{#1}\else\textbf{#1}\fi}
\newcommand{\BestNG}[1]{#1}
\newcommand{\BestNH}[1]{#1}
\newcommand{\BestNI}[1]{#1}
\newcommand{\BestNJ}[1]{#1}
\newcommand{\BestNK}[1]{#1}
\newcommand{\BestNL}[1]{#1}
\newcommand{\BestNM}[1]{#1}
\newcommand{\BestNN}[1]{#1}
\newcommand{\BestNO}[1]{#1}
\newcommand{\BestNP}[1]{#1}
\newcommand{\BestNQ}[1]{#1}
\newcommand{\BestNR}[1]{#1}
\newcommand{\BestNS}[1]{#1}
\newcommand{\BestNT}[1]{#1}
\newcommand{\BestNU}[1]{#1}
\newcommand{\BestNV}[1]{#1}
\newcommand{\BestNW}[1]{#1}
\newcommand{\BestNX}[1]{#1}
\newcommand{\BestNY}[1]{#1}
\newcommand{\BestNZ}[1]{#1}
\newcommand{\BestOA}[1]{#1}
\newcommand{\BestOB}[1]{#1}
\newcommand{\BestOC}[1]{#1}
\newcommand{\BestOD}[1]{#1}
\newcommand{\BestOE}[1]{#1}
\newcommand{\BestOF}[1]{#1}
\newcommand{\BestOG}[1]{#1}
\newcommand{\BestOH}[1]{#1}
\newcommand{\BestOI}[1]{#1}
\newcommand{\BestOJ}[1]{#1}
\newcommand{\BestOK}[1]{#1}
\newcommand{\BestOL}[1]{#1}
\newcommand{\BestOM}[1]{\ifmmode\mathbf{#1}\else\textbf{#1}\fi}
\newcommand{\BestON}[1]{\ifmmode\mathbf{#1}\else\textbf{#1}\fi}
\newcommand{\BestOO}[1]{\ifmmode\mathbf{#1}\else\textbf{#1}\fi}
\newcommand{\BestOP}[1]{#1}
\newcommand{\BestOQ}[1]{#1}
\newcommand{\BestOR}[1]{#1}
\newcommand{\BestOS}[1]{#1}
\newcommand{\BestOT}[1]{#1}
\newcommand{\BestOU}[1]{#1}
\newcommand{\BestOV}[1]{#1}
\newcommand{\BestOW}[1]{\ifmmode\mathbf{#1}\else\textbf{#1}\fi}
\newcommand{\BestOX}[1]{\ifmmode\mathbf{#1}\else\textbf{#1}\fi}
\newcommand{\BestOY}[1]{#1}
\newcommand{\BestOZ}[1]{#1}
\newcommand{\BestPA}[1]{#1}
\newcommand{\BestPB}[1]{#1}
\newcommand{\BestPC}[1]{#1}
\newcommand{\BestPD}[1]{#1}
\newcommand{\BestPE}[1]{#1}
\newcommand{\BestPF}[1]{#1}
\newcommand{\BestPG}[1]{#1}
\newcommand{\BestPH}[1]{#1}
\newcommand{\BestPI}[1]{#1}
\newcommand{\BestPJ}[1]{#1}
\newcommand{\BestPK}[1]{#1}
\newcommand{\BestPL}[1]{#1}
\newcommand{\BestPM}[1]{#1}
\newcommand{\BestPN}[1]{#1}
\newcommand{\BestPO}[1]{#1}
\newcommand{\BestPP}[1]{#1}
\newcommand{\BestPQ}[1]{#1}
\newcommand{\BestPR}[1]{#1}
\newcommand{\BestPS}[1]{#1}
\newcommand{\BestPT}[1]{#1}
\newcommand{\BestPU}[1]{#1}
\newcommand{\BestPV}[1]{#1}
\newcommand{\BestPW}[1]{#1}
\newcommand{\BestPX}[1]{#1}
\newcommand{\BestPY}[1]{#1}
\newcommand{\BestPZ}[1]{#1}
\newcommand{\BestQA}[1]{#1}
\newcommand{\BestQB}[1]{#1}
\newcommand{\BestQC}[1]{#1}
\newcommand{\BestQD}[1]{#1}
\newcommand{\BestQE}[1]{#1}
\newcommand{\BestQF}[1]{#1}
\newcommand{\BestQG}[1]{#1}
\newcommand{\BestQH}[1]{#1}
\newcommand{\BestQI}[1]{#1}
\newcommand{\BestQJ}[1]{#1}
\newcommand{\BestQK}[1]{#1}
\newcommand{\BestQL}[1]{#1}
\newcommand{\BestQM}[1]{#1}
\newcommand{\BestQN}[1]{#1}
\newcommand{\BestQO}[1]{#1}
\newcommand{\BestQP}[1]{#1}
\newcommand{\BestQQ}[1]{#1}
\newcommand{\BestQR}[1]{#1}
\newcommand{\BestQS}[1]{#1}
\newcommand{\BestQT}[1]{#1}
\newcommand{\BestQU}[1]{#1}
\newcommand{\BestQV}[1]{#1}
\newcommand{\BestQW}[1]{#1}
\newcommand{\BestQX}[1]{#1}
\newcommand{\BestQY}[1]{#1}
\newcommand{\BestQZ}[1]{#1}
\newcommand{\BestRA}[1]{#1}
\newcommand{\BestRB}[1]{#1}
\newcommand{\BestRC}[1]{#1}
\newcommand{\BestRD}[1]{#1}
\newcommand{\BestRE}[1]{\ifmmode\mathbf{#1}\else\textbf{#1}\fi}
\newcommand{\BestRF}[1]{\ifmmode\mathbf{#1}\else\textbf{#1}\fi}
\newcommand{\BestRG}[1]{#1}
\newcommand{\BestRH}[1]{#1}
\newcommand{\BestRI}[1]{#1}
\newcommand{\BestRJ}[1]{#1}
\newcommand{\BestRK}[1]{#1}
\newcommand{\BestRL}[1]{\ifmmode\mathbf{#1}\else\textbf{#1}\fi}
\newcommand{\BestRM}[1]{\ifmmode\mathbf{#1}\else\textbf{#1}\fi}
\newcommand{\BestRN}[1]{\ifmmode\mathbf{#1}\else\textbf{#1}\fi}
\newcommand{\BestRO}[1]{#1}
\newcommand{\BestRP}[1]{#1}
\newcommand{\BestRQ}[1]{#1}
\newcommand{\BestRR}[1]{#1}
\newcommand{\BestRS}[1]{#1}
\newcommand{\BestRT}[1]{#1}
\newcommand{\BestRU}[1]{#1}
\newcommand{\BestRV}[1]{#1}
\newcommand{\BestRW}[1]{#1}
\newcommand{\BestRX}[1]{#1}
\newcommand{\BestRY}[1]{\ifmmode\mathbf{#1}\else\textbf{#1}\fi}
\newcommand{\BestRZ}[1]{\ifmmode\mathbf{#1}\else\textbf{#1}\fi}
\newcommand{\BestSA}[1]{#1}
\newcommand{\BestSB}[1]{#1}
\newcommand{\BestSC}[1]{#1}
\newcommand{\BestSD}[1]{#1}
\newcommand{\BestSE}[1]{#1}
\newcommand{\BestSF}[1]{#1}
\newcommand{\BestSG}[1]{#1}
\newcommand{\BestSH}[1]{#1}
\newcommand{\BestSI}[1]{#1}
\newcommand{\BestSJ}[1]{\ifmmode\mathbf{#1}\else\textbf{#1}\fi}
\newcommand{\BestSK}[1]{#1}
\newcommand{\BestSL}[1]{#1}
\newcommand{\BestSM}[1]{#1}
\newcommand{\BestSN}[1]{#1}
\newcommand{\BestSO}[1]{#1}
\newcommand{\BestSP}[1]{#1}
\newcommand{\BestSQ}[1]{\ifmmode\mathbf{#1}\else\textbf{#1}\fi}
\newcommand{\BestSR}[1]{#1}
\newcommand{\BestSS}[1]{#1}
\newcommand{\BestST}[1]{#1}
\newcommand{\BestSU}[1]{#1}
\newcommand{\BestSV}[1]{#1}
\newcommand{\BestSW}[1]{#1}
\newcommand{\BestSX}[1]{\ifmmode\mathbf{#1}\else\textbf{#1}\fi}
\newcommand{\BestSY}[1]{#1}
\newcommand{\BestSZ}[1]{#1}
\newcommand{\BestTA}[1]{#1}
\newcommand{\BestTB}[1]{#1}
\newcommand{\BestTC}[1]{#1}
\newcommand{\BestTD}[1]{\ifmmode\mathbf{#1}\else\textbf{#1}\fi}
\newcommand{\BestTE}[1]{#1}
\newcommand{\BestTF}[1]{#1}
\newcommand{\BestTG}[1]{#1}
\newcommand{\BestTH}[1]{#1}
\newcommand{\BestTI}[1]{#1}
\newcommand{\BestTJ}[1]{#1}
\newcommand{\BestTK}[1]{#1}
\newcommand{\BestTL}[1]{#1}
\newcommand{\BestTM}[1]{\ifmmode\mathbf{#1}\else\textbf{#1}\fi}
\newcommand{\BestTN}[1]{#1}
\newcommand{\BestTO}[1]{#1}
\newcommand{\BestTP}[1]{#1}
\newcommand{\BestTQ}[1]{#1}
\newcommand{\BestTR}[1]{#1}
\newcommand{\BestTS}[1]{\ifmmode\mathbf{#1}\else\textbf{#1}\fi}
\newcommand{\BestTT}[1]{#1}
\newcommand{\BestTU}[1]{#1}
\newcommand{\BestTV}[1]{#1}
\newcommand{\BestTW}[1]{#1}
\newcommand{\BestTX}[1]{#1}
\newcommand{\BestTY}[1]{#1}
\newcommand{\BestTZ}[1]{\ifmmode\mathbf{#1}\else\textbf{#1}\fi}
\newcommand{\BestUA}[1]{#1}
\newcommand{\BestUB}[1]{#1}
\newcommand{\BestUC}[1]{#1}
\newcommand{\BestUD}[1]{#1}
\newcommand{\BestUE}[1]{#1}
\newcommand{\BestUF}[1]{#1}
\newcommand{\BestUG}[1]{#1}
\newcommand{\BestUH}[1]{#1}
\newcommand{\BestUI}[1]{#1}
\newcommand{\BestUJ}[1]{#1}
\newcommand{\BestUK}[1]{#1}
\newcommand{\BestUL}[1]{#1}
\newcommand{\BestUM}[1]{#1}
\newcommand{\BestUN}[1]{#1}
\newcommand{\BestUO}[1]{\ifmmode\mathbf{#1}\else\textbf{#1}\fi}
\newcommand{\BestUP}[1]{#1}
\newcommand{\BestUQ}[1]{#1}
\newcommand{\BestUR}[1]{#1}
\newcommand{\BestUS}[1]{#1}
\newcommand{\BestUT}[1]{#1}
\newcommand{\BestUU}[1]{#1}
\newcommand{\BestUV}[1]{\ifmmode\mathbf{#1}\else\textbf{#1}\fi}
\newcommand{\BestUW}[1]{#1}
\newcommand{\BestUX}[1]{#1}
\newcommand{\BestUY}[1]{#1}
\newcommand{\BestUZ}[1]{#1}
\newcommand{\BestVA}[1]{#1}
\newcommand{\BestVB}[1]{#1}
\newcommand{\BestVC}[1]{\ifmmode\mathbf{#1}\else\textbf{#1}\fi}
\newcommand{\BestVD}[1]{#1}
\newcommand{\BestVE}[1]{#1}
\newcommand{\BestVF}[1]{#1}
\newcommand{\BestVG}[1]{#1}
\newcommand{\BestVH}[1]{#1}
\newcommand{\BestVI}[1]{#1}
\newcommand{\BestVJ}[1]{\ifmmode\mathbf{#1}\else\textbf{#1}\fi}
\newcommand{\BestVK}[1]{#1}
\newcommand{\BestVL}[1]{#1}
\newcommand{\BestVM}[1]{#1}
\newcommand{\BestVN}[1]{#1}
\newcommand{\BestVO}[1]{#1}
\newcommand{\BestVP}[1]{\ifmmode\mathbf{#1}\else\textbf{#1}\fi}
\newcommand{\BestVQ}[1]{#1}
\newcommand{\BestVR}[1]{#1}
\newcommand{\BestVS}[1]{#1}
\newcommand{\BestVT}[1]{#1}
\newcommand{\BestVU}[1]{\ifmmode\mathbf{#1}\else\textbf{#1}\fi}
\newcommand{\BestVV}[1]{#1}
\newcommand{\BestVW}[1]{#1}
\newcommand{\BestVX}[1]{#1}
\newcommand{\BestVY}[1]{\ifmmode\mathbf{#1}\else\textbf{#1}\fi}
\newcommand{\BestVZ}[1]{#1}
\newcommand{\BestWA}[1]{#1}
\newcommand{\BestWB}[1]{#1}
\newcommand{\BestWC}[1]{#1}
\newcommand{\BestWD}[1]{#1}
\newcommand{\BestWE}[1]{\ifmmode\mathbf{#1}\else\textbf{#1}\fi}
\newcommand{\BestWF}[1]{#1}
\newcommand{\BestWG}[1]{#1}
\newcommand{\BestWH}[1]{#1}
\newcommand{\BestWI}[1]{#1}
\newcommand{\BestWJ}[1]{#1}
\newcommand{\BestWK}[1]{#1}
\newcommand{\BestWL}[1]{\ifmmode\mathbf{#1}\else\textbf{#1}\fi}
\newcommand{\BestWM}[1]{#1}
\newcommand{\BestWN}[1]{#1}
\newcommand{\BestWO}[1]{#1}
\newcommand{\BestWP}[1]{#1}
\newcommand{\BestWQ}[1]{#1}
\newcommand{\BestWR}[1]{#1}
\newcommand{\BestWS}[1]{#1}
\newcommand{\BestWT}[1]{#1}
\newcommand{\BestWU}[1]{#1}
\newcommand{\BestWV}[1]{#1}
\newcommand{\BestWW}[1]{#1}
\newcommand{\BestWX}[1]{#1}
\newcommand{\BestWY}[1]{#1}
\newcommand{\BestWZ}[1]{#1}
\newcommand{\BestXA}[1]{#1}
\newcommand{\BestXB}[1]{#1}
\newcommand{\BestXC}[1]{#1}
\newcommand{\BestXD}[1]{#1}
\newcommand{\BestXE}[1]{#1}
\newcommand{\BestXF}[1]{#1}
\newcommand{\BestXG}[1]{#1}
\newcommand{\BestXH}[1]{#1}
\newcommand{\BestXI}[1]{\ifmmode\mathbf{#1}\else\textbf{#1}\fi}
\newcommand{\BestXJ}[1]{\ifmmode\mathbf{#1}\else\textbf{#1}\fi}
\newcommand{\BestXK}[1]{#1}
\newcommand{\BestXL}[1]{#1}
\newcommand{\BestXM}[1]{\ifmmode\mathbf{#1}\else\textbf{#1}\fi}
\newcommand{\BestXN}[1]{\ifmmode\mathbf{#1}\else\textbf{#1}\fi}
\newcommand{\BestXO}[1]{\ifmmode\mathbf{#1}\else\textbf{#1}\fi}
\newcommand{\BestXP}[1]{\ifmmode\mathbf{#1}\else\textbf{#1}\fi}
\newcommand{\BestXQ}[1]{\ifmmode\mathbf{#1}\else\textbf{#1}\fi}
\newcommand{\BestXR}[1]{\ifmmode\mathbf{#1}\else\textbf{#1}\fi}
\newcommand{\BestXS}[1]{\ifmmode\mathbf{#1}\else\textbf{#1}\fi}
\newcommand{\BestXT}[1]{\ifmmode\mathbf{#1}\else\textbf{#1}\fi}
\newcommand{\BestXU}[1]{\ifmmode\mathbf{#1}\else\textbf{#1}\fi}
\newcommand{\BestXV}[1]{\ifmmode\mathbf{#1}\else\textbf{#1}\fi}
\newcommand{\BestXW}[1]{\ifmmode\mathbf{#1}\else\textbf{#1}\fi}
\newcommand{\BestXX}[1]{\ifmmode\mathbf{#1}\else\textbf{#1}\fi}
\newcommand{\BestXY}[1]{\ifmmode\mathbf{#1}\else\textbf{#1}\fi}
\newcommand{\BestXZ}[1]{\ifmmode\mathbf{#1}\else\textbf{#1}\fi}
\newcommand{\BestYA}[1]{\ifmmode\mathbf{#1}\else\textbf{#1}\fi}
\newcommand{\BestYB}[1]{\ifmmode\mathbf{#1}\else\textbf{#1}\fi}
\newcommand{\BestYC}[1]{#1}
\newcommand{\BestYD}[1]{#1}
\newcommand{\BestYE}[1]{#1}
\newcommand{\BestYF}[1]{#1}
\newcommand{\BestYG}[1]{#1}
\newcommand{\BestYH}[1]{#1}
\newcommand{\BestYI}[1]{#1}
\newcommand{\BestYJ}[1]{#1}
\newcommand{\BestYK}[1]{#1}
\newcommand{\BestYL}[1]{#1}
\newcommand{\BestYM}[1]{#1}
\newcommand{\BestYN}[1]{#1}
\newcommand{\BestYO}[1]{#1}
\newcommand{\BestYP}[1]{#1}
\newcommand{\BestYQ}[1]{#1}
\newcommand{\BestYR}[1]{#1}
\newcommand{\BestYS}[1]{#1}
\newcommand{\BestYT}[1]{#1}
\newcommand{\BestYU}[1]{#1}
\newcommand{\BestYV}[1]{#1}
\newcommand{\BestYW}[1]{#1}
\newcommand{\BestYX}[1]{#1}
\newcommand{\BestYY}[1]{#1}
\newcommand{\BestYZ}[1]{#1}
\newcommand{\BestZA}[1]{\ifmmode\mathbf{#1}\else\textbf{#1}\fi}
\newcommand{\BestZB}[1]{#1}
\newcommand{\BestZC}[1]{#1}
\newcommand{\BestZD}[1]{#1}
\newcommand{\BestZE}[1]{#1}
\newcommand{\BestZF}[1]{#1}
\newcommand{\BestZG}[1]{\ifmmode\mathbf{#1}\else\textbf{#1}\fi}
\newcommand{\BestZH}[1]{#1}
\newcommand{\BestZI}[1]{#1}
\newcommand{\BestZJ}[1]{#1}
\newcommand{\BestZK}[1]{#1}
\newcommand{\BestZL}[1]{#1}
\newcommand{\BestZM}[1]{#1}
\newcommand{\BestZN}[1]{\ifmmode\mathbf{#1}\else\textbf{#1}\fi}
\newcommand{\BestZO}[1]{#1}
\newcommand{\BestZP}[1]{#1}
\newcommand{\BestZQ}[1]{#1}
\newcommand{\BestZR}[1]{#1}
\newcommand{\BestZS}[1]{#1}
\newcommand{\BestZT}[1]{\ifmmode\mathbf{#1}\else\textbf{#1}\fi}
\newcommand{\BestZU}[1]{#1}
\newcommand{\BestZV}[1]{#1}
\newcommand{\BestZW}[1]{#1}
\newcommand{\BestZX}[1]{#1}
\newcommand{\BestZY}[1]{#1}
\newcommand{\BestZZ}[1]{\ifmmode\mathbf{#1}\else\textbf{#1}\fi}
\newcommand{\BestAAA}[1]{#1}
\newcommand{\BestAAB}[1]{#1}
\newcommand{\BestAAC}[1]{#1}
\newcommand{\BestAAD}[1]{#1}
\newcommand{\BestAAE}[1]{#1}
\newcommand{\BestAAF}[1]{#1}
\newcommand{\BestAAG}[1]{#1}
\newcommand{\BestAAH}[1]{#1}
\newcommand{\BestAAI}[1]{#1}
\newcommand{\BestAAJ}[1]{#1}
\newcommand{\BestAAK}[1]{\ifmmode\mathbf{#1}\else\textbf{#1}\fi}
\newcommand{\BestAAL}[1]{#1}
\newcommand{\BestAAM}[1]{#1}
\newcommand{\BestAAN}[1]{#1}
\newcommand{\BestAAO}[1]{#1}
\newcommand{\BestAAP}[1]{#1}
\newcommand{\BestAAQ}[1]{\ifmmode\mathbf{#1}\else\textbf{#1}\fi}
\newcommand{\BestAAR}[1]{#1}
\newcommand{\BestAAS}[1]{#1}
\newcommand{\BestAAT}[1]{#1}
\newcommand{\BestAAU}[1]{#1}
\newcommand{\BestAAV}[1]{\ifmmode\mathbf{#1}\else\textbf{#1}\fi}
\newcommand{\BestAAW}[1]{#1}
\newcommand{\BestAAX}[1]{#1}
\newcommand{\BestAAY}[1]{#1}
\newcommand{\BestAAZ}[1]{#1}
\newcommand{\BestABA}[1]{#1}
\newcommand{\BestABB}[1]{#1}
\newcommand{\BestABC}[1]{\ifmmode\mathbf{#1}\else\textbf{#1}\fi}
\newcommand{\BestABD}[1]{\ifmmode\mathbf{#1}\else\textbf{#1}\fi}
\newcommand{\BestABE}[1]{\ifmmode\mathbf{#1}\else\textbf{#1}\fi}
\newcommand{\BestABF}[1]{\ifmmode\mathbf{#1}\else\textbf{#1}\fi}
\newcommand{\BestABG}[1]{\ifmmode\mathbf{#1}\else\textbf{#1}\fi}
\newcommand{\BestABH}[1]{#1}
\newcommand{\BestABI}[1]{#1}
\newcommand{\BestABJ}[1]{\ifmmode\mathbf{#1}\else\textbf{#1}\fi}
\newcommand{\BestABK}[1]{#1}
\newcommand{\BestABL}[1]{#1}
\newcommand{\BestABM}[1]{#1}
\newcommand{\BestABN}[1]{\ifmmode\mathbf{#1}\else\textbf{#1}\fi}
\newcommand{\BestABO}[1]{\ifmmode\mathbf{#1}\else\textbf{#1}\fi}
\newcommand{\BestABP}[1]{#1}
\newcommand{\BestABQ}[1]{#1}
\newcommand{\BestABR}[1]{#1}
\newcommand{\BestABS}[1]{#1}
\newcommand{\BestABT}[1]{#1}
\newcommand{\BestABU}[1]{#1}
\newcommand{\BestABV}[1]{#1}
\newcommand{\BestABW}[1]{#1}
\newcommand{\BestABX}[1]{#1}
\newcommand{\BestABY}[1]{#1}
\newcommand{\BestABZ}[1]{#1}
\newcommand{\BestACA}[1]{#1}
\newcommand{\BestACB}[1]{#1}
\newcommand{\BestACC}[1]{#1}
\newcommand{\BestACD}[1]{#1}
\newcommand{\BestACE}[1]{#1}
\newcommand{\BestACF}[1]{#1}
\newcommand{\BestACG}[1]{#1}
\newcommand{\BestACH}[1]{#1}
\newcommand{\BestACI}[1]{#1}
\newcommand{\BestACJ}[1]{#1}
\newcommand{\BestACK}[1]{#1}
\newcommand{\BestACL}[1]{#1}
\newcommand{\BestACM}[1]{#1}
\newcommand{\BestACN}[1]{#1}
\newcommand{\BestACO}[1]{#1}
\newcommand{\BestACP}[1]{#1}
\newcommand{\BestACQ}[1]{#1}
\newcommand{\BestACR}[1]{#1}
\newcommand{\BestACS}[1]{#1}
\newcommand{\BestACT}[1]{#1}
\newcommand{\BestACU}[1]{#1}
\newcommand{\BestACV}[1]{#1}
\newcommand{\BestACW}[1]{#1}
\newcommand{\BestACX}[1]{#1}
\newcommand{\BestACY}[1]{#1}
\newcommand{\BestACZ}[1]{#1}
\newcommand{\BestADA}[1]{#1}
\newcommand{\BestADB}[1]{#1}
\newcommand{\BestADC}[1]{#1}
\newcommand{\BestADD}[1]{#1}
\newcommand{\BestADE}[1]{#1}
\newcommand{\BestADF}[1]{#1}
\newcommand{\BestADG}[1]{#1}
\newcommand{\BestADH}[1]{#1}
\newcommand{\BestADI}[1]{#1}
\newcommand{\BestADJ}[1]{#1}
\newcommand{\BestADK}[1]{#1}
\newcommand{\BestADL}[1]{#1}
\newcommand{\BestADM}[1]{#1}
\newcommand{\BestADN}[1]{#1}
\newcommand{\BestADO}[1]{#1}
\newcommand{\BestADP}[1]{#1}
\newcommand{\BestADQ}[1]{#1}
\newcommand{\BestADR}[1]{#1}
\newcommand{\BestADS}[1]{#1}
\newcommand{\BestADT}[1]{#1}
\newcommand{\BestADU}[1]{#1}
\newcommand{\BestADV}[1]{#1}
\newcommand{\BestADW}[1]{#1}
\newcommand{\BestADX}[1]{#1}
\newcommand{\BestADY}[1]{#1}
\newcommand{\BestADZ}[1]{\ifmmode\mathbf{#1}\else\textbf{#1}\fi}
\newcommand{\BestAEA}[1]{#1}
\newcommand{\BestAEB}[1]{#1}
\newcommand{\BestAEC}[1]{\ifmmode\mathbf{#1}\else\textbf{#1}\fi}
\newcommand{\BestAED}[1]{#1}
\newcommand{\BestAEE}[1]{#1}
\newcommand{\BestAEF}[1]{#1}
\newcommand{\BestAEG}[1]{#1}
\newcommand{\BestAEH}[1]{\ifmmode\mathbf{#1}\else\textbf{#1}\fi}
\newcommand{\BestAEI}[1]{#1}
\newcommand{\BestAEJ}[1]{#1}
\newcommand{\BestAEK}[1]{#1}
\newcommand{\BestAEL}[1]{#1}
\newcommand{\BestAEM}[1]{#1}
\newcommand{\BestAEN}[1]{#1}
\newcommand{\BestAEO}[1]{#1}
\newcommand{\BestAEP}[1]{#1}
\newcommand{\BestAEQ}[1]{#1}
\newcommand{\BestAER}[1]{#1}
\newcommand{\BestAES}[1]{#1}
\newcommand{\BestAET}[1]{#1}
\newcommand{\BestAEU}[1]{#1}
\newcommand{\BestAEV}[1]{#1}
\newcommand{\BestAEW}[1]{#1}
\newcommand{\BestAEX}[1]{#1}
\newcommand{\BestAEY}[1]{#1}
\newcommand{\BestAEZ}[1]{#1}
\newcommand{\BestAFA}[1]{#1}
\newcommand{\BestAFB}[1]{#1}
\newcommand{\BestAFC}[1]{#1}
\newcommand{\BestAFD}[1]{#1}
\newcommand{\BestAFE}[1]{#1}
\newcommand{\BestAFF}[1]{#1}
\newcommand{\BestAFG}[1]{#1}
\newcommand{\BestAFH}[1]{#1}
\newcommand{\BestAFI}[1]{#1}
\newcommand{\BestAFJ}[1]{#1}
\newcommand{\BestAFK}[1]{#1}
\newcommand{\BestAFL}[1]{#1}
\newcommand{\BestAFM}[1]{#1}
\newcommand{\BestAFN}[1]{#1}
\newcommand{\BestAFO}[1]{#1}
\newcommand{\BestAFP}[1]{#1}
\newcommand{\BestAFQ}[1]{#1}
\newcommand{\BestAFR}[1]{\ifmmode\mathbf{#1}\else\textbf{#1}\fi}
\newcommand{\BestAFS}[1]{#1}
\newcommand{\BestAFT}[1]{#1}
\newcommand{\BestAFU}[1]{#1}
\newcommand{\BestAFV}[1]{\ifmmode\mathbf{#1}\else\textbf{#1}\fi}
\newcommand{\BestAFW}[1]{#1}
\newcommand{\BestAFX}[1]{#1}
\newcommand{\BestAFY}[1]{#1}
\newcommand{\BestAFZ}[1]{#1}
\newcommand{\BestAGA}[1]{\ifmmode\mathbf{#1}\else\textbf{#1}\fi}
\newcommand{\BestAGB}[1]{#1}
\newcommand{\BestAGC}[1]{\ifmmode\mathbf{#1}\else\textbf{#1}\fi}
\newcommand{\BestAGD}[1]{#1}
\newcommand{\BestAGE}[1]{#1}
\newcommand{\BestAGF}[1]{\ifmmode\mathbf{#1}\else\textbf{#1}\fi}
\newcommand{\BestAGG}[1]{#1}
\newcommand{\BestAGH}[1]{#1}
\newcommand{\BestAGI}[1]{#1}
\newcommand{\BestAGJ}[1]{#1}
\newcommand{\BestAGK}[1]{#1}
\newcommand{\BestAGL}[1]{#1}
\newcommand{\BestAGM}[1]{#1}
\newcommand{\BestAGN}[1]{#1}
\newcommand{\BestAGO}[1]{#1}
\newcommand{\BestAGP}[1]{#1}
\newcommand{\BestAGQ}[1]{#1}
\newcommand{\BestAGR}[1]{#1}
\newcommand{\BestAGS}[1]{#1}
\newcommand{\BestAGT}[1]{#1}
\newcommand{\BestAGU}[1]{#1}
\newcommand{\BestAGV}[1]{#1}
\newcommand{\BestAGW}[1]{#1}
\newcommand{\BestAGX}[1]{#1}
\newcommand{\BestAGY}[1]{\ifmmode\mathbf{#1}\else\textbf{#1}\fi}
\newcommand{\BestAGZ}[1]{#1}
\newcommand{\BestAHA}[1]{#1}
\newcommand{\BestAHB}[1]{#1}
\newcommand{\BestAHC}[1]{#1}
\newcommand{\BestAHD}[1]{#1}
\newcommand{\BestAHE}[1]{#1}
\newcommand{\BestAHF}[1]{#1}
\newcommand{\BestAHG}[1]{#1}
\newcommand{\BestAHH}[1]{#1}
\newcommand{\BestAHI}[1]{#1}
\newcommand{\BestAHJ}[1]{#1}
\newcommand{\BestAHK}[1]{#1}
\newcommand{\BestAHL}[1]{#1}
\newcommand{\BestAHM}[1]{#1}
\newcommand{\BestAHN}[1]{#1}
\newcommand{\BestAHO}[1]{#1}
\newcommand{\BestAHP}[1]{#1}
\newcommand{\BestAHQ}[1]{#1}
\newcommand{\BestAHR}[1]{#1}
\newcommand{\BestAHS}[1]{#1}
\newcommand{\BestAHT}[1]{#1}
\newcommand{\BestAHU}[1]{#1}
\newcommand{\BestAHV}[1]{#1}
\newcommand{\BestAHW}[1]{#1}
\newcommand{\BestAHX}[1]{#1}
\newcommand{\BestAHY}[1]{#1}
\newcommand{\BestAHZ}[1]{#1}
\newcommand{\BestAIA}[1]{#1}
\newcommand{\BestAIB}[1]{#1}
\newcommand{\BestAIC}[1]{#1}
\newcommand{\BestAID}[1]{#1}
\newcommand{\BestAIE}[1]{#1}
\newcommand{\BestAIF}[1]{#1}
\newcommand{\BestAIG}[1]{#1}
\newcommand{\BestAIH}[1]{#1}
\newcommand{\BestAII}[1]{#1}
\newcommand{\BestAIJ}[1]{\ifmmode\mathbf{#1}\else\textbf{#1}\fi}
\newcommand{\BestAIK}[1]{#1}
\newcommand{\BestAIL}[1]{#1}
\newcommand{\BestAIM}[1]{\ifmmode\mathbf{#1}\else\textbf{#1}\fi}
\newcommand{\BestAIN}[1]{#1}
\newcommand{\BestAIO}[1]{#1}
\newcommand{\BestAIP}[1]{#1}
\newcommand{\BestAIQ}[1]{#1}
\newcommand{\BestAIR}[1]{#1}
\newcommand{\BestAIS}[1]{#1}
\newcommand{\BestAIT}[1]{#1}
\newcommand{\BestAIU}[1]{#1}
\newcommand{\BestAIV}[1]{#1}
\newcommand{\BestAIW}[1]{#1}
\newcommand{\BestAIX}[1]{#1}
\newcommand{\BestAIY}[1]{#1}
\newcommand{\BestAIZ}[1]{#1}
\newcommand{\BestAJA}[1]{#1}
\newcommand{\BestAJB}[1]{#1}
\newcommand{\BestAJC}[1]{#1}
\newcommand{\BestAJD}[1]{#1}
\newcommand{\BestAJE}[1]{#1}
\newcommand{\BestAJF}[1]{#1}
\newcommand{\BestAJG}[1]{#1}
\newcommand{\BestAJH}[1]{#1}
\newcommand{\BestAJI}[1]{#1}
\newcommand{\BestAJJ}[1]{#1}
\newcommand{\BestAJK}[1]{#1}
\newcommand{\BestAJL}[1]{#1}
\newcommand{\BestAJM}[1]{#1}
\newcommand{\BestAJN}[1]{#1}
\newcommand{\BestAJO}[1]{#1}
\newcommand{\BestAJP}[1]{#1}
\newcommand{\BestAJQ}[1]{#1}
\newcommand{\BestAJR}[1]{#1}
\newcommand{\BestAJS}[1]{#1}
\newcommand{\BestAJT}[1]{#1}
\newcommand{\BestAJU}[1]{#1}
\newcommand{\BestAJV}[1]{#1}
\newcommand{\BestAJW}[1]{#1}
\newcommand{\BestAJX}[1]{#1}
\newcommand{\BestAJY}[1]{#1}
\newcommand{\BestAJZ}[1]{#1}
\newcommand{\BestAKA}[1]{#1}
\newcommand{\BestAKB}[1]{#1}
\newcommand{\BestAKC}[1]{#1}
\newcommand{\BestAKD}[1]{#1}
\newcommand{\BestAKE}[1]{#1}
\newcommand{\BestAKF}[1]{#1}
\newcommand{\BestAKG}[1]{\ifmmode\mathbf{#1}\else\textbf{#1}\fi}
\newcommand{\BestAKH}[1]{\ifmmode\mathbf{#1}\else\textbf{#1}\fi}
\newcommand{\BestAKI}[1]{#1}
\newcommand{\BestAKJ}[1]{#1}
\newcommand{\BestAKK}[1]{#1}
\newcommand{\BestAKL}[1]{#1}
\newcommand{\BestAKM}[1]{#1}
\newcommand{\BestAKN}[1]{#1}
\newcommand{\BestAKO}[1]{#1}
\newcommand{\BestAKP}[1]{#1}
\newcommand{\BestAKQ}[1]{\ifmmode\mathbf{#1}\else\textbf{#1}\fi}
\newcommand{\BestAKR}[1]{#1}
\newcommand{\BestAKS}[1]{#1}
\newcommand{\BestAKT}[1]{#1}
\newcommand{\BestAKU}[1]{#1}
\newcommand{\BestAKV}[1]{\ifmmode\mathbf{#1}\else\textbf{#1}\fi}
\newcommand{\BestAKW}[1]{\ifmmode\mathbf{#1}\else\textbf{#1}\fi}
\newcommand{\BestAKX}[1]{\ifmmode\mathbf{#1}\else\textbf{#1}\fi}
\newcommand{\BestAKY}[1]{#1}
\newcommand{\BestAKZ}[1]{#1}
\newcommand{\BestALA}[1]{\ifmmode\mathbf{#1}\else\textbf{#1}\fi}
\newcommand{\BestALB}[1]{\ifmmode\mathbf{#1}\else\textbf{#1}\fi}
\newcommand{\BestALC}[1]{#1}
\newcommand{\BestALD}[1]{#1}
\newcommand{\BestALE}[1]{\ifmmode\mathbf{#1}\else\textbf{#1}\fi}
\newcommand{\BestALF}[1]{\ifmmode\mathbf{#1}\else\textbf{#1}\fi}
\newcommand{\BestALG}[1]{#1}
\newcommand{\BestALH}[1]{#1}
\newcommand{\BestALI}[1]{\ifmmode\mathbf{#1}\else\textbf{#1}\fi}
\newcommand{\BestALJ}[1]{\ifmmode\mathbf{#1}\else\textbf{#1}\fi}
\newcommand{\BestALK}[1]{#1}
\newcommand{\BestALL}[1]{#1}

\newcommand{\checkpointbreak}{\textunderscore\allowbreak}

\definecolor{linkblue}{rgb}{0.21,0.49,0.74}
\usepackage[breaklinks,colorlinks,allcolors=linkblue]{hyperref}
\usepackage[capitalize]{cleveref}
\crefname{section}{Sec.}{Secs.}
\crefname{table}{Tab.}{Tabs.}
\providecommand{\etal}{\emph{et al}.\xspace}
\newcommand{\AccGridFirst}{0.494\xspace}
\newcommand{\AccGridFirstLo}{0.477\xspace}
\newcommand{\AccGridFirstHi}{0.511\xspace}
\newcommand{\AccRandomFirst}{0.498\xspace}
\newcommand{\AccRandomFirstLo}{0.481\xspace}
\newcommand{\AccRandomFirstHi}{0.515\xspace}
\newcommand{\AccHardFirst}{0.424\xspace}
\newcommand{\AccHardFirstLo}{0.398\xspace}
\newcommand{\AccHardFirstHi}{0.450\xspace}
\newcommand{\AccEasyFirst}{0.579\xspace}
\newcommand{\AccEasyFirstLo}{0.551\xspace}
\newcommand{\AccEasyFirstHi}{0.607\xspace}
\newcommand{\DropFirst}{0.074\xspace}
\newcommand{\DropFirstLow}{0.052\xspace}
\newcommand{\AdmittedFirst}{18\xspace}

\newcommand{\AccGridNew}{0.499\xspace}
\newcommand{\AccGridNewLo}{0.482\xspace}
\newcommand{\AccGridNewHi}{0.514\xspace}
\newcommand{\AccRandomNew}{0.503\xspace}
\newcommand{\AccRandomNewLo}{0.485\xspace}
\newcommand{\AccRandomNewHi}{0.518\xspace}
\newcommand{\AccHardNew}{0.437\xspace}
\newcommand{\AccHardNewLo}{0.410\xspace}
\newcommand{\AccHardNewHi}{0.466\xspace}
\newcommand{\AccEasyNew}{0.577\xspace}
\newcommand{\AccEasyNewLo}{0.549\xspace}
\newcommand{\AccEasyNewHi}{0.605\xspace}
\newcommand{\DropNew}{0.066\xspace}
\newcommand{\DropNewLow}{0.042\xspace}
\newcommand{\AdmittedNew}{14\xspace}

\newcommand{\DropFirstHigh}{0.096\xspace}
\newcommand{\DropNewHigh}{0.088\xspace}
\newcommand{\Margin}{0.03\xspace}
\newcommand{\KnownFirst}{10\xspace}
\newcommand{\KnownNew}{3\xspace}
\newcommand{\PanelsCompared}{203\xspace}
\newcommand{\ExRefFirst}{0.331\xspace}

\newcommand{\ShamLimit}{8\xspace}
\newcommand{\ShamFloor}{10\xspace}
\newcommand{\RandomPairings}{200\xspace}
\newcommand{\WithinPoint}{0.01\xspace}
\newcommand{\ShuffleLimit}{10\xspace}
\newcommand{\GainNew}{0.074\xspace}
\newcommand{\GainNewLow}{0.052\xspace}

\newcommand{\PairedMean}{0.0007\xspace}

\newcommand{\PairedModels}{14\xspace}
\newcommand{\BothAdmitted}{13\xspace}
\newcommand{\PairedBoth}{0.0012\xspace}
\newcommand{\SelDropNew}{0.066\xspace}

\newcommand{\SelAdmittedNew}{17\xspace}
\newcommand{\LineGridFirst}{0.760\xspace}
\newcommand{\LineHardFirst}{0.897\xspace}
\newcommand{\LineEasyFirst}{0.416\xspace}
\newcommand{\PairsFirst}{153\xspace}
\newcommand{\ResolvedFirst}{127\xspace}

\newcommand{\ReversedFirst}{0\xspace}
\newcommand{\LineGridNew}{0.873\xspace}
\newcommand{\LineHardNew}{0.928\xspace}
\newcommand{\LineEasyNew}{0.661\xspace}
\newcommand{\PairsNew}{91\xspace}
\newcommand{\ResolvedNew}{79\xspace}

\newcommand{\ReversedNew}{0\xspace}
\newcommand{\LineGate}{0.8\xspace}
\newcommand{\LineMargin}{0.2\xspace}
\newcommand{\ReversalMargin}{0.02\xspace}
\newcommand{\ShareMargin}{0.1\xspace}
\newcommand{\MinResolved}{20\xspace}

\newcommand{\DropMinFirst}{0.013\xspace}
\newcommand{\DropMaxFirst}{0.112\xspace}

\newcommand{\RankFirst}{-0.91\xspace}
\newcommand{\RandomMaxFirst}{0.051\xspace}

\newcommand{\DropMinNew}{0.018\xspace}
\newcommand{\DropMaxNew}{0.115\xspace}

\newcommand{\RankNew}{-0.80\xspace}
\newcommand{\RandomMaxNew}{0.057\xspace}

\newcommand{\LabAFirst}{-0.013\xspace}
\newcommand{\LabBFirst}{0.072\xspace}
\newcommand{\LabCFirst}{-0.004\xspace}
\newcommand{\LabDFirst}{0.147\xspace}
\newcommand{\LabEFirst}{0.110\xspace}
\newcommand{\LabFFirst}{0.232\xspace}
\newcommand{\LabGFirst}{0.032\xspace}
\newcommand{\LabHFirst}{0.015\xspace}
\newcommand{\LabANew}{-0.018\xspace}
\newcommand{\LabBNew}{0.069\xspace}
\newcommand{\LabCNew}{0.013\xspace}
\newcommand{\LabDNew}{0.139\xspace}
\newcommand{\LabENew}{0.064\xspace}
\newcommand{\LabFNew}{0.229\xspace}
\newcommand{\LabGNew}{0.026\xspace}
\newcommand{\LabHNew}{0.002\xspace}
\newcommand{\EffAa}{-0.121\xspace}

\newcommand{\EffAb}{-0.133\xspace}

\newcommand{\EffAc}{-0.210\xspace}

\newcommand{\EffAd}{-0.189\xspace}

\newcommand{\EffAe}{-0.122\xspace}

\newcommand{\EffAf}{-0.094\xspace}

\newcommand{\EffAg}{-0.146\xspace}

\newcommand{\EffAh}{-0.121\xspace}

\newcommand{\EffBa}{-0.064\xspace}

\newcommand{\EffBb}{-0.037\xspace}

\newcommand{\EffBc}{-0.160\xspace}

\newcommand{\EffBd}{-0.133\xspace}

\newcommand{\EffBe}{-0.041\xspace}

\newcommand{\EffBf}{-0.097\xspace}

\newcommand{\EffBg}{-0.139\xspace}

\newcommand{\EffBh}{-0.060\xspace}

\newcommand{\EffCa}{-0.004\xspace}

\newcommand{\EffCb}{-0.001\xspace}

\newcommand{\EffCc}{-0.016\xspace}

\newcommand{\EffCd}{0.003\xspace}

\newcommand{\EffCe}{0.004\xspace}

\newcommand{\EffCf}{0.004\xspace}

\newcommand{\EffCg}{-0.021\xspace}

\newcommand{\EffCh}{0.002\xspace}

\newcommand{\EffDa}{-0.005\xspace}

\newcommand{\EffDb}{-0.023\xspace}

\newcommand{\EffDc}{-0.148\xspace}

\newcommand{\EffDd}{-0.189\xspace}

\newcommand{\EffDe}{-0.109\xspace}

\newcommand{\EffDf}{-0.214\xspace}

\newcommand{\EffDg}{-0.065\xspace}

\newcommand{\EffDh}{-0.013\xspace}

\newcommand{\EffEa}{-0.062\xspace}

\newcommand{\EffEb}{-0.141\xspace}

\newcommand{\EffEc}{-0.020\xspace}

\newcommand{\EffEd}{-0.239\xspace}

\newcommand{\EffEe}{-0.203\xspace}

\newcommand{\EffEf}{-0.237\xspace}

\newcommand{\EffEg}{-0.135\xspace}

\newcommand{\EffEh}{-0.080\xspace}

\newcommand{\EffFa}{-0.055\xspace}

\newcommand{\EffFb}{-0.394\xspace}

\newcommand{\EffFc}{-0.269\xspace}

\newcommand{\EffFd}{-0.441\xspace}

\newcommand{\EffFe}{-0.302\xspace}

\newcommand{\EffFf}{-0.365\xspace}

\newcommand{\EffFg}{-0.513\xspace}

\newcommand{\EffFh}{-0.042\xspace}

\newcommand{\EffGa}{-0.035\xspace}

\newcommand{\EffGb}{-0.008\xspace}

\newcommand{\EffGc}{-0.145\xspace}

\newcommand{\EffGd}{-0.090\xspace}

\newcommand{\EffGe}{-0.107\xspace}

\newcommand{\EffGf}{-0.086\xspace}

\newcommand{\EffGg}{-0.138\xspace}

\newcommand{\EffGh}{-0.029\xspace}

\newcommand{\EffHa}{-0.021\xspace}

\newcommand{\EffHb}{-0.033\xspace}

\newcommand{\EffHc}{0.010\xspace}

\newcommand{\EffHd}{-0.036\xspace}

\newcommand{\EffHe}{-0.031\xspace}

\newcommand{\EffHf}{-0.038\xspace}

\newcommand{\EffHg}{0.013\xspace}

\newcommand{\EffHh}{-0.018\xspace}

\newcommand{\EffMax}{0.013\xspace}
\newcommand{\EffPositive}{6\xspace}
\newcommand{\EffLowMax}{0.000\xspace}
\newcommand{\HardSum}{-1.39\xspace}
\newcommand{\EasySum}{-0.20\xspace}

\newcommand{\RelNamedFirst}{4\xspace}
\newcommand{\RelNamedNew}{4\xspace}
\newcommand{\RelNamedBoth}{3\xspace}
\newcommand{\RelMargin}{0.01\xspace}
\newcommand{\RelAaFirst}{0.021\xspace}
\newcommand{\RelAbFirst}{0.009\xspace}
\newcommand{\RelAcFirst}{-0.068\xspace}
\newcommand{\RelAdFirst}{-0.047\xspace}
\newcommand{\RelAeFirst}{0.020\xspace}
\newcommand{\RelAfFirst}{0.048\xspace}
\newcommand{\RelAgFirst}{-0.005\xspace}
\newcommand{\RelAhFirst}{0.021\xspace}
\newcommand{\RelBaFirst}{0.027\xspace}
\newcommand{\RelBbFirst}{0.054\xspace}
\newcommand{\RelBcFirst}{-0.068\xspace}
\newcommand{\RelBdFirst}{-0.041\xspace}
\newcommand{\RelBeFirst}{0.050\xspace}
\newcommand{\RelBfFirst}{-0.006\xspace}
\newcommand{\RelBgFirst}{-0.048\xspace}
\newcommand{\RelBhFirst}{0.032\xspace}
\newcommand{\RelCaFirst}{0.000\xspace}
\newcommand{\RelCbFirst}{0.003\xspace}
\newcommand{\RelCcFirst}{-0.012\xspace}
\newcommand{\RelCdFirst}{0.006\xspace}
\newcommand{\RelCeFirst}{0.007\xspace}
\newcommand{\RelCfFirst}{0.008\xspace}
\newcommand{\RelCgFirst}{-0.017\xspace}
\newcommand{\RelChFirst}{0.006\xspace}
\newcommand{\RelDaFirst}{0.090\xspace}
\newcommand{\RelDbFirst}{0.072\xspace}
\newcommand{\RelDcFirst}{-0.052\xspace}
\newcommand{\RelDdFirst}{-0.093\xspace}
\newcommand{\RelDeFirst}{-0.013\xspace}
\newcommand{\RelDfFirst}{-0.118\xspace}
\newcommand{\RelDgFirst}{0.030\xspace}
\newcommand{\RelDhFirst}{0.083\xspace}
\newcommand{\RelEaFirst}{0.077\xspace}
\newcommand{\RelEbFirst}{-0.001\xspace}
\newcommand{\RelEcFirst}{0.119\xspace}
\newcommand{\RelEdFirst}{-0.099\xspace}
\newcommand{\RelEeFirst}{-0.064\xspace}
\newcommand{\RelEfFirst}{-0.098\xspace}
\newcommand{\RelEgFirst}{0.005\xspace}
\newcommand{\RelEhFirst}{0.060\xspace}
\newcommand{\RelFaFirst}{0.243\xspace}
\newcommand{\RelFbFirst}{-0.096\xspace}
\newcommand{\RelFcFirst}{0.029\xspace}
\newcommand{\RelFdFirst}{-0.143\xspace}
\newcommand{\RelFeFirst}{-0.005\xspace}
\newcommand{\RelFfFirst}{-0.067\xspace}
\newcommand{\RelFgFirst}{-0.216\xspace}
\newcommand{\RelFhFirst}{0.256\xspace}
\newcommand{\RelGaFirst}{0.044\xspace}
\newcommand{\RelGbFirst}{0.071\xspace}
\newcommand{\RelGcFirst}{-0.065\xspace}
\newcommand{\RelGdFirst}{-0.011\xspace}
\newcommand{\RelGeFirst}{-0.027\xspace}
\newcommand{\RelGfFirst}{-0.006\xspace}
\newcommand{\RelGgFirst}{-0.058\xspace}
\newcommand{\RelGhFirst}{0.050\xspace}
\newcommand{\RelHaFirst}{-0.002\xspace}
\newcommand{\RelHbFirst}{-0.014\xspace}
\newcommand{\RelHcFirst}{0.030\xspace}
\newcommand{\RelHdFirst}{-0.016\xspace}
\newcommand{\RelHeFirst}{-0.011\xspace}
\newcommand{\RelHfFirst}{-0.019\xspace}
\newcommand{\RelHgFirst}{0.032\xspace}
\newcommand{\RelHhFirst}{0.001\xspace}
\newcommand{\RelAaNew}{0.001\xspace}
\newcommand{\RelAbNew}{0.009\xspace}
\newcommand{\RelAcNew}{-0.033\xspace}
\newcommand{\RelAdNew}{-0.043\xspace}
\newcommand{\RelAeNew}{0.009\xspace}
\newcommand{\RelAfNew}{0.055\xspace}
\newcommand{\RelAgNew}{-0.027\xspace}
\newcommand{\RelAhNew}{0.028\xspace}
\newcommand{\RelBaNew}{0.017\xspace}
\newcommand{\RelBbNew}{0.040\xspace}
\newcommand{\RelBcNew}{-0.063\xspace}
\newcommand{\RelBdNew}{-0.063\xspace}
\newcommand{\RelBeNew}{0.063\xspace}
\newcommand{\RelBfNew}{0.013\xspace}
\newcommand{\RelBgNew}{-0.035\xspace}
\newcommand{\RelBhNew}{0.030\xspace}
\newcommand{\RelCaNew}{-0.005\xspace}
\newcommand{\RelCbNew}{-0.003\xspace}
\newcommand{\RelCcNew}{-0.012\xspace}
\newcommand{\RelCdNew}{0.012\xspace}
\newcommand{\RelCeNew}{0.008\xspace}
\newcommand{\RelCfNew}{0.013\xspace}
\newcommand{\RelCgNew}{-0.015\xspace}
\newcommand{\RelChNew}{0.000\xspace}
\newcommand{\RelDaNew}{0.081\xspace}
\newcommand{\RelDbNew}{0.062\xspace}
\newcommand{\RelDcNew}{-0.016\xspace}
\newcommand{\RelDdNew}{-0.149\xspace}
\newcommand{\RelDeNew}{-0.003\xspace}
\newcommand{\RelDfNew}{-0.117\xspace}
\newcommand{\RelDgNew}{0.040\xspace}
\newcommand{\RelDhNew}{0.100\xspace}
\newcommand{\RelEaNew}{0.078\xspace}
\newcommand{\RelEbNew}{-0.002\xspace}
\newcommand{\RelEcNew}{0.131\xspace}
\newcommand{\RelEdNew}{-0.055\xspace}
\newcommand{\RelEeNew}{-0.072\xspace}
\newcommand{\RelEfNew}{-0.098\xspace}
\newcommand{\RelEgNew}{-0.011\xspace}
\newcommand{\RelEhNew}{0.029\xspace}
\newcommand{\RelFaNew}{0.249\xspace}
\newcommand{\RelFbNew}{-0.106\xspace}
\newcommand{\RelFcNew}{0.033\xspace}
\newcommand{\RelFdNew}{-0.162\xspace}
\newcommand{\RelFeNew}{0.014\xspace}
\newcommand{\RelFfNew}{-0.043\xspace}
\newcommand{\RelFgNew}{-0.235\xspace}
\newcommand{\RelFhNew}{0.250\xspace}
\newcommand{\RelGaNew}{0.043\xspace}
\newcommand{\RelGbNew}{0.036\xspace}
\newcommand{\RelGcNew}{-0.062\xspace}
\newcommand{\RelGdNew}{0.001\xspace}
\newcommand{\RelGeNew}{-0.024\xspace}
\newcommand{\RelGfNew}{-0.012\xspace}
\newcommand{\RelGgNew}{-0.040\xspace}
\newcommand{\RelGhNew}{0.058\xspace}
\newcommand{\RelHaNew}{-0.003\xspace}
\newcommand{\RelHbNew}{-0.005\xspace}
\newcommand{\RelHcNew}{0.018\xspace}
\newcommand{\RelHdNew}{-0.013\xspace}
\newcommand{\RelHeNew}{-0.003\xspace}
\newcommand{\RelHfNew}{-0.016\xspace}
\newcommand{\RelHgNew}{0.022\xspace}
\newcommand{\RelHhNew}{0.000\xspace}

\newcommand{\CoocAa}{0.55\xspace}
\newcommand{\CoocAb}{0.69\xspace}
\newcommand{\CoocAc}{0.22\xspace}
\newcommand{\CoocAd}{0.32\xspace}
\newcommand{\CoocAe}{0.03\xspace}
\newcommand{\CoocAf}{0.02\xspace}
\newcommand{\CoocAg}{0.61\xspace}
\newcommand{\CoocAh}{0.16\xspace}

\newcommand{\CoocBa}{0.60\xspace}
\newcommand{\CoocBb}{0.66\xspace}
\newcommand{\CoocBc}{0.19\xspace}
\newcommand{\CoocBd}{0.23\xspace}
\newcommand{\CoocBe}{0.02\xspace}
\newcommand{\CoocBf}{0.01\xspace}
\newcommand{\CoocBg}{0.69\xspace}
\newcommand{\CoocBh}{0.20\xspace}

\newcommand{\CoocCa}{0.18\xspace}
\newcommand{\CoocCb}{0.27\xspace}
\newcommand{\CoocCc}{0.28\xspace}
\newcommand{\CoocCd}{0.27\xspace}
\newcommand{\CoocCe}{0.03\xspace}
\newcommand{\CoocCf}{0.07\xspace}
\newcommand{\CoocCg}{0.11\xspace}
\newcommand{\CoocCh}{0.05\xspace}

\newcommand{\CoocDa}{0.58\xspace}
\newcommand{\CoocDb}{0.67\xspace}
\newcommand{\CoocDc}{0.19\xspace}
\newcommand{\CoocDd}{0.35\xspace}
\newcommand{\CoocDe}{0.02\xspace}
\newcommand{\CoocDf}{0.02\xspace}
\newcommand{\CoocDg}{0.58\xspace}
\newcommand{\CoocDh}{0.19\xspace}

\newcommand{\CoocEa}{0.03\xspace}
\newcommand{\CoocEb}{0.05\xspace}
\newcommand{\CoocEc}{0.45\xspace}
\newcommand{\CoocEd}{0.05\xspace}
\newcommand{\CoocEe}{0.00\xspace}
\newcommand{\CoocEf}{0.00\xspace}
\newcommand{\CoocEg}{0.02\xspace}
\newcommand{\CoocEh}{0.00\xspace}

\newcommand{\CoocFa}{0.64\xspace}
\newcommand{\CoocFb}{0.33\xspace}
\newcommand{\CoocFc}{0.05\xspace}
\newcommand{\CoocFd}{0.31\xspace}
\newcommand{\CoocFe}{0.01\xspace}
\newcommand{\CoocFf}{0.06\xspace}
\newcommand{\CoocFg}{0.37\xspace}
\newcommand{\CoocFh}{0.31\xspace}

\newcommand{\CoocGa}{0.47\xspace}
\newcommand{\CoocGb}{0.58\xspace}
\newcommand{\CoocGc}{0.12\xspace}
\newcommand{\CoocGd}{0.49\xspace}
\newcommand{\CoocGe}{0.03\xspace}
\newcommand{\CoocGf}{0.07\xspace}
\newcommand{\CoocGg}{0.28\xspace}
\newcommand{\CoocGh}{0.14\xspace}

\newcommand{\CoocHa}{0.59\xspace}
\newcommand{\CoocHb}{0.60\xspace}
\newcommand{\CoocHc}{0.18\xspace}
\newcommand{\CoocHd}{0.33\xspace}
\newcommand{\CoocHe}{0.02\xspace}
\newcommand{\CoocHf}{0.02\xspace}
\newcommand{\CoocHg}{0.53\xspace}
\newcommand{\CoocHh}{0.21\xspace}
\newcommand{\ClassesA}{9\xspace}
\newcommand{\ClassesB}{3\xspace}
\newcommand{\ClassesC}{118\xspace}
\newcommand{\ClassesD}{7\xspace}
\newcommand{\ClassesE}{7\xspace}

\newcommand{\TrainImages}{118,287\xspace}
\newcommand{\LicenceImages}{19,470\xspace}
\newcommand{\UnitsPerLabel}{100\xspace}
\newcommand{\LabelsPerSide}{8\xspace}
\newcommand{\Canvas}{224\xspace}
\newcommand{\FgLongSide}{134\xspace}
\newcommand{\MinFgLong}{96\xspace}
\newcommand{\MinMaskFill}{0.25\xspace}
\newcommand{\BgMinSide}{160\xspace}
\newcommand{\BgSideStep}{16\xspace}
\newcommand{\BgStride}{8\xspace}
\newcommand{\BgPurity}{0.9\xspace}
\newcommand{\BgUnlabelled}{0.05\xspace}
\newcommand{\Grey}{128\xspace}
\newcommand{\FgCandidates}{24,522\xspace}
\newcommand{\FgAdmitted}{7,410\xspace}
\newcommand{\FgSources}{3,198\xspace}
\newcommand{\BgSearched}{16,272\xspace}
\newcommand{\BgCandidates}{84,668\xspace}
\newcommand{\BgAdmitted}{11,966\xspace}
\newcommand{\CompositesAll}{6,400\xspace}
\newcommand{\PanelSize}{800\xspace}
\newcommand{\Cells}{64\xspace}
\newcommand{\UsedImages}{1,804\xspace}

\newcommand{\ImagesFirst}{1,600\xspace}
\newcommand{\ImagesNew}{1,600\xspace}
\newcommand{\Foregrounds}{800\xspace}
\newcommand{\Windows}{800\xspace}
\newcommand{\ZooRows}{1,556\xspace}
\newcommand{\ZooEligible}{710\xspace}
\newcommand{\ZooFamilies}{103\xspace}
\newcommand{\ZooBins}{8\xspace}
\newcommand{\InputSide}{224\xspace}
\newcommand{\Strata}{20\xspace}
\newcommand{\Models}{40\xspace}
\newcommand{\ImagesScored}{8,800\xspace}
\newcommand{\HalfSize}{20\xspace}
\newcommand{\TopMin}{63.1\xspace}
\newcommand{\TopMax}{88.9\xspace}
\newcommand{\CkptAa}{lcnet\_050.ra2\_in1k\xspace}
\newcommand{\TopAa}{63.1\xspace}

\newcommand{\DropAaFirst}{0.066\xspace}
\newcommand{\DropAaFirstLo}{0.040\xspace}
\newcommand{\DropAaFirstHi}{0.091\xspace}
\newcommand{\ShamAaFirst}{-10\xspace}
\newcommand{\CkptAb}{mobilenetv3\_small\_100.lamb\_in1k\xspace}
\newcommand{\TopAb}{67.6\xspace}

\newcommand{\DropAbFirst}{0.099\xspace}
\newcommand{\DropAbFirstLo}{0.072\xspace}
\newcommand{\DropAbFirstHi}{0.127\xspace}
\newcommand{\ShamAbFirst}{-8\xspace}
\newcommand{\CkptAc}{ghostnet\_100.in1k\xspace}
\newcommand{\TopAc}{74.0\xspace}

\newcommand{\DropAcFirst}{0.102\xspace}
\newcommand{\DropAcFirstLo}{0.073\xspace}
\newcommand{\DropAcFirstHi}{0.132\xspace}
\newcommand{\ShamAcFirst}{5\xspace}
\newcommand{\CkptAd}{vgg16\_bn.tv\_in1k\xspace}
\newcommand{\TopAd}{73.4\xspace}

\newcommand{\DropAdFirst}{0.110\xspace}
\newcommand{\DropAdFirstLo}{0.077\xspace}
\newcommand{\DropAdFirstHi}{0.143\xspace}
\newcommand{\ShamAdFirst}{-7\xspace}
\newcommand{\CkptAe}{mnasnet\_100.rmsp\_in1k\xspace}
\newcommand{\TopAe}{74.7\xspace}

\newcommand{\DropAeFirst}{0.106\xspace}
\newcommand{\DropAeFirstLo}{0.073\xspace}
\newcommand{\DropAeFirstHi}{0.138\xspace}
\newcommand{\ShamAeFirst}{5\xspace}
\newcommand{\CkptAf}{fbnetc\_100.rmsp\_in1k\xspace}
\newcommand{\TopAf}{75.1\xspace}

\newcommand{\DropAfFirst}{0.072\xspace}
\newcommand{\DropAfFirstLo}{0.040\xspace}
\newcommand{\DropAfFirstHi}{0.100\xspace}
\newcommand{\ShamAfFirst}{0\xspace}
\newcommand{\CkptAg}{starnet\_s2.in1k\xspace}
\newcommand{\TopAg}{74.7\xspace}

\newcommand{\DropAgFirst}{0.095\xspace}
\newcommand{\DropAgFirstLo}{0.062\xspace}
\newcommand{\DropAgFirstHi}{0.128\xspace}
\newcommand{\ShamAgFirst}{-6\xspace}
\newcommand{\CkptAh}{skresnet18.ra\_in1k\xspace}
\newcommand{\TopAh}{73.0\xspace}

\newcommand{\DropAhFirst}{0.104\xspace}
\newcommand{\DropAhFirstLo}{0.072\xspace}
\newcommand{\DropAhFirstHi}{0.136\xspace}
\newcommand{\ShamAhFirst}{6\xspace}
\newcommand{\CkptAi}{densenet121.ra\_in1k\xspace}
\newcommand{\TopAi}{75.6\xspace}

\newcommand{\DropAiFirst}{0.088\xspace}
\newcommand{\DropAiFirstLo}{0.056\xspace}
\newcommand{\DropAiFirstHi}{0.120\xspace}
\newcommand{\ShamAiFirst}{0\xspace}
\newcommand{\CkptAj}{spnasnet\_100.rmsp\_in1k\xspace}
\newcommand{\TopAj}{74.1\xspace}

\newcommand{\DropAjFirst}{0.101\xspace}
\newcommand{\DropAjFirstLo}{0.071\xspace}
\newcommand{\DropAjFirstHi}{0.131\xspace}
\newcommand{\ShamAjFirst}{3\xspace}
\newcommand{\CkptAk}{levit\_128s.fb\_dist\_in1k\xspace}
\newcommand{\TopAk}{76.5\xspace}

\newcommand{\DropAkFirst}{0.112\xspace}
\newcommand{\DropAkFirstLo}{0.077\xspace}
\newcommand{\DropAkFirstHi}{0.145\xspace}
\newcommand{\ShamAkFirst}{1\xspace}
\newcommand{\CkptAl}{fbnetv3\_d.ra2\_in1k\xspace}
\newcommand{\TopAl}{78.7\xspace}

\newcommand{\DropAlFirst}{0.083\xspace}
\newcommand{\DropAlFirstLo}{0.049\xspace}
\newcommand{\DropAlFirstHi}{0.117\xspace}
\newcommand{\ShamAlFirst}{1\xspace}
\newcommand{\CkptAm}{tf\_mixnet\_m.in1k\xspace}
\newcommand{\TopAm}{77.0\xspace}

\newcommand{\DropAmFirst}{0.098\xspace}
\newcommand{\DropAmFirstLo}{0.067\xspace}
\newcommand{\DropAmFirstHi}{0.129\xspace}
\newcommand{\ShamAmFirst}{4\xspace}
\newcommand{\CkptAn}{densenetblur121d.ra\_in1k\xspace}
\newcommand{\TopAn}{76.6\xspace}

\newcommand{\DropAnFirst}{0.102\xspace}
\newcommand{\DropAnFirstLo}{0.069\xspace}
\newcommand{\DropAnFirstHi}{0.137\xspace}
\newcommand{\ShamAnFirst}{1\xspace}
\newcommand{\CkptAo}{dla60x.in1k\xspace}
\newcommand{\TopAo}{78.2\xspace}

\newcommand{\DropAoFirst}{0.114\xspace}
\newcommand{\DropAoFirstLo}{0.080\xspace}
\newcommand{\DropAoFirstHi}{0.148\xspace}
\newcommand{\ShamAoFirst}{1\xspace}
\newcommand{\CkptAp}{repghostnet\_130.in1k\xspace}
\newcommand{\TopAp}{76.4\xspace}

\newcommand{\DropApFirst}{0.077\xspace}
\newcommand{\DropApFirstLo}{0.046\xspace}
\newcommand{\DropApFirstHi}{0.109\xspace}
\newcommand{\ShamApFirst}{-4\xspace}
\newcommand{\CkptAq}{cait\_xxs24\_224.fb\_dist\_in1k\xspace}
\newcommand{\TopAq}{78.4\xspace}

\newcommand{\DropAqFirst}{0.095\xspace}
\newcommand{\DropAqFirstLo}{0.061\xspace}
\newcommand{\DropAqFirstHi}{0.131\xspace}
\newcommand{\ShamAqFirst}{4\xspace}
\newcommand{\CkptAr}{fasternet\_t2.in1k\xspace}
\newcommand{\TopAr}{78.7\xspace}

\newcommand{\DropArFirst}{0.076\xspace}
\newcommand{\DropArFirstLo}{0.043\xspace}
\newcommand{\DropArFirstHi}{0.107\xspace}
\newcommand{\ShamArFirst}{-3\xspace}
\newcommand{\CkptAs}{coatnext\_nano\_rw\_224.sw\_in1k\xspace}
\newcommand{\TopAs}{82.0\xspace}

\newcommand{\DropAsFirst}{0.073\xspace}
\newcommand{\DropAsFirstLo}{0.037\xspace}
\newcommand{\DropAsFirstHi}{0.108\xspace}
\newcommand{\ShamAsFirst}{-4\xspace}
\newcommand{\CkptAt}{resmlp\_36\_224.fb\_distilled\_in1k\xspace}
\newcommand{\TopAt}{81.2\xspace}

\newcommand{\DropAtFirst}{0.077\xspace}
\newcommand{\DropAtFirstLo}{0.043\xspace}
\newcommand{\DropAtFirstHi}{0.111\xspace}
\newcommand{\ShamAtFirst}{4\xspace}
\newcommand{\CkptAu}{poolformer\_s24.sail\_in1k\xspace}
\newcommand{\TopAu}{80.3\xspace}

\newcommand{\DropAuFirst}{0.081\xspace}
\newcommand{\DropAuFirstLo}{0.047\xspace}
\newcommand{\DropAuFirstHi}{0.117\xspace}
\newcommand{\ShamAuFirst}{7\xspace}
\newcommand{\CkptAv}{wide\_resnet50\_2.racm\_in1k\xspace}
\newcommand{\TopAv}{81.5\xspace}

\newcommand{\DropAvFirst}{0.077\xspace}
\newcommand{\DropAvFirstLo}{0.042\xspace}
\newcommand{\DropAvFirstHi}{0.112\xspace}
\newcommand{\ShamAvFirst}{9\xspace}
\newcommand{\CkptAw}{poolformerv2\_s24.sail\_in1k\xspace}
\newcommand{\TopAw}{80.7\xspace}

\newcommand{\DropAwFirst}{0.071\xspace}
\newcommand{\DropAwFirstLo}{0.037\xspace}
\newcommand{\DropAwFirstHi}{0.106\xspace}
\newcommand{\ShamAwFirst}{10\xspace}
\newcommand{\CkptAx}{rexnet\_130.nav\_in1k\xspace}
\newcommand{\TopAx}{79.5\xspace}

\newcommand{\DropAxFirst}{0.090\xspace}
\newcommand{\DropAxFirstLo}{0.057\xspace}
\newcommand{\DropAxFirstHi}{0.124\xspace}
\newcommand{\ShamAxFirst}{1\xspace}
\newcommand{\CkptAy}{coatnet\_bn\_0\_rw\_224.sw\_in1k\xspace}
\newcommand{\TopAy}{82.4\xspace}

\newcommand{\DropAyFirst}{0.059\xspace}
\newcommand{\DropAyFirstLo}{0.024\xspace}
\newcommand{\DropAyFirstHi}{0.093\xspace}
\newcommand{\ShamAyFirst}{3\xspace}
\newcommand{\CkptAz}{resnetaa50.a1h\_in1k\xspace}
\newcommand{\TopAz}{80.6\xspace}

\newcommand{\DropAzFirst}{0.072\xspace}
\newcommand{\DropAzFirstLo}{0.038\xspace}
\newcommand{\DropAzFirstHi}{0.108\xspace}
\newcommand{\ShamAzFirst}{7\xspace}
\newcommand{\CkptBa}{inception\_next\_small.sail\_in1k\xspace}
\newcommand{\TopBa}{83.6\xspace}

\newcommand{\DropBaFirst}{0.025\xspace}
\newcommand{\DropBaFirstLo}{-0.011\xspace}
\newcommand{\DropBaFirstHi}{0.060\xspace}
\newcommand{\ShamBaFirst}{6\xspace}
\newcommand{\CkptBb}{regnetx\_160.tv2\_in1k\xspace}
\newcommand{\TopBb}{82.6\xspace}

\newcommand{\DropBbFirst}{0.061\xspace}
\newcommand{\DropBbFirstLo}{0.029\xspace}
\newcommand{\DropBbFirstHi}{0.096\xspace}
\newcommand{\ShamBbFirst}{3\xspace}
\newcommand{\CkptBc}{beitv2\_base\_patch16\_224.in1k\_ft\_in1k\xspace}
\newcommand{\TopBc}{85.6\xspace}

\newcommand{\DropBcFirst}{0.034\xspace}
\newcommand{\DropBcFirstLo}{0.000\xspace}
\newcommand{\DropBcFirstHi}{0.072\xspace}
\newcommand{\ShamBcFirst}{-5\xspace}
\newcommand{\CkptBd}{efficientformerv2\_l.snap\_dist\_in1k\xspace}
\newcommand{\TopBd}{83.6\xspace}

\newcommand{\DropBdFirst}{0.084\xspace}
\newcommand{\DropBdFirstLo}{0.049\xspace}
\newcommand{\DropBdFirstHi}{0.119\xspace}
\newcommand{\ShamBdFirst}{3\xspace}
\newcommand{\CkptBe}{focalnet\_base\_srf.ms\_in1k\xspace}
\newcommand{\TopBe}{83.8\xspace}

\newcommand{\DropBeFirst}{0.040\xspace}
\newcommand{\DropBeFirstLo}{0.005\xspace}
\newcommand{\DropBeFirstHi}{0.074\xspace}
\newcommand{\ShamBeFirst}{-1\xspace}
\newcommand{\CkptBf}{rexnetr\_200.sw\_in12k\_ft\_in1k\xspace}
\newcommand{\TopBf}{82.6\xspace}

\newcommand{\DropBfFirst}{0.068\xspace}
\newcommand{\DropBfFirstLo}{0.033\xspace}
\newcommand{\DropBfFirstHi}{0.104\xspace}
\newcommand{\ShamBfFirst}{0\xspace}
\newcommand{\CkptBg}{xcit\_small\_24\_p16\_224.fb\_in1k\xspace}
\newcommand{\TopBg}{82.6\xspace}

\newcommand{\DropBgFirst}{0.039\xspace}
\newcommand{\DropBgFirstLo}{0.004\xspace}
\newcommand{\DropBgFirstHi}{0.073\xspace}
\newcommand{\ShamBgFirst}{-2\xspace}
\newcommand{\CkptBh}{hgnetv2\_b6.ssld\_stage2\_ft\_in1k\xspace}
\newcommand{\TopBh}{86.2\xspace}

\newcommand{\DropBhFirst}{0.033\xspace}
\newcommand{\DropBhFirstLo}{-0.001\xspace}
\newcommand{\DropBhFirstHi}{0.067\xspace}
\newcommand{\ShamBhFirst}{-2\xspace}
\newcommand{\CkptBi}{volo\_d5\_224.sail\_in1k\xspace}
\newcommand{\TopBi}{86.1\xspace}

\newcommand{\DropBiFirst}{0.043\xspace}
\newcommand{\DropBiFirstLo}{0.007\xspace}
\newcommand{\DropBiFirstHi}{0.078\xspace}
\newcommand{\ShamBiFirst}{-2\xspace}
\newcommand{\CkptBj}{eva\_giant\_patch14\_224.clip\_ft\_in1k\xspace}
\newcommand{\TopBj}{88.9\xspace}

\newcommand{\DropBjFirst}{0.017\xspace}
\newcommand{\DropBjFirstLo}{-0.018\xspace}
\newcommand{\DropBjFirstHi}{0.052\xspace}
\newcommand{\ShamBjFirst}{2\xspace}
\newcommand{\CkptBk}{convnext\_xlarge.fb\_in22k\_ft\_in1k\xspace}
\newcommand{\TopBk}{87.0\xspace}

\newcommand{\DropBkFirst}{0.042\xspace}
\newcommand{\DropBkFirstLo}{0.010\xspace}
\newcommand{\DropBkFirstHi}{0.075\xspace}
\newcommand{\ShamBkFirst}{13\xspace}
\newcommand{\CkptBl}{beit3\_large\_patch16\_224.indomain\_in22k\_ft\_in1k\xspace}
\newcommand{\TopBl}{87.5\xspace}

\newcommand{\DropBlFirst}{0.002\xspace}
\newcommand{\DropBlFirstLo}{-0.032\xspace}
\newcommand{\DropBlFirstHi}{0.036\xspace}
\newcommand{\ShamBlFirst}{4\xspace}
\newcommand{\CkptBm}{caformer\_s36.sail\_in22k\_ft\_in1k\xspace}
\newcommand{\TopBm}{85.8\xspace}

\newcommand{\DropBmFirst}{0.049\xspace}
\newcommand{\DropBmFirstLo}{0.015\xspace}
\newcommand{\DropBmFirstHi}{0.082\xspace}
\newcommand{\ShamBmFirst}{11\xspace}
\newcommand{\CkptBn}{maxxvitv2\_rmlp\_base\_rw\_224.sw\_in12k\_ft\_in1k\xspace}
\newcommand{\TopBn}{86.6\xspace}

\newcommand{\DropBnFirst}{0.013\xspace}
\newcommand{\DropBnFirstLo}{-0.020\xspace}
\newcommand{\DropBnFirstHi}{0.049\xspace}
\newcommand{\ShamBnFirst}{2\xspace}
\newcommand{\DropAaNew}{0.065\xspace}
\newcommand{\DropAaNewLo}{0.040\xspace}
\newcommand{\DropAaNewHi}{0.088\xspace}
\newcommand{\ShamAaNew}{-3\xspace}
\newcommand{\DropAbNew}{0.085\xspace}
\newcommand{\DropAbNewLo}{0.058\xspace}
\newcommand{\DropAbNewHi}{0.113\xspace}
\newcommand{\ShamAbNew}{-1\xspace}
\newcommand{\DropAcNew}{0.080\xspace}
\newcommand{\DropAcNewLo}{0.052\xspace}
\newcommand{\DropAcNewHi}{0.108\xspace}
\newcommand{\ShamAcNew}{2\xspace}
\newcommand{\DropAdNew}{0.115\xspace}
\newcommand{\DropAdNewLo}{0.081\xspace}
\newcommand{\DropAdNewHi}{0.147\xspace}
\newcommand{\ShamAdNew}{2\xspace}
\newcommand{\DropAeNew}{0.097\xspace}
\newcommand{\DropAeNewLo}{0.065\xspace}
\newcommand{\DropAeNewHi}{0.127\xspace}
\newcommand{\ShamAeNew}{-6\xspace}
\newcommand{\DropAfNew}{0.073\xspace}
\newcommand{\DropAfNewLo}{0.043\xspace}
\newcommand{\DropAfNewHi}{0.103\xspace}
\newcommand{\ShamAfNew}{-3\xspace}
\newcommand{\DropAgNew}{0.078\xspace}
\newcommand{\DropAgNewLo}{0.045\xspace}
\newcommand{\DropAgNewHi}{0.110\xspace}
\newcommand{\ShamAgNew}{5\xspace}
\newcommand{\DropAhNew}{0.098\xspace}
\newcommand{\DropAhNewLo}{0.066\xspace}
\newcommand{\DropAhNewHi}{0.129\xspace}
\newcommand{\ShamAhNew}{0\xspace}
\newcommand{\DropAiNew}{0.090\xspace}
\newcommand{\DropAiNewLo}{0.056\xspace}
\newcommand{\DropAiNewHi}{0.123\xspace}
\newcommand{\ShamAiNew}{-2\xspace}
\newcommand{\DropAjNew}{0.095\xspace}
\newcommand{\DropAjNewLo}{0.064\xspace}
\newcommand{\DropAjNewHi}{0.125\xspace}
\newcommand{\ShamAjNew}{-9\xspace}
\newcommand{\DropAkNew}{0.099\xspace}
\newcommand{\DropAkNewLo}{0.063\xspace}
\newcommand{\DropAkNewHi}{0.133\xspace}
\newcommand{\ShamAkNew}{1\xspace}
\newcommand{\DropAlNew}{0.066\xspace}
\newcommand{\DropAlNewLo}{0.030\xspace}
\newcommand{\DropAlNewHi}{0.101\xspace}
\newcommand{\ShamAlNew}{-1\xspace}
\newcommand{\DropAmNew}{0.084\xspace}
\newcommand{\DropAmNewLo}{0.052\xspace}
\newcommand{\DropAmNewHi}{0.115\xspace}
\newcommand{\ShamAmNew}{1\xspace}
\newcommand{\DropAnNew}{0.099\xspace}
\newcommand{\DropAnNewLo}{0.064\xspace}
\newcommand{\DropAnNewHi}{0.133\xspace}
\newcommand{\ShamAnNew}{-10\xspace}
\newcommand{\DropAoNew}{0.111\xspace}
\newcommand{\DropAoNewLo}{0.077\xspace}
\newcommand{\DropAoNewHi}{0.145\xspace}
\newcommand{\ShamAoNew}{7\xspace}
\newcommand{\DropApNew}{0.071\xspace}
\newcommand{\DropApNewLo}{0.039\xspace}
\newcommand{\DropApNewHi}{0.101\xspace}
\newcommand{\ShamApNew}{-4\xspace}
\newcommand{\DropAqNew}{0.077\xspace}
\newcommand{\DropAqNewLo}{0.040\xspace}
\newcommand{\DropAqNewHi}{0.112\xspace}
\newcommand{\ShamAqNew}{11\xspace}
\newcommand{\DropArNew}{0.082\xspace}
\newcommand{\DropArNewLo}{0.048\xspace}
\newcommand{\DropArNewHi}{0.115\xspace}
\newcommand{\ShamArNew}{-3\xspace}
\newcommand{\DropAsNew}{0.054\xspace}
\newcommand{\DropAsNewLo}{0.019\xspace}
\newcommand{\DropAsNewHi}{0.091\xspace}
\newcommand{\ShamAsNew}{16\xspace}
\newcommand{\DropAtNew}{0.070\xspace}
\newcommand{\DropAtNewLo}{0.033\xspace}
\newcommand{\DropAtNewHi}{0.105\xspace}
\newcommand{\ShamAtNew}{5\xspace}
\newcommand{\DropAuNew}{0.084\xspace}
\newcommand{\DropAuNewLo}{0.048\xspace}
\newcommand{\DropAuNewHi}{0.118\xspace}
\newcommand{\ShamAuNew}{17\xspace}
\newcommand{\DropAvNew}{0.060\xspace}
\newcommand{\DropAvNewLo}{0.022\xspace}
\newcommand{\DropAvNewHi}{0.093\xspace}
\newcommand{\ShamAvNew}{15\xspace}
\newcommand{\DropAwNew}{0.047\xspace}
\newcommand{\DropAwNewLo}{0.012\xspace}
\newcommand{\DropAwNewHi}{0.081\xspace}
\newcommand{\ShamAwNew}{14\xspace}
\newcommand{\DropAxNew}{0.079\xspace}
\newcommand{\DropAxNewLo}{0.044\xspace}
\newcommand{\DropAxNewHi}{0.113\xspace}
\newcommand{\ShamAxNew}{4\xspace}
\newcommand{\DropAyNew}{0.056\xspace}
\newcommand{\DropAyNewLo}{0.020\xspace}
\newcommand{\DropAyNewHi}{0.093\xspace}
\newcommand{\ShamAyNew}{7\xspace}
\newcommand{\DropAzNew}{0.070\xspace}
\newcommand{\DropAzNewLo}{0.035\xspace}
\newcommand{\DropAzNewHi}{0.104\xspace}
\newcommand{\ShamAzNew}{5\xspace}
\newcommand{\DropBaNew}{0.036\xspace}
\newcommand{\DropBaNewLo}{-0.001\xspace}
\newcommand{\DropBaNewHi}{0.070\xspace}
\newcommand{\ShamBaNew}{3\xspace}
\newcommand{\DropBbNew}{0.062\xspace}
\newcommand{\DropBbNewLo}{0.028\xspace}
\newcommand{\DropBbNewHi}{0.096\xspace}
\newcommand{\ShamBbNew}{12\xspace}
\newcommand{\DropBcNew}{0.044\xspace}
\newcommand{\DropBcNewLo}{0.008\xspace}
\newcommand{\DropBcNewHi}{0.080\xspace}
\newcommand{\ShamBcNew}{7\xspace}
\newcommand{\DropBdNew}{0.071\xspace}
\newcommand{\DropBdNewLo}{0.035\xspace}
\newcommand{\DropBdNewHi}{0.108\xspace}
\newcommand{\ShamBdNew}{1\xspace}
\newcommand{\DropBeNew}{0.018\xspace}
\newcommand{\DropBeNewLo}{-0.017\xspace}
\newcommand{\DropBeNewHi}{0.053\xspace}
\newcommand{\ShamBeNew}{6\xspace}
\newcommand{\DropBfNew}{0.077\xspace}
\newcommand{\DropBfNewLo}{0.040\xspace}
\newcommand{\DropBfNewHi}{0.114\xspace}
\newcommand{\ShamBfNew}{-4\xspace}
\newcommand{\DropBgNew}{0.032\xspace}
\newcommand{\DropBgNewLo}{-0.003\xspace}
\newcommand{\DropBgNewHi}{0.066\xspace}
\newcommand{\ShamBgNew}{2\xspace}
\newcommand{\DropBhNew}{0.045\xspace}
\newcommand{\DropBhNewLo}{0.010\xspace}
\newcommand{\DropBhNewHi}{0.082\xspace}
\newcommand{\ShamBhNew}{1\xspace}
\newcommand{\DropBiNew}{0.028\xspace}
\newcommand{\DropBiNewLo}{-0.004\xspace}
\newcommand{\DropBiNewHi}{0.061\xspace}
\newcommand{\ShamBiNew}{4\xspace}
\newcommand{\DropBjNew}{0.046\xspace}
\newcommand{\DropBjNewLo}{0.011\xspace}
\newcommand{\DropBjNewHi}{0.081\xspace}
\newcommand{\ShamBjNew}{-1\xspace}
\newcommand{\DropBkNew}{0.028\xspace}
\newcommand{\DropBkNewLo}{-0.004\xspace}
\newcommand{\DropBkNewHi}{0.064\xspace}
\newcommand{\ShamBkNew}{10\xspace}
\newcommand{\DropBlNew}{0.004\xspace}
\newcommand{\DropBlNewLo}{-0.030\xspace}
\newcommand{\DropBlNewHi}{0.037\xspace}
\newcommand{\ShamBlNew}{0\xspace}
\newcommand{\DropBmNew}{0.054\xspace}
\newcommand{\DropBmNewLo}{0.020\xspace}
\newcommand{\DropBmNewHi}{0.090\xspace}
\newcommand{\ShamBmNew}{0\xspace}
\newcommand{\DropBnNew}{0.024\xspace}
\newcommand{\DropBnNewLo}{-0.009\xspace}
\newcommand{\DropBnNewHi}{0.061\xspace}
\newcommand{\ShamBnNew}{2\xspace}
\newcommand{\SecondsFirst}{1,254\xspace}
\newcommand{\SecondsNew}{1,208\xspace}
\newcommand{\VerTorch}{2.13.0\xspace}
\newcommand{\VerTorchvision}{0.28.0\xspace}
\newcommand{\VerTimm}{1.0.28\xspace}
\newcommand{\VerNumpy}{2.1.3\xspace}
\newcommand{\VerPillow}{11.1.0\xspace}

\newcommand{\VerPython}{3.11.12\xspace}
\newcommand{\DeviceSxm}{NVIDIA A100-SXM4-80GB\xspace}
\newcommand{\DevicePcie}{NVIDIA A100 80GB PCIe\xspace}
\newcommand{\ScoreBatch}{64\xspace}
\newcommand{\BelowZeroFirst}{3\xspace}

\newcommand{\PointBeyondFirst}{3\xspace}

\newcommand{\WorstFirst}{-0.061\xspace}
\newcommand{\WorstUpperFirst}{-0.014\xspace}

\newcommand{\PointBeyondNew}{2\xspace}
\newcommand{\OrderCorrNew}{0.95\xspace}

\newcommand{\AbsChangeNew}{0.0112\xspace}

\newcommand{\CohortFirst}{0.068\xspace}
\newcommand{\CohortNew}{0.066\xspace}
\newcommand{\CohortDiffLo}{-0.031\xspace}
\newcommand{\CohortDiffHi}{0.032\xspace}

\newcommand{\MovedAFirst}{0.017\xspace}
\newcommand{\MovedANew}{0.046\xspace}

\newcommand{\MovedBFirst}{0.040\xspace}
\newcommand{\MovedBNew}{0.018\xspace}
\newcommand{\SplitCount}{200\xspace}
\newcommand{\SplitDistinct}{200\xspace}

\newcommand{\SplitSame}{40\xspace}

\newcommand{\SplitDropMin}{0.058\xspace}

\newcommand{\SplitDropMax}{0.076\xspace}
\newcommand{\SplitBoundMin}{0.035\xspace}
\newcommand{\SplitCellAa}{69\xspace}
\newcommand{\SplitCellAb}{81\xspace}
\newcommand{\SplitCellAc}{4\xspace}
\newcommand{\SplitCellAd}{6\xspace}
\newcommand{\SplitCellAe}{0\xspace}
\newcommand{\SplitCellAf}{0\xspace}
\newcommand{\SplitCellAg}{0\xspace}
\newcommand{\SplitCellAh}{40\xspace}
\newcommand{\SplitCellBa}{9\xspace}
\newcommand{\SplitCellBb}{0\xspace}
\newcommand{\SplitCellBc}{189\xspace}
\newcommand{\SplitCellBd}{1\xspace}
\newcommand{\SplitCellBe}{0\xspace}
\newcommand{\SplitCellBf}{0\xspace}
\newcommand{\SplitCellBg}{0\xspace}
\newcommand{\SplitCellBh}{1\xspace}
\newcommand{\SplitCellCa}{84\xspace}
\newcommand{\SplitCellCb}{0\xspace}
\newcommand{\SplitCellCc}{0\xspace}
\newcommand{\SplitCellCd}{0\xspace}
\newcommand{\SplitCellCe}{0\xspace}
\newcommand{\SplitCellCf}{0\xspace}
\newcommand{\SplitCellCg}{0\xspace}
\newcommand{\SplitCellCh}{116\xspace}
\newcommand{\SplitCellDa}{0\xspace}
\newcommand{\SplitCellDb}{0\xspace}
\newcommand{\SplitCellDc}{0\xspace}
\newcommand{\SplitCellDd}{0\xspace}
\newcommand{\SplitCellDe}{0\xspace}
\newcommand{\SplitCellDf}{200\xspace}
\newcommand{\SplitCellDg}{0\xspace}
\newcommand{\SplitCellDh}{0\xspace}
\newcommand{\SplitCellEa}{0\xspace}
\newcommand{\SplitCellEb}{0\xspace}
\newcommand{\SplitCellEc}{0\xspace}
\newcommand{\SplitCellEd}{193\xspace}
\newcommand{\SplitCellEe}{7\xspace}
\newcommand{\SplitCellEf}{0\xspace}
\newcommand{\SplitCellEg}{0\xspace}
\newcommand{\SplitCellEh}{0\xspace}
\newcommand{\SplitCellFa}{0\xspace}
\newcommand{\SplitCellFb}{0\xspace}
\newcommand{\SplitCellFc}{0\xspace}
\newcommand{\SplitCellFd}{0\xspace}
\newcommand{\SplitCellFe}{0\xspace}
\newcommand{\SplitCellFf}{0\xspace}
\newcommand{\SplitCellFg}{200\xspace}
\newcommand{\SplitCellFh}{0\xspace}
\newcommand{\SplitCellGa}{0\xspace}
\newcommand{\SplitCellGb}{0\xspace}
\newcommand{\SplitCellGc}{7\xspace}
\newcommand{\SplitCellGd}{0\xspace}
\newcommand{\SplitCellGe}{193\xspace}
\newcommand{\SplitCellGf}{0\xspace}
\newcommand{\SplitCellGg}{0\xspace}
\newcommand{\SplitCellGh}{0\xspace}
\newcommand{\SplitCellHa}{38\xspace}
\newcommand{\SplitCellHb}{119\xspace}
\newcommand{\SplitCellHc}{0\xspace}
\newcommand{\SplitCellHd}{0\xspace}
\newcommand{\SplitCellHe}{0\xspace}
\newcommand{\SplitCellHf}{0\xspace}
\newcommand{\SplitCellHg}{0\xspace}
\newcommand{\SplitCellHh}{43\xspace}
\newcommand{\SplitFirstMin}{0.062\xspace}
\newcommand{\SplitFirstMax}{0.079\xspace}
\newcommand{\TransRwLoFirst}{3\xspace}
\newcommand{\TransRwHiFirst}{15\xspace}
\newcommand{\TransWrLoFirst}{2\xspace}
\newcommand{\TransWrHiFirst}{22\xspace}

\newcommand{\TransRwLoNew}{3\xspace}
\newcommand{\TransRwHiNew}{11\xspace}
\newcommand{\TransWrLoNew}{3\xspace}
\newcommand{\TransWrHiNew}{16\xspace}

\newcommand{\RwAaFirst}{14\xspace}
\newcommand{\WrAaFirst}{4\xspace}
\newcommand{\ChgAaFirst}{29\xspace}
\newcommand{\RwAbFirst}{13\xspace}
\newcommand{\WrAbFirst}{5\xspace}
\newcommand{\ChgAbFirst}{33\xspace}
\newcommand{\RwAcFirst}{8\xspace}
\newcommand{\WrAcFirst}{13\xspace}
\newcommand{\ChgAcFirst}{33\xspace}
\newcommand{\RwAdFirst}{13\xspace}
\newcommand{\WrAdFirst}{6\xspace}
\newcommand{\ChgAdFirst}{21\xspace}
\newcommand{\RwAeFirst}{3\xspace}
\newcommand{\WrAeFirst}{8\xspace}
\newcommand{\ChgAeFirst}{21\xspace}
\newcommand{\RwAfFirst}{10\xspace}
\newcommand{\WrAfFirst}{10\xspace}
\newcommand{\ChgAfFirst}{30\xspace}
\newcommand{\RwAgFirst}{14\xspace}
\newcommand{\WrAgFirst}{8\xspace}
\newcommand{\ChgAgFirst}{30\xspace}
\newcommand{\RwAhFirst}{4\xspace}
\newcommand{\WrAhFirst}{10\xspace}
\newcommand{\ChgAhFirst}{21\xspace}
\newcommand{\RwAiFirst}{5\xspace}
\newcommand{\WrAiFirst}{5\xspace}
\newcommand{\ChgAiFirst}{15\xspace}
\newcommand{\RwAjFirst}{11\xspace}
\newcommand{\WrAjFirst}{14\xspace}
\newcommand{\ChgAjFirst}{34\xspace}
\newcommand{\RwAkFirst}{7\xspace}
\newcommand{\WrAkFirst}{8\xspace}
\newcommand{\ChgAkFirst}{25\xspace}
\newcommand{\RwAlFirst}{11\xspace}
\newcommand{\WrAlFirst}{12\xspace}
\newcommand{\ChgAlFirst}{30\xspace}
\newcommand{\RwAmFirst}{3\xspace}
\newcommand{\WrAmFirst}{7\xspace}
\newcommand{\ChgAmFirst}{13\xspace}
\newcommand{\RwAnFirst}{8\xspace}
\newcommand{\WrAnFirst}{9\xspace}
\newcommand{\ChgAnFirst}{26\xspace}
\newcommand{\RwAoFirst}{8\xspace}
\newcommand{\WrAoFirst}{9\xspace}
\newcommand{\ChgAoFirst}{20\xspace}
\newcommand{\RwApFirst}{10\xspace}
\newcommand{\WrApFirst}{6\xspace}
\newcommand{\ChgApFirst}{22\xspace}
\newcommand{\RwAqFirst}{5\xspace}
\newcommand{\WrAqFirst}{9\xspace}
\newcommand{\ChgAqFirst}{19\xspace}
\newcommand{\RwArFirst}{11\xspace}
\newcommand{\WrArFirst}{8\xspace}
\newcommand{\ChgArFirst}{25\xspace}
\newcommand{\RwAsFirst}{8\xspace}
\newcommand{\WrAsFirst}{4\xspace}
\newcommand{\ChgAsFirst}{15\xspace}
\newcommand{\RwAtFirst}{5\xspace}
\newcommand{\WrAtFirst}{9\xspace}
\newcommand{\ChgAtFirst}{21\xspace}
\newcommand{\RwAuFirst}{15\xspace}
\newcommand{\WrAuFirst}{22\xspace}
\newcommand{\ChgAuFirst}{53\xspace}
\newcommand{\RwAvFirst}{4\xspace}
\newcommand{\WrAvFirst}{13\xspace}
\newcommand{\ChgAvFirst}{24\xspace}
\newcommand{\RwAwFirst}{9\xspace}
\newcommand{\WrAwFirst}{19\xspace}
\newcommand{\ChgAwFirst}{40\xspace}
\newcommand{\RwAxFirst}{9\xspace}
\newcommand{\WrAxFirst}{10\xspace}
\newcommand{\ChgAxFirst}{21\xspace}
\newcommand{\RwAyFirst}{7\xspace}
\newcommand{\WrAyFirst}{10\xspace}
\newcommand{\ChgAyFirst}{23\xspace}
\newcommand{\RwAzFirst}{8\xspace}
\newcommand{\WrAzFirst}{15\xspace}
\newcommand{\ChgAzFirst}{28\xspace}
\newcommand{\RwBaFirst}{2\xspace}
\newcommand{\WrBaFirst}{8\xspace}
\newcommand{\ChgBaFirst}{16\xspace}
\newcommand{\RwBbFirst}{10\xspace}
\newcommand{\WrBbFirst}{13\xspace}
\newcommand{\ChgBbFirst}{34\xspace}
\newcommand{\RwBcFirst}{8\xspace}
\newcommand{\WrBcFirst}{3\xspace}
\newcommand{\ChgBcFirst}{14\xspace}
\newcommand{\RwBdFirst}{7\xspace}
\newcommand{\WrBdFirst}{10\xspace}
\newcommand{\ChgBdFirst}{26\xspace}
\newcommand{\RwBeFirst}{8\xspace}
\newcommand{\WrBeFirst}{7\xspace}
\newcommand{\ChgBeFirst}{19\xspace}
\newcommand{\RwBfFirst}{6\xspace}
\newcommand{\WrBfFirst}{6\xspace}
\newcommand{\ChgBfFirst}{15\xspace}
\newcommand{\RwBgFirst}{9\xspace}
\newcommand{\WrBgFirst}{7\xspace}
\newcommand{\ChgBgFirst}{27\xspace}
\newcommand{\RwBhFirst}{4\xspace}
\newcommand{\WrBhFirst}{2\xspace}
\newcommand{\ChgBhFirst}{9\xspace}
\newcommand{\RwBiFirst}{5\xspace}
\newcommand{\WrBiFirst}{3\xspace}
\newcommand{\ChgBiFirst}{10\xspace}
\newcommand{\RwBjFirst}{3\xspace}
\newcommand{\WrBjFirst}{5\xspace}
\newcommand{\ChgBjFirst}{10\xspace}
\newcommand{\RwBkFirst}{6\xspace}
\newcommand{\WrBkFirst}{19\xspace}
\newcommand{\ChgBkFirst}{29\xspace}
\newcommand{\RwBlFirst}{1\xspace}
\newcommand{\WrBlFirst}{5\xspace}
\newcommand{\ChgBlFirst}{6\xspace}
\newcommand{\RwBmFirst}{7\xspace}
\newcommand{\WrBmFirst}{18\xspace}
\newcommand{\ChgBmFirst}{30\xspace}
\newcommand{\RwBnFirst}{4\xspace}
\newcommand{\WrBnFirst}{6\xspace}
\newcommand{\ChgBnFirst}{11\xspace}
\newcommand{\RankGridAb}{14\xspace}
\newcommand{\RankHardAb}{14\xspace}
\newcommand{\RankGridAd}{8\xspace}
\newcommand{\RankHardAd}{10\xspace}
\newcommand{\RankGridAe}{13\xspace}
\newcommand{\RankHardAe}{13\xspace}
\newcommand{\RankGridAi}{12\xspace}
\newcommand{\RankHardAi}{12\xspace}
\newcommand{\RankGridAk}{10\xspace}
\newcommand{\RankHardAk}{11\xspace}
\newcommand{\RankGridAl}{11\xspace}
\newcommand{\RankHardAl}{9\xspace}
\newcommand{\RankGridAz}{9\xspace}
\newcommand{\RankHardAz}{8\xspace}
\newcommand{\RankGridBe}{6\xspace}
\newcommand{\RankHardBe}{4\xspace}
\newcommand{\RankGridBf}{4\xspace}
\newcommand{\RankHardBf}{6\xspace}
\newcommand{\RankGridBg}{7\xspace}
\newcommand{\RankHardBg}{7\xspace}
\newcommand{\RankGridBh}{1\xspace}
\newcommand{\RankHardBh}{2\xspace}
\newcommand{\RankGridBj}{5\xspace}
\newcommand{\RankHardBj}{5\xspace}
\newcommand{\RankGridBm}{2\xspace}
\newcommand{\RankHardBm}{3\xspace}
\newcommand{\RankGridBn}{3\xspace}
\newcommand{\RankHardBn}{1\xspace}
\newcommand{\RwAaNew}{10\xspace}
\newcommand{\WrAaNew}{7\xspace}
\newcommand{\ChgAaNew}{29\xspace}
\newcommand{\RwAbNew}{8\xspace}
\newcommand{\WrAbNew}{7\xspace}
\newcommand{\ChgAbNew}{26\xspace}
\newcommand{\RwAcNew}{8\xspace}
\newcommand{\WrAcNew}{10\xspace}
\newcommand{\ChgAcNew}{28\xspace}
\newcommand{\RwAdNew}{6\xspace}
\newcommand{\WrAdNew}{8\xspace}
\newcommand{\ChgAdNew}{21\xspace}
\newcommand{\RwAeNew}{10\xspace}
\newcommand{\WrAeNew}{4\xspace}
\newcommand{\ChgAeNew}{23\xspace}
\newcommand{\RwAfNew}{9\xspace}
\newcommand{\WrAfNew}{6\xspace}
\newcommand{\ChgAfNew}{25\xspace}
\newcommand{\RwAgNew}{5\xspace}
\newcommand{\WrAgNew}{10\xspace}
\newcommand{\ChgAgNew}{25\xspace}
\newcommand{\RwAhNew}{5\xspace}
\newcommand{\WrAhNew}{5\xspace}
\newcommand{\ChgAhNew}{23\xspace}
\newcommand{\RwAiNew}{9\xspace}
\newcommand{\WrAiNew}{7\xspace}
\newcommand{\ChgAiNew}{22\xspace}
\newcommand{\RwAjNew}{14\xspace}
\newcommand{\WrAjNew}{5\xspace}
\newcommand{\ChgAjNew}{32\xspace}
\newcommand{\RwAkNew}{10\xspace}
\newcommand{\WrAkNew}{11\xspace}
\newcommand{\ChgAkNew}{25\xspace}
\newcommand{\RwAlNew}{5\xspace}
\newcommand{\WrAlNew}{4\xspace}
\newcommand{\ChgAlNew}{16\xspace}
\newcommand{\RwAmNew}{5\xspace}
\newcommand{\WrAmNew}{6\xspace}
\newcommand{\ChgAmNew}{15\xspace}
\newcommand{\RwAnNew}{17\xspace}
\newcommand{\WrAnNew}{7\xspace}
\newcommand{\ChgAnNew}{34\xspace}
\newcommand{\RwAoNew}{3\xspace}
\newcommand{\WrAoNew}{10\xspace}
\newcommand{\ChgAoNew}{19\xspace}
\newcommand{\RwApNew}{10\xspace}
\newcommand{\WrApNew}{6\xspace}
\newcommand{\ChgApNew}{23\xspace}
\newcommand{\RwAqNew}{5\xspace}
\newcommand{\WrAqNew}{16\xspace}
\newcommand{\ChgAqNew}{30\xspace}
\newcommand{\RwArNew}{13\xspace}
\newcommand{\WrArNew}{10\xspace}
\newcommand{\ChgArNew}{35\xspace}
\newcommand{\RwAsNew}{5\xspace}
\newcommand{\WrAsNew}{21\xspace}
\newcommand{\ChgAsNew}{40\xspace}
\newcommand{\RwAtNew}{2\xspace}
\newcommand{\WrAtNew}{7\xspace}
\newcommand{\ChgAtNew}{12\xspace}
\newcommand{\RwAuNew}{9\xspace}
\newcommand{\WrAuNew}{26\xspace}
\newcommand{\ChgAuNew}{43\xspace}
\newcommand{\RwAvNew}{6\xspace}
\newcommand{\WrAvNew}{21\xspace}
\newcommand{\ChgAvNew}{38\xspace}
\newcommand{\RwAwNew}{4\xspace}
\newcommand{\WrAwNew}{18\xspace}
\newcommand{\ChgAwNew}{33\xspace}
\newcommand{\RwAxNew}{6\xspace}
\newcommand{\WrAxNew}{10\xspace}
\newcommand{\ChgAxNew}{21\xspace}
\newcommand{\RwAyNew}{5\xspace}
\newcommand{\WrAyNew}{12\xspace}
\newcommand{\ChgAyNew}{21\xspace}
\newcommand{\RwAzNew}{11\xspace}
\newcommand{\WrAzNew}{16\xspace}
\newcommand{\ChgAzNew}{36\xspace}
\newcommand{\RwBaNew}{11\xspace}
\newcommand{\WrBaNew}{14\xspace}
\newcommand{\ChgBaNew}{29\xspace}
\newcommand{\RwBbNew}{3\xspace}
\newcommand{\WrBbNew}{15\xspace}
\newcommand{\ChgBbNew}{25\xspace}
\newcommand{\RwBcNew}{4\xspace}
\newcommand{\WrBcNew}{11\xspace}
\newcommand{\ChgBcNew}{16\xspace}
\newcommand{\RwBdNew}{6\xspace}
\newcommand{\WrBdNew}{7\xspace}
\newcommand{\ChgBdNew}{18\xspace}
\newcommand{\RwBeNew}{7\xspace}
\newcommand{\WrBeNew}{13\xspace}
\newcommand{\ChgBeNew}{23\xspace}
\newcommand{\RwBfNew}{7\xspace}
\newcommand{\WrBfNew}{3\xspace}
\newcommand{\ChgBfNew}{13\xspace}
\newcommand{\RwBgNew}{8\xspace}
\newcommand{\WrBgNew}{10\xspace}
\newcommand{\ChgBgNew}{23\xspace}
\newcommand{\RwBhNew}{3\xspace}
\newcommand{\WrBhNew}{4\xspace}
\newcommand{\ChgBhNew}{9\xspace}
\newcommand{\RwBiNew}{2\xspace}
\newcommand{\WrBiNew}{6\xspace}
\newcommand{\ChgBiNew}{11\xspace}
\newcommand{\RwBjNew}{4\xspace}
\newcommand{\WrBjNew}{3\xspace}
\newcommand{\ChgBjNew}{7\xspace}
\newcommand{\RwBkNew}{7\xspace}
\newcommand{\WrBkNew}{17\xspace}
\newcommand{\ChgBkNew}{29\xspace}
\newcommand{\RwBlNew}{3\xspace}
\newcommand{\WrBlNew}{3\xspace}
\newcommand{\ChgBlNew}{7\xspace}
\newcommand{\RwBmNew}{9\xspace}
\newcommand{\WrBmNew}{9\xspace}
\newcommand{\ChgBmNew}{20\xspace}
\newcommand{\RwBnNew}{4\xspace}
\newcommand{\WrBnNew}{6\xspace}
\newcommand{\ChgBnNew}{13\xspace}

\newcommand{\GridAbNew}{0.312\xspace}
\newcommand{\HardAbNew}{0.232\xspace}

\newcommand{\GridAdNew}{0.496\xspace}
\newcommand{\HardAdNew}{0.385\xspace}
\newcommand{\GridAeNew}{0.405\xspace}
\newcommand{\HardAeNew}{0.314\xspace}

\newcommand{\GridAiNew}{0.455\xspace}
\newcommand{\HardAiNew}{0.370\xspace}

\newcommand{\GridAkNew}{0.472\xspace}
\newcommand{\HardAkNew}{0.379\xspace}
\newcommand{\GridAlNew}{0.469\xspace}
\newcommand{\HardAlNew}{0.406\xspace}

\newcommand{\GridAzNew}{0.486\xspace}
\newcommand{\HardAzNew}{0.419\xspace}

\newcommand{\GridBeNew}{0.529\xspace}
\newcommand{\HardBeNew}{0.515\xspace}
\newcommand{\GridBfNew}{0.550\xspace}
\newcommand{\HardBfNew}{0.477\xspace}
\newcommand{\GridBgNew}{0.499\xspace}
\newcommand{\HardBgNew}{0.470\xspace}
\newcommand{\GridBhNew}{0.594\xspace}
\newcommand{\HardBhNew}{0.551\xspace}

\newcommand{\GridBjNew}{0.538\xspace}
\newcommand{\HardBjNew}{0.494\xspace}

\newcommand{\GridBmNew}{0.591\xspace}
\newcommand{\HardBmNew}{0.541\xspace}
\newcommand{\GridBnNew}{0.590\xspace}
\newcommand{\HardBnNew}{0.567\xspace}
\newcommand{\BootReps}{2,000\xspace}
\newcommand{\CueMargin}{0.01\xspace}
\newcommand{\VerScipy}{1.18.0\xspace}
\newcommand{\VerNumpyHere}{2.1.3\xspace}
\newcommand{\ExcludedEarlier}{204\xspace}
\newcommand{\ExcludedFirstSet}{1,600\xspace}
\newcommand{\SeedUnitsSecond}{20261030\xspace}
\newcommand{\SeedUnitsAdded}{20261040\xspace}
\newcommand{\SeedCells}{20261041\xspace}
\newcommand{\SeedFamilies}{20261032\xspace}
\newcommand{\SeedBins}{20261033\xspace}
\newcommand{\SeedSplit}{20261042\xspace}
\newcommand{\SeedEffect}{20261043\xspace}
\newcommand{\SeedUnit}{20261044\xspace}
\newcommand{\SeedModels}{20261045\xspace}
\newcommand{\SeedRandom}{20261046\xspace}
\newcommand{\SeedNewUnits}{20261050\xspace}
\newcommand{\SeedPaired}{20261052\xspace}
\newcommand{\SeedPairing}{20261101\xspace}
\newcommand{\SeedSplits}{20261102\xspace}
\newcommand{\RelLowAaFirst}{-0.060\xspace}
\newcommand{\RelLowAbFirst}{-0.070\xspace}
\newcommand{\RelLowAcFirst}{-0.150\xspace}
\newcommand{\RelLowAdFirst}{-0.137\xspace}
\newcommand{\RelLowAeFirst}{-0.063\xspace}
\newcommand{\RelLowAfFirst}{-0.037\xspace}
\newcommand{\RelLowAgFirst}{-0.082\xspace}
\newcommand{\RelLowAhFirst}{-0.063\xspace}
\newcommand{\RelLowBaFirst}{-0.050\xspace}
\newcommand{\RelLowBbFirst}{-0.016\xspace}
\newcommand{\RelLowBcFirst}{-0.144\xspace}
\newcommand{\RelLowBdFirst}{-0.116\xspace}
\newcommand{\RelLowBeFirst}{-0.026\xspace}
\newcommand{\RelLowBfFirst}{-0.078\xspace}
\newcommand{\RelLowBgFirst}{-0.121\xspace}
\newcommand{\RelLowBhFirst}{-0.042\xspace}
\newcommand{\RelLowCaFirst}{-0.009\xspace}
\newcommand{\RelLowCbFirst}{-0.004\xspace}
\newcommand{\RelLowCcFirst}{-0.034\xspace}
\newcommand{\RelLowCdFirst}{0.002\xspace}
\newcommand{\RelLowCeFirst}{0.003\xspace}
\newcommand{\RelLowCfFirst}{0.004\xspace}
\newcommand{\RelLowCgFirst}{-0.036\xspace}
\newcommand{\RelLowChFirst}{0.001\xspace}
\newcommand{\RelLowDaFirst}{0.015\xspace}
\newcommand{\RelLowDbFirst}{0.002\xspace}
\newcommand{\RelLowDcFirst}{-0.123\xspace}
\newcommand{\RelLowDdFirst}{-0.168\xspace}
\newcommand{\RelLowDeFirst}{-0.089\xspace}
\newcommand{\RelLowDfFirst}{-0.190\xspace}
\newcommand{\RelLowDgFirst}{-0.048\xspace}
\newcommand{\RelLowDhFirst}{0.005\xspace}
\newcommand{\RelLowEaFirst}{0.007\xspace}
\newcommand{\RelLowEbFirst}{-0.078\xspace}
\newcommand{\RelLowEcFirst}{0.049\xspace}
\newcommand{\RelLowEdFirst}{-0.170\xspace}
\newcommand{\RelLowEeFirst}{-0.141\xspace}
\newcommand{\RelLowEfFirst}{-0.173\xspace}
\newcommand{\RelLowEgFirst}{-0.069\xspace}
\newcommand{\RelLowEhFirst}{-0.011\xspace}
\newcommand{\RelLowFaFirst}{0.173\xspace}
\newcommand{\RelLowFbFirst}{-0.172\xspace}
\newcommand{\RelLowFcFirst}{-0.050\xspace}
\newcommand{\RelLowFdFirst}{-0.215\xspace}
\newcommand{\RelLowFeFirst}{-0.093\xspace}
\newcommand{\RelLowFfFirst}{-0.140\xspace}
\newcommand{\RelLowFgFirst}{-0.280\xspace}
\newcommand{\RelLowFhFirst}{0.187\xspace}
\newcommand{\RelLowGaFirst}{-0.009\xspace}
\newcommand{\RelLowGbFirst}{0.005\xspace}
\newcommand{\RelLowGcFirst}{-0.110\xspace}
\newcommand{\RelLowGdFirst}{-0.065\xspace}
\newcommand{\RelLowGeFirst}{-0.078\xspace}
\newcommand{\RelLowGfFirst}{-0.056\xspace}
\newcommand{\RelLowGgFirst}{-0.103\xspace}
\newcommand{\RelLowGhFirst}{-0.002\xspace}
\newcommand{\RelLowHaFirst}{-0.021\xspace}
\newcommand{\RelLowHbFirst}{-0.032\xspace}
\newcommand{\RelLowHcFirst}{0.002\xspace}
\newcommand{\RelLowHdFirst}{-0.033\xspace}
\newcommand{\RelLowHeFirst}{-0.029\xspace}
\newcommand{\RelLowHfFirst}{-0.034\xspace}
\newcommand{\RelLowHgFirst}{-0.001\xspace}
\newcommand{\RelLowHhFirst}{-0.020\xspace}
\newcommand{\RelLowAaNew}{-0.078\xspace}
\newcommand{\RelLowAbNew}{-0.069\xspace}
\newcommand{\RelLowAcNew}{-0.120\xspace}
\newcommand{\RelLowAdNew}{-0.123\xspace}
\newcommand{\RelLowAeNew}{-0.075\xspace}
\newcommand{\RelLowAfNew}{-0.027\xspace}
\newcommand{\RelLowAgNew}{-0.106\xspace}
\newcommand{\RelLowAhNew}{-0.050\xspace}
\newcommand{\RelLowBaNew}{-0.056\xspace}
\newcommand{\RelLowBbNew}{-0.030\xspace}
\newcommand{\RelLowBcNew}{-0.128\xspace}
\newcommand{\RelLowBdNew}{-0.131\xspace}
\newcommand{\RelLowBeNew}{-0.011\xspace}
\newcommand{\RelLowBfNew}{-0.062\xspace}
\newcommand{\RelLowBgNew}{-0.108\xspace}
\newcommand{\RelLowBhNew}{-0.043\xspace}
\newcommand{\RelLowCaNew}{-0.032\xspace}
\newcommand{\RelLowCbNew}{-0.023\xspace}
\newcommand{\RelLowCcNew}{-0.036\xspace}
\newcommand{\RelLowCdNew}{0.005\xspace}
\newcommand{\RelLowCeNew}{-0.001\xspace}
\newcommand{\RelLowCfNew}{0.006\xspace}
\newcommand{\RelLowCgNew}{-0.033\xspace}
\newcommand{\RelLowChNew}{-0.014\xspace}
\newcommand{\RelLowDaNew}{0.003\xspace}
\newcommand{\RelLowDbNew}{-0.019\xspace}
\newcommand{\RelLowDcNew}{-0.086\xspace}
\newcommand{\RelLowDdNew}{-0.212\xspace}
\newcommand{\RelLowDeNew}{-0.072\xspace}
\newcommand{\RelLowDfNew}{-0.185\xspace}
\newcommand{\RelLowDgNew}{-0.034\xspace}
\newcommand{\RelLowDhNew}{0.027\xspace}
\newcommand{\RelLowEaNew}{0.003\xspace}
\newcommand{\RelLowEbNew}{-0.075\xspace}
\newcommand{\RelLowEcNew}{0.061\xspace}
\newcommand{\RelLowEdNew}{-0.129\xspace}
\newcommand{\RelLowEeNew}{-0.150\xspace}
\newcommand{\RelLowEfNew}{-0.173\xspace}
\newcommand{\RelLowEgNew}{-0.085\xspace}
\newcommand{\RelLowEhNew}{-0.043\xspace}
\newcommand{\RelLowFaNew}{0.185\xspace}
\newcommand{\RelLowFbNew}{-0.181\xspace}
\newcommand{\RelLowFcNew}{-0.037\xspace}
\newcommand{\RelLowFdNew}{-0.233\xspace}
\newcommand{\RelLowFeNew}{-0.062\xspace}
\newcommand{\RelLowFfNew}{-0.119\xspace}
\newcommand{\RelLowFgNew}{-0.302\xspace}
\newcommand{\RelLowFhNew}{0.187\xspace}
\newcommand{\RelLowGaNew}{-0.019\xspace}
\newcommand{\RelLowGbNew}{-0.024\xspace}
\newcommand{\RelLowGcNew}{-0.113\xspace}
\newcommand{\RelLowGdNew}{-0.060\xspace}
\newcommand{\RelLowGeNew}{-0.084\xspace}
\newcommand{\RelLowGfNew}{-0.069\xspace}
\newcommand{\RelLowGgNew}{-0.097\xspace}
\newcommand{\RelLowGhNew}{-0.011\xspace}
\newcommand{\RelLowHaNew}{-0.021\xspace}
\newcommand{\RelLowHbNew}{-0.024\xspace}
\newcommand{\RelLowHcNew}{-0.006\xspace}
\newcommand{\RelLowHdNew}{-0.028\xspace}
\newcommand{\RelLowHeNew}{-0.020\xspace}
\newcommand{\RelLowHfNew}{-0.031\xspace}
\newcommand{\RelLowHgNew}{-0.006\xspace}
\newcommand{\RelLowHhNew}{-0.018\xspace}
\newcommand{\KnownLiftErr}{0.04\xspace}
\newcommand{\KnownReplayDrop}{0.0740\xspace}
\newcommand{\KnownReplayLow}{0.0522\xspace}
\newcommand{\KnownPlantDrop}{0.148\xspace}
\newcommand{\KnownPlantUnder}{0.018\xspace}
\newcommand{\KnownBuiltComposites}{24\xspace}
\newcommand{\KnownBuiltControls}{6\xspace}
\newcommand{\KnownBuiltImages}{3\xspace}
\newcommand{\EdgeRadius}{1\xspace}
\newcommand{\EdgeModels}{14\xspace}
\newcommand{\EdgeOld}{0.066\xspace}
\newcommand{\EdgeTreated}{0.068\xspace}
\newcommand{\EdgeDifference}{0.002\xspace}

\newcommand{\EdgeOldLo}{0.042\xspace}
\newcommand{\EdgeOldHi}{0.088\xspace}
\newcommand{\EdgeTreatedLo}{0.046\xspace}
\newcommand{\EdgeTreatedHi}{0.089\xspace}
\newcommand{\EdgeDifferenceLo}{-0.004\xspace}
\newcommand{\EdgeDifferenceHi}{0.009\xspace}
\newcommand{\EdgeOldAb}{0.085\xspace}
\newcommand{\EdgeNewAb}{0.086\xspace}
\newcommand{\EdgeDeltaAb}{0.001\xspace}
\newcommand{\EdgeRwAb}{9\xspace}
\newcommand{\EdgeWrAb}{4\xspace}
\newcommand{\EdgeNetAb}{-5\xspace}
\newcommand{\EdgeChangedAb}{23\xspace}
\newcommand{\EdgeOldAd}{0.115\xspace}
\newcommand{\EdgeNewAd}{0.118\xspace}
\newcommand{\EdgeDeltaAd}{0.003\xspace}
\newcommand{\EdgeRwAd}{6\xspace}
\newcommand{\EdgeWrAd}{10\xspace}
\newcommand{\EdgeNetAd}{4\xspace}
\newcommand{\EdgeChangedAd}{25\xspace}
\newcommand{\EdgeOldAe}{0.097\xspace}
\newcommand{\EdgeNewAe}{0.087\xspace}
\newcommand{\EdgeDeltaAe}{-0.010\xspace}
\newcommand{\EdgeRwAe}{10\xspace}
\newcommand{\EdgeWrAe}{7\xspace}
\newcommand{\EdgeNetAe}{-3\xspace}
\newcommand{\EdgeChangedAe}{25\xspace}
\newcommand{\EdgeOldAi}{0.090\xspace}
\newcommand{\EdgeNewAi}{0.083\xspace}
\newcommand{\EdgeDeltaAi}{-0.007\xspace}
\newcommand{\EdgeRwAi}{7\xspace}
\newcommand{\EdgeWrAi}{8\xspace}
\newcommand{\EdgeNetAi}{1\xspace}
\newcommand{\EdgeChangedAi}{20\xspace}
\newcommand{\EdgeOldAk}{0.099\xspace}
\newcommand{\EdgeNewAk}{0.108\xspace}
\newcommand{\EdgeDeltaAk}{0.009\xspace}
\newcommand{\EdgeRwAk}{9\xspace}
\newcommand{\EdgeWrAk}{11\xspace}
\newcommand{\EdgeNetAk}{2\xspace}
\newcommand{\EdgeChangedAk}{25\xspace}
\newcommand{\EdgeOldAl}{0.066\xspace}
\newcommand{\EdgeNewAl}{0.070\xspace}
\newcommand{\EdgeDeltaAl}{0.003\xspace}
\newcommand{\EdgeRwAl}{7\xspace}
\newcommand{\EdgeWrAl}{7\xspace}
\newcommand{\EdgeNetAl}{0\xspace}
\newcommand{\EdgeChangedAl}{20\xspace}
\newcommand{\EdgeOldAz}{0.070\xspace}
\newcommand{\EdgeNewAz}{0.072\xspace}
\newcommand{\EdgeDeltaAz}{0.002\xspace}
\newcommand{\EdgeRwAz}{14\xspace}
\newcommand{\EdgeWrAz}{17\xspace}
\newcommand{\EdgeNetAz}{3\xspace}
\newcommand{\EdgeChangedAz}{41\xspace}
\newcommand{\EdgeOldBe}{0.018\xspace}
\newcommand{\EdgeNewBe}{0.027\xspace}
\newcommand{\EdgeDeltaBe}{0.010\xspace}
\newcommand{\EdgeRwBe}{9\xspace}
\newcommand{\EdgeWrBe}{12\xspace}
\newcommand{\EdgeNetBe}{3\xspace}
\newcommand{\EdgeChangedBe}{24\xspace}
\newcommand{\EdgeOldBf}{0.077\xspace}
\newcommand{\EdgeNewBf}{0.079\xspace}
\newcommand{\EdgeDeltaBf}{0.002\xspace}
\newcommand{\EdgeRwBf}{7\xspace}
\newcommand{\EdgeWrBf}{3\xspace}
\newcommand{\EdgeNetBf}{-4\xspace}
\newcommand{\EdgeChangedBf}{13\xspace}
\newcommand{\EdgeOldBg}{0.032\xspace}
\newcommand{\EdgeNewBg}{0.041\xspace}
\newcommand{\EdgeDeltaBg}{0.010\xspace}
\newcommand{\EdgeRwBg}{9\xspace}
\newcommand{\EdgeWrBg}{9\xspace}
\newcommand{\EdgeNetBg}{0\xspace}
\newcommand{\EdgeChangedBg}{22\xspace}
\newcommand{\EdgeOldBh}{0.045\xspace}
\newcommand{\EdgeNewBh}{0.056\xspace}
\newcommand{\EdgeDeltaBh}{0.011\xspace}
\newcommand{\EdgeRwBh}{3\xspace}
\newcommand{\EdgeWrBh}{3\xspace}
\newcommand{\EdgeNetBh}{0\xspace}
\newcommand{\EdgeChangedBh}{8\xspace}
\newcommand{\EdgeOldBj}{0.046\xspace}
\newcommand{\EdgeNewBj}{0.044\xspace}
\newcommand{\EdgeDeltaBj}{-0.002\xspace}
\newcommand{\EdgeRwBj}{5\xspace}
\newcommand{\EdgeWrBj}{3\xspace}
\newcommand{\EdgeNetBj}{-2\xspace}
\newcommand{\EdgeChangedBj}{8\xspace}
\newcommand{\EdgeOldBm}{0.054\xspace}
\newcommand{\EdgeNewBm}{0.056\xspace}
\newcommand{\EdgeDeltaBm}{0.002\xspace}
\newcommand{\EdgeRwBm}{8\xspace}
\newcommand{\EdgeWrBm}{9\xspace}
\newcommand{\EdgeNetBm}{1\xspace}
\newcommand{\EdgeChangedBm}{19\xspace}
\newcommand{\EdgeOldBn}{0.024\xspace}
\newcommand{\EdgeNewBn}{0.023\xspace}
\newcommand{\EdgeDeltaBn}{-0.002\xspace}
\newcommand{\EdgeRwBn}{5\xspace}
\newcommand{\EdgeWrBn}{6\xspace}
\newcommand{\EdgeNetBn}{1\xspace}
\newcommand{\EdgeChangedBn}{13\xspace}
\newcommand{\MatchInside}{171\xspace}
\newcommand{\MatchLabelFirst}{0.066\xspace}
\newcommand{\MatchImageFirst}{0.118\xspace}
\newcommand{\MatchDeltaFirst}{-0.052\xspace}
\newcommand{\MatchLabelFirstLo}{0.045\xspace}
\newcommand{\MatchLabelFirstHi}{0.088\xspace}
\newcommand{\MatchImageFirstLo}{0.096\xspace}
\newcommand{\MatchImageFirstHi}{0.140\xspace}
\newcommand{\MatchDeltaFirstLo}{-0.081\xspace}
\newcommand{\MatchDeltaFirstHi}{-0.021\xspace}
\newcommand{\MatchLabelNew}{0.066\xspace}
\newcommand{\MatchImageNew}{0.028\xspace}
\newcommand{\MatchDeltaNew}{0.037\xspace}
\newcommand{\MatchLabelNewLo}{0.042\xspace}
\newcommand{\MatchLabelNewHi}{0.088\xspace}
\newcommand{\MatchImageNewLo}{0.006\xspace}
\newcommand{\MatchImageNewHi}{0.051\xspace}
\newcommand{\MatchDeltaNewLo}{0.007\xspace}
\newcommand{\MatchDeltaNewHi}{0.068\xspace}
\newcommand{\TrainingAa}{ImageNet 1k\xspace}
\newcommand{\TeacherAa}{Not stated\xspace}
\newcommand{\NormMeanAa}{0.485, 0.456, 0.406\xspace}
\newcommand{\NormStdAa}{0.229, 0.224, 0.225\xspace}

\newcommand{\TeacherAk}{Distillation, teacher not stated\xspace}

\newcommand{\NormMeanAs}{0.5, 0.5, 0.5\xspace}
\newcommand{\NormStdAs}{0.5, 0.5, 0.5\xspace}

\newcommand{\TeacherBc}{CLIP B/16 through a VQ-KD tokenizer\xspace}

\newcommand{\TrainingBf}{ImageNet 12k, ImageNet 1k\xspace}

\newcommand{\TrainingBh}{Mined ImageNet 22k, ImageNet 1k\xspace}
\newcommand{\TeacherBh}{SSLD distillation, teacher not stated\xspace}

\newcommand{\TeacherBi}{Token labelling, teacher not stated\xspace}

\newcommand{\TrainingBj}{LAION 400M, ImageNet 1k\xspace}
\newcommand{\TeacherBj}{Pretrained image and text towers, identities not stated\xspace}
\newcommand{\NormMeanBj}{0.48145466, 0.4578275, 0.40821073\xspace}
\newcommand{\NormStdBj}{0.26862954, 0.26130258, 0.27577711\xspace}
\newcommand{\TrainingBk}{ImageNet 22k, ImageNet 1k\xspace}

\newcommand{\TrainingBl}{ImageNet 22k, 160 GB text unidentified on the card, LAION 400M, COYO 700M, CC15M, COCO, Visual Genome, ImageNet 1k\xspace}

\newcommand{\BinEdgeA}{63.13800\xspace}
\newcommand{\BinEdgeB}{66.35775\xspace}
\newcommand{\BinEdgeC}{69.57750\xspace}
\newcommand{\BinEdgeD}{72.79725\xspace}
\newcommand{\BinEdgeE}{76.01700\xspace}
\newcommand{\BinEdgeF}{79.23675\xspace}
\newcommand{\BinEdgeG}{82.45650\xspace}
\newcommand{\BinEdgeH}{85.67625\xspace}
\newcommand{\BinEdgeI}{88.89600\xspace}
\newcommand{\BinQuotaA}{1\xspace}
\newcommand{\BinQuotaB}{1\xspace}
\newcommand{\BinQuotaC}{0\xspace}
\newcommand{\BinQuotaD}{8\xspace}
\newcommand{\BinQuotaE}{8\xspace}
\newcommand{\BinQuotaF}{8\xspace}
\newcommand{\BinQuotaG}{7\xspace}
\newcommand{\BinQuotaH}{7\xspace}
\newcommand{\KnownLiftTolerance}{0.05\xspace}
\newcommand{\SplitFirstDropA}{0.062\xspace}
\newcommand{\SplitFirstDropB}{0.066\xspace}
\newcommand{\SplitFirstDropC}{0.069\xspace}
\newcommand{\SplitFirstDropD}{0.071\xspace}
\newcommand{\SplitFirstDropE}{0.074\xspace}
\newcommand{\SplitFirstDropF}{0.076\xspace}
\newcommand{\SplitFirstDropG}{0.079\xspace}
\newcommand{\SplitFirstBoundA}{0.038\xspace}
\newcommand{\SplitFirstBoundB}{0.043\xspace}
\newcommand{\SplitFirstBoundC}{0.047\xspace}
\newcommand{\SplitFirstBoundD}{0.049\xspace}
\newcommand{\SplitFirstBoundE}{0.051\xspace}
\newcommand{\SplitFirstBoundF}{0.054\xspace}
\newcommand{\SplitFirstBoundG}{0.057\xspace}
\newcommand{\SplitNewDropA}{0.058\xspace}
\newcommand{\SplitNewDropB}{0.060\xspace}
\newcommand{\SplitNewDropC}{0.064\xspace}
\newcommand{\SplitNewDropD}{0.067\xspace}
\newcommand{\SplitNewDropE}{0.069\xspace}
\newcommand{\SplitNewDropF}{0.073\xspace}
\newcommand{\SplitNewDropG}{0.076\xspace}
\newcommand{\SplitNewBoundA}{0.035\xspace}
\newcommand{\SplitNewBoundB}{0.037\xspace}
\newcommand{\SplitNewBoundC}{0.041\xspace}
\newcommand{\SplitNewBoundD}{0.045\xspace}
\newcommand{\SplitNewBoundE}{0.047\xspace}
\newcommand{\SplitNewBoundF}{0.050\xspace}
\newcommand{\SplitNewBoundG}{0.053\xspace}

\newcommand{\SplitAdmittedMinFirst}{15\xspace}
\newcommand{\SplitAdmittedMaxFirst}{20\xspace}

\newcommand{\SplitAdmittedMinNew}{12\xspace}
\newcommand{\SplitAdmittedMaxNew}{19\xspace}
\newcommand{\CandidatePairings}{40,320\xspace}
\newcommand{\BandConstant}{0.061\xspace}

\newcommand{\BandPointHardest}{0.074\xspace}
\newcommand{\BandOrdinaryHardest}{0.052\xspace}
\newcommand{\BandSimultaneousHardest}{0.013\xspace}
\newcommand{\BandPointEasiest}{-0.081\xspace}
\newcommand{\BandOrdinaryEasiest}{-0.103\xspace}
\newcommand{\BandSimultaneousEasiest}{-0.142\xspace}
\newcommand{\BandPointRandom}{0.051\xspace}
\newcommand{\BandOrdinaryRandom}{0.028\xspace}
\newcommand{\BandSimultaneousRandom}{-0.010\xspace}
\newcommand{\CarriedDifference}{0.026\xspace}
\newcommand{\CarriedDifferenceLo}{0.005\xspace}
\newcommand{\CarriedDifferenceHi}{0.047\xspace}
\newcommand{\CoverageWorlds}{240\xspace}
\newcommand{\CoverageReps}{499\xspace}
\newcommand{\CoverageSeed}{20261102031\xspace}
\newcommand{\CoverageLayoutSeed}{20261102032\xspace}
\newcommand{\CoverageBandCount}{240\xspace}
\newcommand{\CoverageBand}{1.000\xspace}
\newcommand{\CoverageBandLo}{0.984\xspace}
\newcommand{\CoverageBandHi}{1.000\xspace}
\newcommand{\CoverageOrdinaryCount}{41\xspace}
\newcommand{\CoverageOrdinary}{0.171\xspace}
\newcommand{\CoverageOrdinaryLo}{0.128\xspace}
\newcommand{\CoverageOrdinaryHi}{0.224\xspace}
\newcommand{\RasterVersion}{2.0.11\xspace}
\newcommand{\ReplayAbsolute}{$1.0\times 10^{-6}$\xspace}
\newcommand{\ReplayRelative}{$1.0\times 10^{-5}$\xspace}
\newcommand{\ReplayError}{0.000002056\xspace}
\newcommand{\SyntheticCenter}{0.55\xspace}
\newcommand{\SyntheticSlope}{0.01\xspace}
\newcommand{\SyntheticMidpoint}{3.5\xspace}
\newcommand{\SyntheticPenalty}{0.03\xspace}
\newcommand{\SyntheticSource}{0.12\xspace}
\newcommand{\SyntheticOffset}{0.08\xspace}
\newcommand{\BandLevel}{0.95\xspace}
\newcommand{\ReplayStride}{64\xspace}
\newcommand{\ReplayLast}{99\xspace}
\newcommand{\KnownCoordinateLo}{0.63\xspace}
\newcommand{\KnownCoordinateHi}{0.89\xspace}
\newcommand{\KnownSlope}{1.2\xspace}
\newcommand{\KnownHighIntercept}{0.3\xspace}
\newcommand{\KnownMiddleIntercept}{-0.3\xspace}
\newcommand{\KnownLowIntercept}{-1.0\xspace}
\newcommand{\KnownOffsetLo}{-0.2\xspace}
\newcommand{\KnownOffsetHi}{0.2\xspace}
\newcommand{\KnownGrey}{0.75\xspace}
\newcommand{\KnownDefault}{-0.08\xspace}
\newcommand{\KnownJitterLo}{-0.02\xspace}
\newcommand{\KnownJitterHi}{0.02\xspace}
\newcommand{\KnownLiftHigh}{0.12\xspace}
\newcommand{\KnownLiftLow}{-0.3\xspace}
\newcommand{\KnownHardestCoordinates}{3, 5, 0, 7, 1, 6, 2, 4\xspace}
\newcommand{\KnownSeedSix}{20261037\xspace}
\newcommand{\KnownSeedEight}{20261048\xspace}
\newcommand{\DatasetLaion}{LAION 400M\xspace}
\newcommand{\DatasetCoyo}{COYO 700M\xspace}
\newcommand{\DatasetCombined}{CC15M\xspace}
\newcommand{\ClipVariant}{B/16\xspace}

\newcommand{\DatasetConceptualTwelve}{Conceptual 12M\xspace}
\newcommand{\KnownShamShift}{12\xspace}
\newcommand{\FourthFirstModels}{18\xspace}
\newcommand{\FourthFirstIndex}{14\xspace}
\newcommand{\FourthFirstDelta}{0.023194\xspace}
\newcommand{\FourthFirstDeltaLo}{-0.006656\xspace}
\newcommand{\FourthFirstDeltaHi}{0.053700\xspace}
\newcommand{\FourthFirstSampledAccuracy}{0.498200\xspace}
\newcommand{\FourthFirstSampledDrop}{0.073964\xspace}
\newcommand{\FourthFirstSampledStrong}{0.050770\xspace}
\newcommand{\FourthFirstSampledLower}{0.052217\xspace}
\newcommand{\FourthFirstSampledSimultaneous}{0.013135\xspace}

\newcommand{\FourthFirstUniformAccuracy}{0.493828\xspace}
\newcommand{\FourthFirstUniformDrop}{0.069592\xspace}
\newcommand{\FourthFirstUniformStrong}{0.046398\xspace}
\newcommand{\FourthFirstUniformLower}{0.048562\xspace}
\newcommand{\FourthFirstUniformSimultaneous}{0.009083\xspace}

\newcommand{\FourthConditionalModels}{14\xspace}
\newcommand{\FourthConditionalIndex}{175\xspace}
\newcommand{\FourthConditionalDelta}{0.008482\xspace}
\newcommand{\FourthConditionalDeltaLo}{-0.014904\xspace}
\newcommand{\FourthConditionalDeltaHi}{0.031119\xspace}
\newcommand{\FourthConditionalSampledAccuracy}{0.502805\xspace}
\newcommand{\FourthConditionalSampledDrop}{0.065573\xspace}
\newcommand{\FourthConditionalSampledStrong}{0.057091\xspace}
\newcommand{\FourthConditionalSampledLower}{0.042489\xspace}
\newcommand{\FourthConditionalUniformAccuracy}{0.498850\xspace}
\newcommand{\FourthConditionalUniformDrop}{0.061618\xspace}
\newcommand{\FourthConditionalUniformStrong}{0.053136\xspace}
\newcommand{\FourthConditionalUniformLower}{0.038917\xspace}
\newcommand{\FourthAllModels}{20\xspace}
\newcommand{\FourthAllIndex}{22\xspace}
\newcommand{\FourthAllDelta}{0.007062\xspace}
\newcommand{\FourthAllDeltaLo}{-0.018972\xspace}
\newcommand{\FourthAllDeltaHi}{0.034543\xspace}
\newcommand{\FourthAllSampledAccuracy}{0.500562\xspace}
\newcommand{\FourthAllSampledDrop}{0.069312\xspace}
\newcommand{\FourthAllSampledStrong}{0.062250\xspace}
\newcommand{\FourthAllSampledLower}{0.045995\xspace}
\newcommand{\FourthAllUniformAccuracy}{0.496500\xspace}
\newcommand{\FourthAllUniformDrop}{0.065250\xspace}
\newcommand{\FourthAllUniformStrong}{0.058188\xspace}
\newcommand{\FourthAllUniformLower}{0.042438\xspace}
\newcommand{\FourthFixedModels}{18\xspace}
\newcommand{\FourthFixedIndex}{175\xspace}
\newcommand{\FourthFixedDelta}{0.006458\xspace}
\newcommand{\FourthFixedDeltaLo}{-0.016617\xspace}
\newcommand{\FourthFixedDeltaHi}{0.028472\xspace}
\newcommand{\FourthFixedSampledAccuracy}{0.494950\xspace}
\newcommand{\FourthFixedSampledDrop}{0.070714\xspace}
\newcommand{\FourthFixedSampledStrong}{0.064256\xspace}
\newcommand{\FourthFixedSampledLower}{0.047805\xspace}
\newcommand{\FourthFixedUniformAccuracy}{0.490833\xspace}
\newcommand{\FourthFixedUniformDrop}{0.066597\xspace}
\newcommand{\FourthFixedUniformStrong}{0.060139\xspace}
\newcommand{\FourthFixedUniformLower}{0.044296\xspace}

\newcommand{\FirstEasyAaGrid}{0.326\xspace}
\newcommand{\FirstEasyAaHard}{0.232\xspace}
\newcommand{\FirstEasyAaEasy}{0.449\xspace}

\newcommand{\FirstEasyAbGrid}{0.506\xspace}
\newcommand{\FirstEasyAbHard}{0.401\xspace}
\newcommand{\FirstEasyAbEasy}{0.611\xspace}

\newcommand{\FirstEasyAcGrid}{0.407\xspace}
\newcommand{\FirstEasyAcHard}{0.306\xspace}
\newcommand{\FirstEasyAcEasy}{0.524\xspace}

\newcommand{\FirstEasyAdGrid}{0.468\xspace}
\newcommand{\FirstEasyAdHard}{0.385\xspace}
\newcommand{\FirstEasyAdEasy}{0.566\xspace}

\newcommand{\FirstEasyAeGrid}{0.398\xspace}
\newcommand{\FirstEasyAeHard}{0.302\xspace}
\newcommand{\FirstEasyAeEasy}{0.535\xspace}

\newcommand{\FirstEasyAfGrid}{0.468\xspace}
\newcommand{\FirstEasyAfHard}{0.364\xspace}
\newcommand{\FirstEasyAfEasy}{0.595\xspace}

\newcommand{\FirstEasyAgGrid}{0.479\xspace}
\newcommand{\FirstEasyAgHard}{0.400\xspace}
\newcommand{\FirstEasyAgEasy}{0.566\xspace}

\newcommand{\FirstEasyAhGrid}{0.513\xspace}
\newcommand{\FirstEasyAhHard}{0.415\xspace}
\newcommand{\FirstEasyAhEasy}{0.624\xspace}

\newcommand{\FirstEasyAiGrid}{0.536\xspace}
\newcommand{\FirstEasyAiHard}{0.445\xspace}
\newcommand{\FirstEasyAiEasy}{0.629\xspace}

\newcommand{\FirstEasyAjGrid}{0.510\xspace}
\newcommand{\FirstEasyAjHard}{0.443\xspace}
\newcommand{\FirstEasyAjEasy}{0.581\xspace}

\newcommand{\FirstEasyAkGrid}{0.550\xspace}
\newcommand{\FirstEasyAkHard}{0.474\xspace}
\newcommand{\FirstEasyAkEasy}{0.638\xspace}

\newcommand{\FirstEasyAlGrid}{0.483\xspace}
\newcommand{\FirstEasyAlHard}{0.415\xspace}
\newcommand{\FirstEasyAlEasy}{0.547\xspace}

\newcommand{\FirstEasyAmGrid}{0.527\xspace}
\newcommand{\FirstEasyAmHard}{0.490\xspace}
\newcommand{\FirstEasyAmEasy}{0.565\xspace}

\newcommand{\FirstEasyAnGrid}{0.560\xspace}
\newcommand{\FirstEasyAnHard}{0.496\xspace}
\newcommand{\FirstEasyAnEasy}{0.639\xspace}

\newcommand{\FirstEasyAoGrid}{0.481\xspace}
\newcommand{\FirstEasyAoHard}{0.446\xspace}
\newcommand{\FirstEasyAoEasy}{0.552\xspace}

\newcommand{\FirstEasyApGrid}{0.582\xspace}
\newcommand{\FirstEasyApHard}{0.552\xspace}
\newcommand{\FirstEasyApEasy}{0.629\xspace}

\newcommand{\FirstEasyAqGrid}{0.522\xspace}
\newcommand{\FirstEasyAqHard}{0.506\xspace}
\newcommand{\FirstEasyAqEasy}{0.551\xspace}

\newcommand{\FirstEasyArGrid}{0.573\xspace}
\newcommand{\FirstEasyArHard}{0.562\xspace}
\newcommand{\FirstEasyArEasy}{0.623\xspace}

\newcommand{\NewRefusedAaSampledDrop}{0.095\xspace}
\newcommand{\NewRefusedAaSampledLo}{0.064\xspace}
\newcommand{\NewRefusedAaSampledHi}{0.125\xspace}
\newcommand{\NewRefusedAaUniformDrop}{0.089\xspace}
\newcommand{\NewRefusedAaUniformLo}{0.059\xspace}
\newcommand{\NewRefusedAaUniformHi}{0.119\xspace}

\newcommand{\NewRefusedAbSampledDrop}{0.099\xspace}
\newcommand{\NewRefusedAbSampledLo}{0.064\xspace}
\newcommand{\NewRefusedAbSampledHi}{0.133\xspace}
\newcommand{\NewRefusedAbUniformDrop}{0.094\xspace}
\newcommand{\NewRefusedAbUniformLo}{0.059\xspace}
\newcommand{\NewRefusedAbUniformHi}{0.128\xspace}

\newcommand{\NewRefusedAcSampledDrop}{0.077\xspace}
\newcommand{\NewRefusedAcSampledLo}{0.040\xspace}
\newcommand{\NewRefusedAcSampledHi}{0.112\xspace}
\newcommand{\NewRefusedAcUniformDrop}{0.073\xspace}
\newcommand{\NewRefusedAcUniformLo}{0.037\xspace}
\newcommand{\NewRefusedAcUniformHi}{0.108\xspace}

\newcommand{\NewRefusedAdSampledDrop}{0.054\xspace}
\newcommand{\NewRefusedAdSampledLo}{0.019\xspace}
\newcommand{\NewRefusedAdSampledHi}{0.091\xspace}
\newcommand{\NewRefusedAdUniformDrop}{0.050\xspace}
\newcommand{\NewRefusedAdUniformLo}{0.016\xspace}
\newcommand{\NewRefusedAdUniformHi}{0.086\xspace}

\newcommand{\NewRefusedAeSampledDrop}{0.084\xspace}
\newcommand{\NewRefusedAeSampledLo}{0.048\xspace}
\newcommand{\NewRefusedAeSampledHi}{0.118\xspace}
\newcommand{\NewRefusedAeUniformDrop}{0.080\xspace}
\newcommand{\NewRefusedAeUniformLo}{0.045\xspace}
\newcommand{\NewRefusedAeUniformHi}{0.114\xspace}

\newcommand{\NewRefusedAfSampledDrop}{0.060\xspace}
\newcommand{\NewRefusedAfSampledLo}{0.022\xspace}
\newcommand{\NewRefusedAfSampledHi}{0.093\xspace}
\newcommand{\NewRefusedAfUniformDrop}{0.056\xspace}
\newcommand{\NewRefusedAfUniformLo}{0.020\xspace}
\newcommand{\NewRefusedAfUniformHi}{0.090\xspace}
\newcommand{\FourthKnownFgConstant}{$1.7\times 10^{-18}$\xspace}
\newcommand{\FourthKnownFgBoundError}{$1.1\times 10^{-16}$\xspace}
\newcommand{\FourthKnownCommonConstant}{0.090\xspace}
\newcommand{\FourthKnownCommonOrdinary}{0.110\xspace}
\newcommand{\FourthKnownCommonSimultaneous}{0.110\xspace}
\newcommand{\FourthKnownCellConstant}{$8.4\times 10^{-17}$\xspace}
\newcommand{\FourthKnownCellOrdinary}{0.356\xspace}

\newcommand{\CompactFirstSampledDrop}{0.074\xspace}
\newcommand{\CompactFirstSampledLower}{0.052\xspace}
\newcommand{\CompactFirstSampledSimultaneous}{0.013\xspace}
\newcommand{\CompactFirstUniformDrop}{0.070\xspace}
\newcommand{\CompactFirstUniformLower}{0.049\xspace}
\newcommand{\CompactFirstUniformSimultaneous}{0.009\xspace}
\newcommand{\CompactFirstDelta}{0.023\xspace}
\newcommand{\CompactFirstDeltaLo}{-0.007\xspace}
\newcommand{\CompactFirstDeltaHi}{0.054\xspace}
\newcommand{\CompactConditionalSampledDrop}{0.066\xspace}
\newcommand{\CompactConditionalSampledLower}{0.042\xspace}
\newcommand{\CompactConditionalUniformDrop}{0.062\xspace}
\newcommand{\CompactConditionalUniformLower}{0.039\xspace}
\newcommand{\CompactConditionalDelta}{0.008\xspace}
\newcommand{\CompactConditionalDeltaLo}{-0.015\xspace}
\newcommand{\CompactConditionalDeltaHi}{0.031\xspace}
\newcommand{\CompactAllSampledDrop}{0.069\xspace}
\newcommand{\CompactAllSampledLower}{0.046\xspace}
\newcommand{\CompactAllUniformDrop}{0.065\xspace}
\newcommand{\CompactAllUniformLower}{0.042\xspace}
\newcommand{\CompactAllDelta}{0.007\xspace}
\newcommand{\CompactAllDeltaLo}{-0.019\xspace}
\newcommand{\CompactAllDeltaHi}{0.035\xspace}
\newcommand{\CompactFixedSampledDrop}{0.071\xspace}
\newcommand{\CompactFixedSampledLower}{0.048\xspace}
\newcommand{\CompactFixedUniformDrop}{0.067\xspace}
\newcommand{\CompactFixedUniformLower}{0.044\xspace}
\newcommand{\CompactFixedDelta}{0.006\xspace}
\newcommand{\CompactFixedDeltaLo}{-0.017\xspace}
\newcommand{\CompactFixedDeltaHi}{0.028\xspace}
\newcommand{\OrderChangesFirst}{15\xspace}
\newcommand{\OrderChangesResolvedFirst}{8\xspace}

\newcommand{\OrderRhoFirst}{0.93\xspace}

\newcommand{\PairFirstAAll}{0.023\xspace}
\newcommand{\PairFirstAAllLow}{0.004\xspace}
\newcommand{\PairFirstAHard}{-0.014\xspace}
\newcommand{\PairFirstAHardLo}{-0.052\xspace}
\newcommand{\PairFirstAHardHi}{0.023\xspace}

\newcommand{\PairFirstBAll}{0.025\xspace}
\newcommand{\PairFirstBAllLow}{0.005\xspace}
\newcommand{\PairFirstBHard}{-0.045\xspace}
\newcommand{\PairFirstBHardLo}{-0.087\xspace}
\newcommand{\PairFirstBHardHi}{-0.002\xspace}

\newcommand{\PairFirstCAll}{0.032\xspace}
\newcommand{\PairFirstCAllLow}{0.013\xspace}
\newcommand{\PairFirstCHard}{-0.031\xspace}
\newcommand{\PairFirstCHardLo}{-0.068\xspace}
\newcommand{\PairFirstCHardHi}{0.007\xspace}

\newcommand{\PairFirstDAll}{0.056\xspace}
\newcommand{\PairFirstDAllLow}{0.037\xspace}
\newcommand{\PairFirstDHard}{-0.001\xspace}
\newcommand{\PairFirstDHardLo}{-0.045\xspace}
\newcommand{\PairFirstDHardHi}{0.040\xspace}

\newcommand{\PairFirstEAll}{0.029\xspace}
\newcommand{\PairFirstEAllLow}{0.014\xspace}
\newcommand{\PairFirstEHard}{-0.004\xspace}
\newcommand{\PairFirstEHardLo}{-0.040\xspace}
\newcommand{\PairFirstEHardHi}{0.032\xspace}

\newcommand{\PairFirstFAll}{0.022\xspace}
\newcommand{\PairFirstFAllLow}{0.006\xspace}
\newcommand{\PairFirstFHard}{-0.016\xspace}
\newcommand{\PairFirstFHardLo}{-0.056\xspace}
\newcommand{\PairFirstFHardHi}{0.024\xspace}

\newcommand{\PairFirstGAll}{0.028\xspace}
\newcommand{\PairFirstGAllLow}{0.007\xspace}
\newcommand{\PairFirstGHard}{-0.032\xspace}
\newcommand{\PairFirstGHardLo}{-0.079\xspace}
\newcommand{\PairFirstGHardHi}{0.014\xspace}

\newcommand{\PairFirstHAll}{0.038\xspace}
\newcommand{\PairFirstHAllLow}{0.019\xspace}
\newcommand{\PairFirstHHard}{-0.010\xspace}
\newcommand{\PairFirstHHardLo}{-0.053\xspace}
\newcommand{\PairFirstHHardHi}{0.034\xspace}

\newcommand{\OrderChangesNew}{8\xspace}
\newcommand{\OrderChangesResolvedNew}{2\xspace}

\newcommand{\OrderRhoNew}{0.95\xspace}

\newcommand{\PairNewAAll}{0.026\xspace}
\newcommand{\PairNewAAllLow}{0.008\xspace}
\newcommand{\PairNewAHard}{-0.021\xspace}
\newcommand{\PairNewAHardLo}{-0.058\xspace}
\newcommand{\PairNewAHardHi}{0.016\xspace}

\newcommand{\PairNewBAll}{0.021\xspace}
\newcommand{\PairNewBAllLow}{0.003\xspace}
\newcommand{\PairNewBHard}{-0.038\xspace}
\newcommand{\PairNewBHardLo}{-0.074\xspace}
\newcommand{\PairNewBHardHi}{-0.001\xspace}

\newcommand{\ComparatorDrop}{0.053\xspace}
\newcommand{\ComparatorLower}{0.031\xspace}
\newcommand{\ComparatorDelta}{0.012\xspace}
\newcommand{\ComparatorDeltaLo}{-0.020\xspace}
\newcommand{\ComparatorDeltaHi}{0.043\xspace}
\newcommand{\ComparatorIndex}{14\xspace}
\newcommand{\ChangeSignedLo}{-0.031\xspace}
\newcommand{\ChangeSignedHi}{0.031\xspace}
\newcommand{\ChangeAbsLo}{0.016\xspace}
\newcommand{\ChangeAbsHi}{0.040\xspace}
\newcommand{\RerunUniformDrop}{0.070\xspace}
\newcommand{\RerunUniformLower}{0.048\xspace}
\newcommand{\RerunSampledDrop}{0.075\xspace}
\newcommand{\RerunSampledLower}{0.051\xspace}

\newcommand{\RerunModels}{20\xspace}
\newcommand{\RerunChange}{0.005\xspace}
\newcommand{\RerunChangeLo}{-0.025\xspace}
\newcommand{\RerunChangeHi}{0.036\xspace}
\newcommand{\SeedRerun}{20261109\xspace}
\newcommand{\CoocTotalA}{12,251\xspace}
\newcommand{\CoocTotalB}{3,952\xspace}
\newcommand{\CoocTotalC}{4,385\xspace}
\newcommand{\CoocTotalD}{6,127\xspace}
\newcommand{\CoocTotalE}{4,114\xspace}
\newcommand{\CoocTotalF}{2,986\xspace}
\newcommand{\CoocTotalG}{2,941\xspace}
\newcommand{\CoocTotalH}{3,588\xspace}
\newcommand{\ArchAa}{PP-LCNet 0.5x\xspace}
\newcommand{\LicNameAa}{Apache 2.0\xspace}
\newcommand{\ArchAb}{MobileNetV3-Small\xspace}
\newcommand{\LicNameAb}{Apache 2.0\xspace}
\newcommand{\ArchAc}{GhostNet 1.0x\xspace}
\newcommand{\LicNameAc}{Apache 2.0\xspace}
\newcommand{\ArchAd}{VGG-16\xspace}
\newcommand{\LicNameAd}{BSD 3-Clause\xspace}
\newcommand{\ArchAe}{MnasNet 1.0x\xspace}
\newcommand{\LicNameAe}{Apache 2.0\xspace}
\newcommand{\ArchAf}{FBNet-C\xspace}
\newcommand{\LicNameAf}{Apache 2.0\xspace}
\newcommand{\ArchAg}{StarNet-S2\xspace}
\newcommand{\LicNameAg}{Apache 2.0\xspace}
\newcommand{\ArchAh}{SK-ResNet-18\xspace}
\newcommand{\LicNameAh}{Apache 2.0\xspace}
\newcommand{\ArchAi}{DenseNet-121\xspace}
\newcommand{\LicNameAi}{Apache 2.0\xspace}
\newcommand{\ArchAj}{Single-Path NAS\xspace}
\newcommand{\LicNameAj}{Apache 2.0\xspace}
\newcommand{\ArchAk}{LeViT-128S\xspace}
\newcommand{\LicNameAk}{Apache 2.0\xspace}
\newcommand{\ArchAl}{FBNetV3-D\xspace}
\newcommand{\LicNameAl}{Apache 2.0\xspace}
\newcommand{\ArchAm}{MixNet-M\xspace}
\newcommand{\LicNameAm}{Apache 2.0\xspace}
\newcommand{\ArchAn}{DenseNet-121 Blur\xspace}
\newcommand{\LicNameAn}{Apache 2.0\xspace}
\newcommand{\ArchAo}{DLA-X-60\xspace}
\newcommand{\LicNameAo}{BSD 3-Clause\xspace}
\newcommand{\ArchAp}{RepGhostNet 1.3x\xspace}
\newcommand{\LicNameAp}{MIT\xspace}
\newcommand{\ArchAq}{CaiT-XXS24\xspace}
\newcommand{\LicNameAq}{Apache 2.0\xspace}
\newcommand{\ArchAr}{FasterNet-T2\xspace}
\newcommand{\LicNameAr}{Apache 2.0\xspace}
\newcommand{\ArchAs}{CoAtNeXt Nano\xspace}
\newcommand{\LicNameAs}{Apache 2.0\xspace}
\newcommand{\ArchAt}{ResMLP-36\xspace}
\newcommand{\LicNameAt}{Apache 2.0\xspace}
\newcommand{\ArchAu}{PoolFormer-S24\xspace}
\newcommand{\LicNameAu}{Apache 2.0\xspace}
\newcommand{\ArchAv}{Wide ResNet-50-2\xspace}
\newcommand{\LicNameAv}{Apache 2.0\xspace}
\newcommand{\ArchAw}{PoolFormerV2-S24\xspace}
\newcommand{\LicNameAw}{Apache 2.0\xspace}
\newcommand{\ArchAx}{ReXNet 1.3x\xspace}
\newcommand{\LicNameAx}{MIT\xspace}
\newcommand{\ArchAy}{CoAtNet-0\xspace}
\newcommand{\LicNameAy}{Apache 2.0\xspace}
\newcommand{\ArchAz}{Antialiased ResNet-50\xspace}
\newcommand{\LicNameAz}{Apache 2.0\xspace}
\newcommand{\ArchBa}{InceptionNeXt-S\xspace}
\newcommand{\LicNameBa}{Apache 2.0\xspace}
\newcommand{\ArchBb}{RegNetX-16GF\xspace}
\newcommand{\LicNameBb}{BSD 3-Clause\xspace}
\newcommand{\ArchBc}{BEiT v2 Base\xspace}
\newcommand{\LicNameBc}{Apache 2.0\xspace}
\newcommand{\ArchBd}{EfficientFormerV2-L\xspace}
\newcommand{\LicNameBd}{Apache 2.0\xspace}
\newcommand{\ArchBe}{FocalNet-B\xspace}
\newcommand{\LicNameBe}{MIT\xspace}
\newcommand{\ArchBf}{ReXNet-R 2.0x\xspace}
\newcommand{\LicNameBf}{Apache 2.0\xspace}
\newcommand{\ArchBg}{XCiT-S24\xspace}
\newcommand{\LicNameBg}{Apache 2.0\xspace}
\newcommand{\ArchBh}{PP-HGNetV2-B6\xspace}
\newcommand{\LicNameBh}{Apache 2.0\xspace}
\newcommand{\ArchBi}{VOLO-D5\xspace}
\newcommand{\LicNameBi}{Apache 2.0\xspace}
\newcommand{\ArchBj}{EVA Giant\xspace}
\newcommand{\LicNameBj}{MIT\xspace}
\newcommand{\ArchBk}{ConvNeXt-XL\xspace}
\newcommand{\LicNameBk}{Apache 2.0\xspace}
\newcommand{\ArchBl}{BEiT-3 Large\xspace}
\newcommand{\LicNameBl}{Apache 2.0\xspace}
\newcommand{\ArchBm}{CAFormer-S36\xspace}
\newcommand{\LicNameBm}{Apache 2.0\xspace}
\newcommand{\ArchBn}{MaxxViT-V2 Base\xspace}
\newcommand{\LicNameBn}{Apache 2.0\xspace}
\newcommand{\LineDecreaseFirst}{-0.137\xspace}
\newcommand{\LineDecreaseFirstLow}{-0.290\xspace}
\newcommand{\LineDecreaseFirstHigh}{-0.040\xspace}
\newcommand{\LineDecreaseNew}{-0.056\xspace}
\newcommand{\LineDecreaseNewLow}{-0.146\xspace}
\newcommand{\LineDecreaseNewHigh}{0.013\xspace}
\newcommand{\PairedModelLo}{-0.005\xspace}
\newcommand{\PairedModelHi}{0.006\xspace}
\newcommand{\PoolImages}{1,600\xspace}
\newcommand{\PoolForegrounds}{800\xspace}
\newcommand{\PoolWindows}{800\xspace}
\newcommand{\PoolModels}{20\xspace}

\newcommand{\PoolHardDrop}{0.067\xspace}
\newcommand{\PoolHardLower}{0.045\xspace}
\newcommand{\PoolCompDrop}{0.055\xspace}
\newcommand{\PoolCompLower}{0.032\xspace}
\newcommand{\PoolDelta}{0.012\xspace}
\newcommand{\PoolDeltaLo}{-0.019\xspace}
\newcommand{\PoolDeltaHi}{0.044\xspace}
\newcommand{\SeedPool}{20261110\xspace}
\newcommand{\PoolUsed}{3,404\xspace}
\newcommand{\PoolSmallest}{143\xspace}
\newcommand{\BlurSigma}{2\xspace}
\newcommand{\BlurTruncate}{4\xspace}
\newcommand{\BlurModels}{20\xspace}
\newcommand{\BlurDrop}{0.062\xspace}
\newcommand{\BlurLower}{0.040\xspace}

\newcommand{\BlurChange}{-0.003\xspace}
\newcommand{\BlurChangeLo}{-0.011\xspace}
\newcommand{\BlurChangeHi}{0.004\xspace}
\newcommand{\BlurControlImages}{800\xspace}

\newcommand{\PointBeyondBoth}{5\xspace}
\newcommand{\OrderChangesResolvedBoth}{10\xspace}
\newcommand{\EasyGapFirst}{0.344\xspace}
\newcommand{\EasyGapFirstLo}{0.120\xspace}
\newcommand{\EasyGapFirstHi}{0.743\xspace}
\newcommand{\EasyGapNew}{0.211\xspace}
\newcommand{\EasyGapNewLo}{0.065\xspace}
\newcommand{\EasyGapNewHi}{0.613\xspace}
\newcommand{\EasyAbNew}{0.441\xspace}
\newcommand{\EasyAdNew}{0.602\xspace}
\newcommand{\EasyAeNew}{0.517\xspace}
\newcommand{\EasyAiNew}{0.568\xspace}
\newcommand{\EasyAkNew}{0.596\xspace}
\newcommand{\EasyAlNew}{0.549\xspace}
\newcommand{\EasyAzNew}{0.542\xspace}
\newcommand{\EasyBeNew}{0.574\xspace}
\newcommand{\EasyBfNew}{0.636\xspace}
\newcommand{\EasyBgNew}{0.546\xspace}
\newcommand{\EasyBhNew}{0.649\xspace}
\newcommand{\EasyBjNew}{0.578\xspace}
\newcommand{\EasyBmNew}{0.651\xspace}
\newcommand{\EasyBnNew}{0.631\xspace}
\newcommand{\RemovalPairings}{720\xspace}
\newcommand{\RemovalSum}{-0.66\xspace}
\newcommand{\RemovalShared}{6\xspace}
\newcommand{\RemovalBaseFirst}{0.496\xspace}
\newcommand{\RemovalAccFirst}{0.458\xspace}
\newcommand{\RemovalDropFirst}{0.038\xspace}
\newcommand{\RemovalLowerFirst}{0.014\xspace}

\newcommand{\RemovalBaseNew}{0.498\xspace}
\newcommand{\RemovalAccNew}{0.467\xspace}
\newcommand{\RemovalDropNew}{0.030\xspace}
\newcommand{\RemovalLowerNew}{0.006\xspace}

\newcommand{\RemovalBaseThird}{0.518\xspace}
\newcommand{\RemovalAccThird}{0.486\xspace}
\newcommand{\RemovalDropThird}{0.032\xspace}
\newcommand{\RemovalLowerThird}{0.008\xspace}

\newcommand{\RemovalCells}{36\xspace}
\newcommand{\GroupInOneK}{15\xspace}
\newcommand{\GroupInOneKDropFirst}{0.080\xspace}
\newcommand{\GroupInOneKLowerFirst}{0.058\xspace}
\newcommand{\GroupInOneKDropNew}{0.072\xspace}
\newcommand{\GroupInOneKLowerNew}{0.049\xspace}
\newcommand{\GroupInOneKDropThird}{0.077\xspace}
\newcommand{\GroupInOneKLowerThird}{0.055\xspace}
\newcommand{\GroupNoTeacher}{13\xspace}
\newcommand{\GroupNoTeacherDropFirst}{0.078\xspace}
\newcommand{\GroupNoTeacherLowerFirst}{0.055\xspace}
\newcommand{\GroupNoTeacherDropNew}{0.070\xspace}
\newcommand{\GroupNoTeacherLowerNew}{0.047\xspace}
\newcommand{\GroupNoTeacherDropThird}{0.075\xspace}
\newcommand{\GroupNoTeacherLowerThird}{0.052\xspace}
\newcommand{\GroupDistilled}{2\xspace}
\newcommand{\GroupDistilledDropFirst}{0.098\xspace}
\newcommand{\GroupDistilledLowerFirst}{0.070\xspace}
\newcommand{\GroupDistilledDropNew}{0.083\xspace}
\newcommand{\GroupDistilledLowerNew}{0.052\xspace}
\newcommand{\GroupDistilledDropThird}{0.091\xspace}
\newcommand{\GroupDistilledLowerThird}{0.062\xspace}
\newcommand{\GroupFurther}{5\xspace}
\newcommand{\GroupFurtherDropFirst}{0.033\xspace}
\newcommand{\GroupFurtherLowerFirst}{0.006\xspace}
\newcommand{\GroupFurtherDropNew}{0.046\xspace}
\newcommand{\GroupFurtherLowerNew}{0.020\xspace}
\newcommand{\GroupFurtherDropThird}{0.035\xspace}
\newcommand{\GroupFurtherLowerThird}{0.009\xspace}
\newcommand{\GroupGapFirst}{0.048\xspace}
\newcommand{\GroupGapFirstLo}{0.029\xspace}
\newcommand{\GroupGapFirstHi}{0.066\xspace}
\newcommand{\GroupGapNew}{0.025\xspace}
\newcommand{\GroupGapNewLo}{0.007\xspace}
\newcommand{\GroupGapNewHi}{0.043\xspace}
\newcommand{\GroupGapThird}{0.042\xspace}
\newcommand{\GroupGapThirdLo}{0.023\xspace}
\newcommand{\GroupGapThirdHi}{0.061\xspace}
\newcommand{\GroupFloor}{10\xspace}
\newcommand{\RebuildMasks}{800\xspace}

\newcommand{\PoolHardSampledDrop}{0.071\xspace}
\newcommand{\PoolHardSampledLower}{0.048\xspace}
\newcommand{\PoolWithin}{0\xspace}
\newcommand{\AbsRandomNew}{50.3\xspace}
\newcommand{\AbsHardNew}{43.7\xspace}
\newcommand{\AbsDropNew}{6.6\xspace}

\newcommand{\AbsRandomFirst}{49.8\xspace}
\newcommand{\AbsHardFirst}{42.4\xspace}
\newcommand{\AbsDropFirst}{7.4\xspace}
\newcommand{\AbsReductionNew}{13\xspace}
\newcommand{\AbsReductionFirst}{15\xspace}

\newcommand{\MeanAbsoluteChangePoints}{1.12\xspace}

\title{A Hard Background Pairing Transfers across Classifiers and Images}
\newcommand{\authoremail}{kl3685@columbia.edu}
\author{%
  Karl Li \\
  Columbia University \\
  {\normalfont\small\texttt{\authoremail}}
}
\hypersetup{pdfauthor={Karl Li},pdftitle={A Hard Background Pairing Transfers across Classifiers and Images}}
\date{}
\begin{document}
\maketitle
\begin{abstract}
An image classifier names the object in an image, and its accuracy is the share of images named correctly. A classifier that names a car on a road may miss the same car pasted onto clouds. Composite benchmarks measure such failures by pasting objects cut from photographs onto backgrounds from other photographs. A \emph{pairing} puts every object label on a different background label, such as truck on sea. The pairing on which some classifiers do worst, their \emph{hardest pairing}, makes a strict test, but may be hard only for them or only on the photographs used to choose it. We show that the hardest pairing of some classifiers stays hard for others and on new photographs. We split \Models ImageNet classifiers into halves balanced by reported ImageNet accuracy. One half picks its hardest pairing of \LabelsPerSide object and \LabelsPerSide background labels from Common Objects in Context (COCO). The other half, held out of the choice, compares that pairing with random pairings of the same objects and backgrounds. For the \AdmittedNew held-out classifiers with little net accuracy change after pasting objects back onto their own backgrounds, mean accuracy on composites from new photographs falls from \AbsRandomNew{}\% over random pairings to \AbsHardNew{}\% on the hardest pairing, \AbsDropNew percentage points lower. A benchmark using these composites can compare this pool's held-out classifiers on the fixed and random pairings.
\end{abstract}

\section{Introduction}
\label{sec:intro}

\begin{figure}[t]
  \centering
  \includegraphics{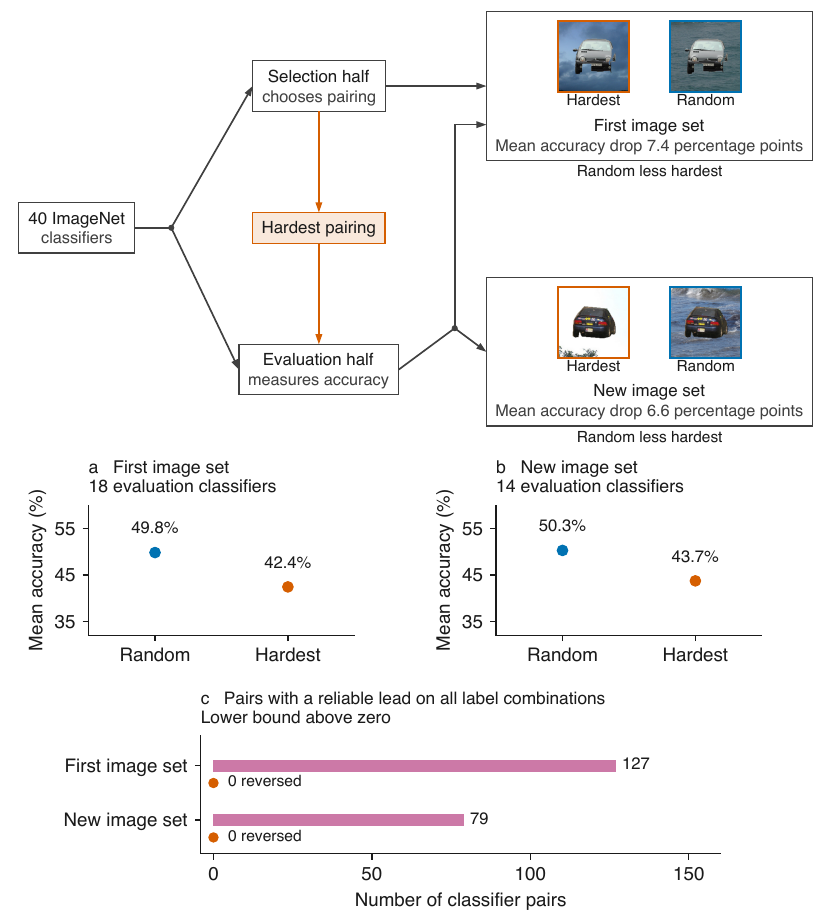}
  \caption{\textbf{The hardest pairing lowers mean accuracy for evaluation classifiers passing each image set's compositing check; accounting for the search puts the first image set's lower bound below the margin.} Each set's first car shows clouds and the first different random background. Panel c counts \ResolvedFirst and \ResolvedNew reliable leads; none has an upper bound below $-\ReversalMargin$ on the hardest pairing.}
  \label{fig:overview}
\end{figure}

\Cref{fig:overview} shows one car, cut from its photograph and pasted onto clouds and onto the sea. An image classifier answers correctly on either picture when it names the car, so the two pictures test the same object against different backgrounds. Benchmarks built from such composite images compare a classifier's accuracy on the same objects across backgrounds. \INnine separates ImageNet objects from their backgrounds and measures accuracy when the background is removed or replaced by one from another image~\cite{xiao2021noise}. Later benchmarks vary the background in generated or edited images~\cite{lynch2023spawrious,dunkel2025cns} and in photographs sorted by their context~\cite{liang2022metashift,wang2024sober}. Each such benchmark must decide which background goes behind which object, for instance whether airplanes appear on buildings or on clouds.

A pairing makes that decision for whole labels, giving each object label a background label that no other object label receives, such as building for airplane and sea for truck. Most benchmarks choose backgrounds at random or by the scene in which an object usually appears. \INnine also searched each image for the background that hurts a given model most~\cite{xiao2021noise}, a choice made for one image and one model rather than for a whole label. A benchmark could instead fix the hardest pairing for some classifiers, the pairing on which they are least accurate. It must fix that pairing once, since a pairing searched anew for each classifier gives each one a different test set, and their accuracies could not be compared. The fixed pairing transfers across classifiers if it lowers the accuracy of classifiers that took no part in choosing it below their accuracy on pairings drawn at random. It transfers across images if it does so on photographs that took no part either (\cref{sec:measures}).

Studies of hard test sets ask whether examples chosen as hard on some models stay hard for models left out of the choice, and they find that the difficulty transfers~\cite{lebras2020adversarial,salaudeen2025aggregation,zhang2024imagenetd}. These studies choose single examples, while a pairing is one rule for every image of a label. None of them fixes a pairing of labels on some classifiers and then measures it on what the choice never saw: other classifiers and new photographs.

We show that the hardest pairing transfers across classifiers and across images. We build the test from photographs of Common Objects in Context (COCO)~\cite{lin2014microsoft} and \Models ImageNet classifiers from the PyTorch Image Models (timm) library~\cite{wightman2026timm}, which we split into two halves balanced by reported ImageNet accuracy. One half, the selection half, scores every combination of an object label and a background label, and we fix the pairing on which that half is least accurate as the hardest pairing. The other half, the evaluation half, compares that pairing with \RandomPairings random pairings that show the same objects and backgrounds at the same label frequencies. It does so on the photographs used in the choice, which we call the first image set, and on a new image set of \ImagesNew photographs that the choice never used. Pasting alone can change a classifier's answers, and a fall in its accuracy would then mix the effect of the paste with that of the background. Our tests therefore use only the evaluation classifiers with little net change in accuracy when we paste each object back onto its own photograph (\cref{sec:halves}). \Cref{fig:overview} shows the design, and we make five contributions.

\begin{enumerate}
  \setlength{\itemsep}{1pt}
  \setlength{\parsep}{0pt}
  \setlength{\topsep}{2pt}
  \item \textbf{A transfer test for a fixed pairing} (\cref{sec:test}). We choose the hardest pairing once, with the selection half, and measure it with the evaluation half against random pairings, on the first image set and on the new image set.
  \item \textbf{Transfer across classifiers} (\cref{sec:first}). On the first image set, the hardest pairing lowers mean accuracy by \AbsDropFirst percentage points against random pairings, over the \AdmittedFirst evaluation classifiers with little net change in accuracy after pasting objects back. We call this fall in mean accuracy the drop. Once we allow for having searched all pairings on those same photographs, the drop is too uncertain to show transfer, because the hardest of many pairings looks harder than it is. So the evidence of transfer rests on the new image set.
  \item \textbf{Transfer across images} (\cref{sec:new}). On the new image set, the hardest pairing lowers mean accuracy by \AbsDropNew percentage points against random pairings, over the \AdmittedNew evaluation classifiers with little net change in accuracy after pasting objects back there. The drop meets our criterion for transfer. A further draw of new photographs, which we call the third image set, also shows the drop.
  \item \textbf{Two label combinations account for most of the drop} (\cref{sec:bylabel}). Airplane on building and truck on sea give the two largest drops of single labels on both image sets. Without the object labels airplane and truck and the background labels building and sea, the hardest pairing of the remaining labels still lowers accuracy on the new image set, but the lower bound of that drop is too low to count as transfer.
  \item \textbf{No pair meets the reversal criterion} (\cref{sec:line}). On the hardest pairing some classifiers change places at their point estimates, but no pair in which one classifier leads reliably over all label combinations meets our reversal criterion.
\end{enumerate}

\section{Related work}
\label{sec:related}

\paragraph{Backgrounds and composites.}
Xiao \etal separate the foregrounds of ImageNet images from their backgrounds and measure accuracy when the background is removed and when it is replaced, either at random or by a search against one model for each image~\cite{xiao2021noise}. In their mixed test sets each image draws its new background from the same class or from a random class, so the background that a class meets varies from image to image. Our pairing instead fixes one background label for every object of a label. The benchmark CounterAnimal sets animals against common and uncommon backgrounds in real photographs, and it finds that the examples hard for one family of contrastive vision and language models are hard for others~\cite{wang2024sober}. The dataset MetaShift groups natural images by their annotated contexts~\cite{liang2022metashift}, and the benchmark ORIC builds questions about COCO photographs whose objects and contexts disagree, to test vision and language models~\cite{li2026oric}. Liao and Chen choose backgrounds by their semantic proximity to an object, for analysts who test a model~\cite{liao2025background}.

Other benchmarks generate or edit their images. The dataset Spawrious generates images whose class and location labels correlate many to many~\cite{lynch2023spawrious}. The benchmark CNS-Bench shifts generated nuisances continuously across many classifiers~\cite{dunkel2025cns}, and the benchmark ContextShift changes the context of objects in detection while keeping their identity~\cite{zlotnikov2026contextshift}. Each of these works draws backgrounds in its own way, but none chooses a pairing on some classifiers and applies it to others. We vary only the pairing of labels, and all pairings share the same label frequencies. We measure one pairing across many classifiers.

\paragraph{Hard sets that transfer to other models.}
A second group asks whether examples chosen as hard for some models stay hard for others. Adversarial filtering removes the examples that a simple model predicts correctly and leaves a harder set~\cite{lebras2020adversarial}. The benchmark ImageNet-D keeps the generated images that several surrogate models fail, the surrogates being the models that make the choice~\cite{zhang2024imagenetd}. Those images also lower the accuracy of a target model, one that played no part in the choice. The method OODSelect selects examples by gradients of model outputs and finds that the accuracy relations of its subsets persist on models left out of the selection~\cite{salaudeen2025aggregation}.

Sets built against particular models can be harder than ordinary sets and still treat models unevenly~\cite{phang2021adversarially}. Worst case subpopulations measure a model on the hardest mixture of a given size~\cite{li2024evaluating}, and inference conditional on selection corrects the optimism of a chosen winner~\cite{bakshi2026flexible}. Our selection half plays the part of their surrogates, and our evaluation half that of their targets. We choose a pairing of labels where they choose examples, and we then test it on photographs that no model saw during the choice.

\paragraph{Accuracy on the line.}
A third group relates a model's accuracy under a shift to its accuracy on the original test set. Miller \etal find that the first is close to a linear function of the second once both are transformed by the probit, the quantile function of the standard normal distribution~\cite{miller2021accuracy}. Salaudeen \etal argue that this relation appears when a benchmark keeps the association between spurious features and labels across its splits, and that reversing that association should break it~\cite{salaudeen2025domain}. The ImageNot study, by contrast, finds that a contrasting dataset keeps the ranking of architectures~\cite{salaudeen2024imagenot}. Gender bias benchmarks also report a score beside its sensitivity to the features they vary~\cite{hirota2025bias}, and model comparisons can be unstable even when each model is stable~\cite{bayle2026relative}. We measure this relation on the hardest pairing, which the selection half chose and the evaluation half did not, as a question apart from whether the drop transfers.

\paragraph{Edits that measure reliance.}
Two works that appeared in October 2026 edit images to measure reliance on cues. Shift-Reliance Alignment makes a cue suggest the wrong answer and compares against a matched control~\cite{mohammad2026shift}, and HullProbe swaps parts of vessel images within a class~\cite{mohammad2026hullprobe}. Those works edit a cue within single images, while a pairing fixes one rule for every image of a label. We still need a control of the same kind for our own edit. We call it the compositing check, and it pastes each object back over its own surroundings. It bounds the net change in a classifier's correct answers, a change that comes from the blending every composite shares and from one more crop and resize of those surroundings.

\section{Method}
\label{sec:test}

We build the test in the order of the subsections below. We make composites that change only the background behind each object, and we split the classifiers into a selection half, which chooses the hardest pairing, and an evaluation half. The evaluation half then measures how much accuracy that pairing costs it against random pairings, first on the images used in the choice and then on new images. The second measurement uses different COCO image identifiers and the pairing chosen on the first image set.

\subsection{Composites}
\label{sec:composites}

We paste each object onto each background at one fixed size, so that only the background varies. Objects and backgrounds both come from the \LicenceImages of the \TrainImages COCO 2017 training images that carry the Creative Commons Attribution licence. COCO's instance masks mark the pixels of each object. Common Objects in Context Stuff (COCO-Stuff) labels every pixel of background material, such as sky or grass, which it calls stuff~\cite{caesar2018cocostuff}.

The object labels are COCO's categories of animals and vehicles. Each object label covers the ImageNet classes that the WordNet lexical database~\cite{miller1995wordnet} places under its first noun sense, and a class under two labels goes to neither. We rank the labels by how many images contain a usable object and keep the first \LabelsPerSide: car, bus, dog, truck, cat, airplane, horse and train.

We call a cut object a foreground and a square crop of a photograph's background a background crop. Intuitively, a background crop should show one kind of background and nothing else. Precisely, it is the largest square in which one COCO-Stuff label covers at least \BgPurity of the pixels, no pixel lies inside an annotated object, and at most \BgUnlabelled of the pixels have no label. We rank the background labels by how many images give a background crop and keep the first \LabelsPerSide: sky, tree, concrete wall, grass, snow, sea, building and clouds. Appendix~\ref{sec:supp-composites} gives the sizes and the search grid of both rules, and the names that COCO-Stuff gives three of these labels.

Each label keeps \UnitsPerLabel foregrounds or \UnitsPerLabel background crops, drawn at random from its candidates. These \Foregrounds foregrounds and \Windows background crops come from \ImagesFirst photographs with distinct COCO identifiers, which form the first image set. We scale each foreground until its longer box side spans \FgLongSide pixels and centre it on a canvas of \Canvas by \Canvas pixels. We resize the foreground's whole photograph by bicubic resampling and its mask by bilinear resampling, and a canvas pixel outside the photograph takes the grey level \Grey. We crop each background crop and resize it to the same canvas by bicubic resampling.

The labels form a table with one row for each object label and one column for each background label, and we call each of its \Cells entries a label combination. For example, each car foreground appears once over sky, once over tree and once over every other background label of its row. Precisely, in each label combination we blend each of the object label's \UnitsPerLabel foregrounds over one of the background label's \UnitsPerLabel background crops, matched by a permutation drawn at random for that combination. Each foreground thus appears once in every combination of its row, and each background crop once in every combination of its column. A pairing therefore changes only which foregrounds meet which background crops, and never which foregrounds or crops appear.

Beside the \CompositesAll composites, each foreground gives three control images. The original image is the canvas itself, which shows the object in its own surroundings. The object on grey blends the foreground over flat grey, which removes the background and keeps the object. The check image blends the foreground back over its own surroundings, the part of its photograph that the original shows, cropped and resized as a background crop is. It differs from the original by the blending and by that one extra crop and resize. Every classifier scores all \ImagesScored images of an image set at \Canvas by \Canvas pixels, each normalized by the classifier's own mean and deviation, with no resize or crop of its own.

\subsection{Classifiers}
\label{sec:halves}

We score \Models ImageNet classifiers whose ImageNet accuracies spread widely, so that weak and strong classifiers both appear. They come from timm's results table at version \VerTimm. Of its \ZooRows rows, we drew one checkpoint from each of \Models architecture families, spread over \ZooBins bins of equal width in ImageNet accuracy. They include convolutional networks, vision transformers, multilayer perceptron mixers and hybrids of convolution and attention, and \cref{tab:checkpoints} in Appendix~\ref{sec:supp-classifiers} names each. A checkpoint's ImageNet accuracy is the accuracy that timm's results table gives it on the validation images of ImageNet~\cite{russakovsky2015imagenet}. Over the \Models checkpoints it runs from \TopMin{}\% to \TopMax{}\%.

Intuitively, a classifier answers with the object label whose ImageNet classes it finds most likely. Precisely, a classifier's score for an object label is the probability that its softmax over the ImageNet classes puts on that label's classes. Its prediction is the object label of largest score, and it is correct when it names the foreground's label.

We split the classifiers so that both halves hold weak and strong classifiers in the same proportions. Precisely, we sort the \Models checkpoints by ImageNet accuracy into \Strata strata of two neighbours and send one checkpoint of each stratum to each half by a coin flip. The selection half alone chooses the pairing and supplies the background effects of \cref{sec:pairings}. The evaluation half measures the pairing and takes no part in choosing it.

Pasting alone can change a classifier's answers, and then its drop would mix the effect of the background with the effect of the paste. The compositing check guards against this by comparing check images with original images, which differ by the paste alone. We exclude an evaluation classifier from an image set when its count of correct answers on the \Foregrounds check images differs from its count on the \Foregrounds original images by more than \ShamLimit images. The evaluation classifiers that remain pass the compositing check on that image set. Our tests use only those classifiers, and they need at least \ShamFloor of them.

\begin{table}[t]
  \caption{\textbf{No background raises the selection half's accuracy above grey by more than \CueMargin at its lower bound.} Each entry is the \emph{background effect} of one label combination on the first image set, the selection half's mean change in accuracy from the same objects on grey. Tint marks the hardest pairing, and bold the most negative effect in each column, the direction $\downarrow$ that a hard pairing seeks.}
  \label{tab:effects}
  \centering
  \small
  \begin{tabular}{@{}lrrrrrrrr@{}}
\toprule
\multicolumn{9}{@{}l}{Entries give the change in fraction correct against the same objects on grey.} \\
& sky $\downarrow$ & tree $\downarrow$ & \shortstack{concrete\\wall} $\downarrow$ & grass $\downarrow$ & snow $\downarrow$ & sea $\downarrow$ & building $\downarrow$ & clouds $\downarrow$ \\
\midrule
car & $\BestLL{\EffAa}$ & $\BestLM{\EffAb}$ & $\BestLN{\EffAc}$ & $\BestLO{\EffAd}$ & $\BestLP{\EffAe}$ & $\BestLQ{\EffAf}$ & $\BestLR{\EffAg}$ & \cellcolor{MethodTint}$\BestLS{\EffAh}$ \\
bus & $\BestLT{\EffBa}$ & $\BestLU{\EffBb}$ & \cellcolor{MethodTint}$\BestLV{\EffBc}$ & $\BestLW{\EffBd}$ & $\BestLX{\EffBe}$ & $\BestLY{\EffBf}$ & $\BestLZ{\EffBg}$ & $\BestMA{\EffBh}$ \\
dog & \cellcolor{MethodTint}$\BestMB{\EffCa}$ & $\BestMC{\EffCb}$ & $\BestMD{\EffCc}$ & $\BestME{\EffCd}$ & $\BestMF{\EffCe}$ & $\BestMG{\EffCf}$ & $\BestMH{\EffCg}$ & $\BestMI{\EffCh}$ \\
truck & $\BestMJ{\EffDa}$ & $\BestMK{\EffDb}$ & $\BestML{\EffDc}$ & $\BestMM{\EffDd}$ & $\BestMN{\EffDe}$ & \cellcolor{MethodTint}$\BestMO{\EffDf}$ & $\BestMP{\EffDg}$ & $\BestMQ{\EffDh}$ \\
cat & $\BestMR{\EffEa}$ & $\BestMS{\EffEb}$ & $\BestMT{\EffEc}$ & \cellcolor{MethodTint}$\BestMU{\EffEd}$ & $\BestMV{\EffEe}$ & $\BestMW{\EffEf}$ & $\BestMX{\EffEg}$ & $\BestMY{\EffEh}$ \\
airplane & $\BestMZ{\EffFa}$ & $\BestNA{\EffFb}$ & $\BestNB{\EffFc}$ & $\BestNC{\EffFd}$ & $\BestND{\EffFe}$ & $\BestNE{\EffFf}$ & \cellcolor{MethodTint}$\BestNF{\EffFg}$ & $\BestNG{\EffFh}$ \\
horse & $\BestNH{\EffGa}$ & $\BestNI{\EffGb}$ & $\BestNJ{\EffGc}$ & $\BestNK{\EffGd}$ & \cellcolor{MethodTint}$\BestNL{\EffGe}$ & $\BestNM{\EffGf}$ & $\BestNN{\EffGg}$ & $\BestNO{\EffGh}$ \\
train & $\BestNP{\EffHa}$ & \cellcolor{MethodTint}$\BestNQ{\EffHb}$ & $\BestNR{\EffHc}$ & $\BestNS{\EffHd}$ & $\BestNT{\EffHe}$ & $\BestNU{\EffHf}$ & $\BestNV{\EffHg}$ & $\BestNW{\EffHh}$ \\
\bottomrule
\end{tabular}

\end{table}

\subsection{Pairings}
\label{sec:pairings}

We now define the pairings we compare and the effect by which the selection half ranks them. Intuitively, a pairing takes one entry in each row and in each column of the table of label combinations. To rank pairings, we first measure how far each background lowers accuracy on each object label compared with no background at all. Let $a_m(f,b)$ be the accuracy of classifier $m$ on the \UnitsPerLabel composites of object label $f$ over background label $b$, and $a_m(f,\varnothing)$ its accuracy on the same foregrounds alone on grey. The background effect of $b$ on $f$ is the selection half's mean change,
\begin{equation}
  e(f,b) = \frac{1}{|S|}\sum_{m\in S}\bigl(a_m(f,b)-a_m(f,\varnothing)\bigr),
  \label{eq:effect}
\end{equation}
where $S$ is the selection half of \HalfSize classifiers. A negative effect means that the background lowers the label's accuracy below its accuracy on grey, and \cref{tab:effects} gives the effect of \cref{eq:effect} for every label combination on the first image set.

Intuitively, the hardest pairing puts each object label on a background that lowers its accuracy, while no two object labels share a background. A pairing's test set is its \LabelsPerSide label combinations, \PanelSize composites in which every object label and every background label has a frequency of one eighth. Precisely, a pairing $\pi$ gives each object label $f$ a different background label $\pi(f)$, and the hardest pairing minimizes the summed background effect of its label combinations,
\begin{equation}
  \pi^{\star} = \operatorname*{arg\,min}_{\pi}\sum_{f} e\bigl(f,\pi(f)\bigr).
  \label{eq:hardest}
\end{equation}
We find it by enumerating all \CandidatePairings pairings, and the linear assignment solver of the SciPy library~\cite{virtanen2020scipy}, which picks one entry in each row and each column with the smallest sum, returns the same pairing. The easiest pairing maximizes the same sum. Every pairing uses each object label once, so the hardest pairing is also the pairing on which the selection half's mean accuracy is lowest.

Random pairings give the baseline. A random pairing shows the same foregrounds and background crops as the hardest pairing at the same label frequencies, so the two differ in the assignment of labels alone. We draw \RandomPairings random pairings, each the hardest pairing of the table of background effects after a random shuffle of its background labels, which gives a uniform draw when the minimum is untied.

\subsection{The drop and its uncertainty}
\label{sec:measures}

Intuitively, the drop says how much accuracy a classifier gives up when the hardest pairing replaces a random one. A positive drop means lower accuracy on the hardest pairing. A classifier's accuracy on the test set of a pairing $\pi$ is $A_m(\pi)=\frac{1}{8}\sum_f a_m\bigl(f,\pi(f)\bigr)$, the mean of its accuracies on the label combinations of $\pi$. Its drop is its mean accuracy over the random pairings less its accuracy on the hardest pairing,
\begin{equation}
  d_m = \frac{1}{R}\sum_{r=1}^{R} A_m(\pi_r) - A_m(\pi^{\star}),
  \label{eq:drop}
\end{equation}
where $\pi_1,\ldots,\pi_R$ are the $R=\RandomPairings$ random pairings. We report the mean drop over the evaluation classifiers that pass the compositing check. Some later analyses take the drop against the mean accuracy over all \Cells label combinations instead of the random pairings. That mean equals the mean over all pairings, since every label combination appears in the same fraction of them, and the mean over the random pairings is an estimate of it.

For example, \ArchAb~\cite{howard2019searching}, the evaluation classifier of lowest ImageNet accuracy at \TopAb{}\%, scores \ExRefFirst on average over the random pairings of the first image set and \FirstEasyAaHard on the hardest pairing, a drop of \DropAbFirst. The drop also splits by object label. A label's drop $d_m(f)$ is its mean accuracy over the label combinations that the random pairings give it less its accuracy in the label combination that $\pi^{\star}$ gives it, and the drop is their mean, $d_m=\frac{1}{8}\sum_f d_m(f)$. A label's drop is a difference in accuracy for that label and not a fraction of the overall drop.

Intuitively, our uncertainty asks how much the drop would change with other photographs of the same labels, for the same \Models checkpoints. We therefore bootstrap over source images, with \BootReps replicates, since one image enters \LabelsPerSide composites. Each replicate draws, for every label, its \UnitsPerLabel foregrounds or background crops with replacement, and it weights each composite by the product of the draws of its foreground and its background crop. A label combination's accuracy is then a weighted mean, and a test set's accuracy is the mean over its label combinations. A lower bound is the value below which one twentieth of the replicates lie, interpolated by the default linear rule of the NumPy library~\cite{harris2020numpy}. An interval spans the middle nine tenths of the replicates. Unless we say otherwise, a bound keeps the pairing, the drawn image matches and the qualifying classifiers fixed, even on the first image set, whose images we used to choose it. On the first image set such a bound ignores the search over pairings, and \cref{sec:first} gives a bound that allows for it.

Intuitively, the hardest pairing transfers when the evaluation half is less accurate on it than on random pairings, and by more than chance would explain. Three things could make a drop look like transfer when it is not one. The drop could be too small to matter. The sampling of source images could produce it, so that other photographs would not repeat it. Many random pairings could lower the evaluation half's accuracy as much, so that the search over pairings added nothing. The drop of a random pairing is the mean accuracy over all random pairings less the accuracy on that pairing.

Formally, the drop of \cref{eq:drop} transfers on an image set when three conditions hold:
\begin{enumerate*}[label=(\roman*)]
\item the drop exceeds a margin of \Margin, the smallest drop that we count as transfer;
\item the lower bound of the drop exceeds the margin; and
\item fewer than \ShuffleLimit of the \RandomPairings random pairings have a drop no more than \WithinPoint below it.
\end{enumerate*}
Conditions (i) to (iii) check these three conditions on the observed data and sampled pairings.

Two further tests ask whether the hardest pairing changes which classifiers look better. Intuitively, if classifiers that do well on ImageNet also do well on the hardest pairing, their two accuracies stay close to a line on probit scales, while a pairing that reorders them scatters the points. The correlation test measures this by Pearson's correlation between the probits of ImageNet accuracy and of accuracy on a test set~\cite{miller2021accuracy}, over the evaluation classifiers that pass the compositing check. The test applies only where that correlation over all \Cells label combinations is at least \LineGate, so that there is a line for the hardest pairing to weaken. It finds weakening where the correlation decreases by more than \LineMargin from all label combinations to the hardest pairing, at the point estimate and at the lower bound of the decrease, whose bootstrap and probit clipping Appendix~\ref{sec:supp-composites} gives.

The second, our test for rank reversals, asks whether the hardest pairing reorders classifiers that are reliably ordered without it. Intuitively, a swap counts only when a classifier reliably ahead on all label combinations falls reliably behind on the hardest pairing, since point estimates alone can change order by chance. One classifier has a reliable lead over another when its lead, its accuracy less the other's, has a lower bound above zero on all \Cells label combinations. A reliable lead is reversed when the upper bound of the same lead on the hardest pairing lies below $-\ReversalMargin$. The test applies when at least \MinResolved pairs have a reliable lead, and it finds reordering where a share of at least \ShareMargin of them is reversed.

\subsection{New images and what we fixed in advance}
\label{sec:newset}

We repeat the comparison on images that no composite had used, since both halves scored the first image set and its images helped choose the pairing. The draw of the new image set excludes the \UsedImages images that earlier composites used. They are the \ExcludedFirstSet images of the first image set and \ExcludedEarlier COCO validation images, which no draw from the training split could select in any case. Those validation images had supplied separate images for choosing and for confirming in an earlier pilot with six object labels. The new image set takes \ImagesNew more COCO training images under the same licence by the same rules, \UnitsPerLabel to a label, so its \Foregrounds foregrounds and \Windows background crops are all new. The halves and pairings stay as the first image set fixed them, and both halves score every new image.

To compare the two image sets, we ask whether a classifier's drop changes from one to the other. We take each classifier's drop on the first image set less its drop on the new image set, and we average this signed difference over the classifiers that pass the new image set's compositing check. We bootstrap that average over those classifiers, so its interval measures variation across classifiers rather than across images.

We fixed our choices before we saw the scores that they concern, and every random draw took a recorded seed. We recorded the hardest pairing and its background effects before any classifier of the evaluation half was scored. At that point we also fixed the margin of \Margin, the bootstrap, the transfer criterion, both tests of the order of classifiers and the rule of \cref{sec:bylabel} for a background that raises accuracy. We fixed the tests on the new image set, the difference between the image sets among them, before any classifier scored its images.

Every other analysis we report is post hoc: we chose it after we saw both image sets' results, and we fixed its computation before it ran. The mark \aftertest names such an analysis wherever it appears, and \cref{tab:posthoc} in Appendix~\ref{sec:supp-after} lists them all. Before scoring each image set, we ran our scoring on constructed cases whose results were fixed by construction, \KnownFirst for the first image set and \KnownNew for the new one. \Cref{tab:known} in Appendix~\ref{sec:supp-after} lists each case with its result.

In all our tables, a subscript gives a lower bound or an interval from the bootstrap, and an arrow in a column heading points in the better direction. Bold shows the best value among those compared, and a tint shows the hardest pairing. Our tables give every accuracy and every difference of accuracies as a fraction of answers, while our figures give accuracies in percent and their differences in percentage points.

\section{Results}
\label{sec:results}

On new images, the hardest pairing lowers mean accuracy among the \AdmittedNew evaluation classifiers passing the compositing check by \DropNew against random pairings, and the drop meets our criterion for transfer. The test keeps two things apart from the choice of the pairing: the evaluation half, which took no part in it, and the new image set, whose photographs took no part either. \Cref{sec:first} measures the first of these alone, on the images used in the choice, and \cref{sec:new} measures both together.

\subsection{Transfer across classifiers}
\label{sec:first}

On the images used to choose it, the hardest pairing lowers the mean accuracy of \AdmittedFirst evaluation classifiers by \DropFirst, from \AccRandomFirst over the random pairings to \AccHardFirst. This comparison tests transfer across classifiers alone, since the evaluation half took no part in the choice but the photographs did. On the percentage scale, accuracy falls from \AbsRandomFirst{}\% to \AbsHardFirst{}\%, which is \AbsDropFirst percentage points or \AbsReductionFirst{}\% of its level over random pairings. These \AdmittedFirst are the evaluation classifiers, out of the evaluation half's \HalfSize, that pass the compositing check on the first image set. \Cref{tab:results} gives the same measures for them and for the \AdmittedNew that pass it on the new image set, and \cref{sec:line} returns to the correlations and the reliable leads that it lists.

\paragraph{With the pairing held fixed the drop transfers.}
Its lower bound of \DropFirstLow lies above the margin, and no random pairing has a drop within \WithinPoint of it. So the drop transfers by the criterion of \cref{sec:measures}, as long as we treat the pairing as fixed. The easiest pairing raises the same classifiers' accuracy above the random pairings on both image sets, so the pairing of labels moves accuracy in both directions.

\begin{table}[t]
  \caption{\textbf{The hardest pairing lowers mean accuracy on both image sets, and the test for rank reversals finds no reliable lead reversed.} Columns cover evaluation classifiers passing each set's compositing check, with \RandomPairings random pairings as the baseline. Rows give mean accuracy, the drop, the Pearson correlation between probit accuracy and probit ImageNet accuracy, and counts of pairs with a reliable lead. Subscripts give the middle nine tenths of \BootReps source image resamples with checkpoints fixed for accuracy and resampled for correlation differences, $\uparrow$ the better direction, bold each block's highest value, tint the hardest pairing and $\ddagger$ a post hoc row.}
  \label{tab:results}
  \centering
  \small
  \setlength{\tabcolsep}{4pt}
  \setlength{\tabcolsep}{2.5pt}
  \begin{tabular}{@{}lcc@{}}
\toprule
\multicolumn{3}{@{}l}{Accuracies and drops are fractions of answers.} \\
& First image set & New image set \\
& \AdmittedFirst classifiers & \AdmittedNew classifiers \\
\midrule
\multicolumn{3}{@{}l}{\emph{Mean accuracy $\uparrow$}} \\
All \Cells combinations & \BestXC{\AccGridFirst}\iv{\AccGridFirstLo to \AccGridFirstHi} & \BestXD{\AccGridNew}\iv{\AccGridNewLo to \AccGridNewHi} \\
Random pairings & \BestXE{\AccRandomFirst}\iv{\AccRandomFirstLo to \AccRandomFirstHi} & \BestXF{\AccRandomNew}\iv{\AccRandomNewLo to \AccRandomNewHi} \\
\rowcolor{MethodTint}Hardest pairing & \BestXG{\AccHardFirst}\iv{\AccHardFirstLo to \AccHardFirstHi} & \BestXH{\AccHardNew}\iv{\AccHardNewLo to \AccHardNewHi} \\
Easiest pairing & \BestXI{\AccEasyFirst}\iv{\AccEasyFirstLo to \AccEasyFirstHi} & \BestXJ{\AccEasyNew}\iv{\AccEasyNewLo to \AccEasyNewHi} \\
\midrule
Drop on the hardest pairing & \DropFirst\iv{\DropFirstLow to \DropFirstHigh} & \DropNew\iv{\DropNewLow to \DropNewHigh} \\
\midrule
\multicolumn{3}{@{}l}{\emph{Correlation with ImageNet accuracy $\uparrow$}} \\
All \Cells combinations & \BestXK{\LineGridFirst} & \BestXL{\LineGridNew} \\
\rowcolor{MethodTint}Hardest pairing & \BestXM{\LineHardFirst} & \BestXN{\LineHardNew} \\
Easiest pairing & \LineEasyFirst & \LineEasyNew \\
All combinations less hardest\textsuperscript{\ddag} & $\LineDecreaseFirst$ & $\LineDecreaseNew$ \\
Its lower and upper ends & $\LineDecreaseFirstLow$, $\LineDecreaseFirstHigh$ & $\LineDecreaseNewLow$, \LineDecreaseNewHigh \\
All combinations less easiest\textsuperscript{\ddag} & \EasyGapFirst & \EasyGapNew \\
Its lower and upper ends & \EasyGapFirstLo, \EasyGapFirstHi & \EasyGapNewLo, \EasyGapNewHi \\
\midrule
Pairs with a reliable lead & \ResolvedFirst of \PairsFirst & \ResolvedNew of \PairsNew \\
\quad reversed on the hardest pairing & \ReversedFirst & \ReversedNew \\
\bottomrule
\end{tabular}

\end{table}

\paragraph{A bound counting the search misses the margin.}
We chose the pairing from all \CandidatePairings pairings on these same images, and the bound above ignores that search. Intuitively, the best of many candidates looks better than it is, so a fair bound must allow for the search. In each resample, a pairing's shortfall is its drop on the image set less its drop in that resample. A simultaneous lower bound uses all pairings to allow for the search: it subtracts from each drop the ninety fifth percentile across resamples of the largest shortfall of any pairing, as Appendix~\ref{sec:supp-selection} defines. It is \CompactFirstSampledSimultaneous against the random pairings and \CompactFirstUniformSimultaneous against the mean over all \Cells label combinations, both below the margin. So the evidence of transfer rests on the new images of \cref{sec:new}.

\begin{figure}[t]
  \centering
  \includegraphics[width=\linewidth]{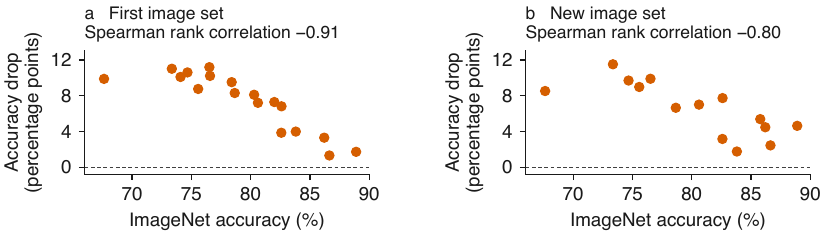}
  \caption{\textbf{Every evaluation classifier that passes the compositing check has a positive estimated drop on both image sets, and the largest drops come at low ImageNet accuracy.}\aftertest For the \AdmittedFirst and \AdmittedNew qualifying classifiers, each point plots the drop from the random pairing mean against its ImageNet accuracy, and the dashed line marks a drop of zero. The mark $\ddagger$ names a post hoc analysis.}
  \label{fig:ranking}
\end{figure}

\paragraph{Every qualifying classifier has a positive estimated drop.}
On both image sets, every evaluation classifier that passes the compositing check has a positive drop at its point estimate, as \cref{fig:ranking} shows. For \BelowZeroFirst of these classifiers on each image set, the interval over source images extends below zero, and \cref{tab:checkpoints} in Appendix~\ref{sec:supp-classifiers} marks them. The classifiers weakest on ImageNet have the largest drops, and the Spearman rank correlation between drop and ImageNet accuracy is $\RankFirst$ on the first image set and $\RankNew$ on the new one.

\begin{table}[t]
  \caption{\textbf{Six robustness analyses leave the drop on new images above the margin.}\aftertest Each drop is mean accuracy over all \Cells label combinations less accuracy on the hardest pairing, except in the last block, where the hardest random pairing takes the hardest pairing's place. Each difference is a row's drop less the original drop, or in the last block the hardest pairing's drop less that row's. Subscripts give lower bounds or intervals, $\uparrow$ the better direction, \S{} the row that uses random pairings and the \EdgeModels classifiers that pass the compositing check, and $\ddagger$ a post hoc analysis.}
  \label{tab:checks}
  \centering
  \small
  \setlength{\tabcolsep}{2.5pt}
  \begin{tabular}{@{}lcc@{}}
\toprule
\multicolumn{3}{@{}l}{Drops and differences are changes in fraction correct.} \\
Analysis & Drop $\uparrow$ & Difference \\
\midrule
\multicolumn{3}{@{}l}{\emph{Groups of classifiers fixed in advance}} \\
All \FourthAllModels evaluation classifiers & \CompactAllUniformDrop\iv{\CompactAllUniformLower} & \\
The \FourthFixedModels that pass the first image set's check & \CompactFixedUniformDrop\iv{\CompactFixedUniformLower} & \\
\midrule
\multicolumn{3}{@{}l}{\emph{New matches, new photographs and new mask edges}} \\
New matches within label combinations & \RerunUniformDrop\iv{\RerunUniformLower} & \RerunChange\iv{$[\RerunChangeLo,\RerunChangeHi]$} \\
Third image set & \PoolHardDrop\iv{\PoolHardLower} & \\
Blurred mask & \BlurDrop\iv{\BlurLower} & $\BlurChange$\iv{$[\BlurChangeLo,\BlurChangeHi]$} \\
Eroded mask\textsuperscript{\S} & \EdgeTreated\iv{\EdgeTreatedLo} & \EdgeDifference\iv{$[\EdgeDifferenceLo,\EdgeDifferenceHi]$} \\
\midrule
\multicolumn{3}{@{}l}{\emph{Hardest random pairing in place of the hardest pairing}} \\
On the new image set & \ComparatorDrop\iv{\ComparatorLower} & \ComparatorDelta\iv{$[\ComparatorDeltaLo,\ComparatorDeltaHi]$} \\
On the third image set & \PoolCompDrop\iv{\PoolCompLower} & \PoolDelta\iv{$[\PoolDeltaLo,\PoolDeltaHi]$} \\
\bottomrule
\end{tabular}

\end{table}

\subsection{Transfer to new images}
\label{sec:new}

On new images, the hardest pairing lowers mean accuracy among the \AdmittedNew evaluation classifiers passing the compositing check by \DropNew, from \AccRandomNew over the random pairings to \AccHardNew, with a lower bound of \DropNewLow above the margin. This comparison tests both kinds of transfer at once, since neither the evaluation half nor the new photographs took part in the choice. On the percentage scale of the abstract, accuracy falls from \AbsRandomNew{}\% to \AbsHardNew{}\%, which is \AbsDropNew percentage points or \AbsReductionNew{}\% of its level over random pairings. The mean covers the \AdmittedNew of the \HalfSize evaluation classifiers that pass the compositing check on the new image set. Only two of the \RandomPairings random pairings have a drop within \WithinPoint of the hardest pairing's, so the drop transfers on images that took no part in the choice. The selection half, which fixed the pairing on the first image set, loses as much on these new images, \SelDropNew over the \SelAdmittedNew of its classifiers that pass the check.

\begin{figure}[!t]
  \centering
  \includegraphics{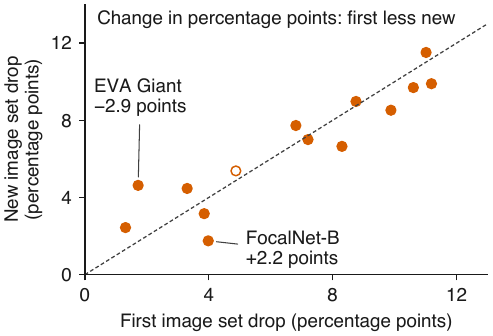}
  \caption{\textbf{The two classifiers whose drops change most between the image sets change in opposite directions.}\aftertest For the \AdmittedNew classifiers passing the new image set's check, each point plots the drop from the random pairing mean on the new image set against the drop on the first, and the dashed line marks equal drops. A labelled change is the first image set's drop less the new image set's, and the open circle marks a classifier that the first image set's compositing check excludes. The mean absolute change is \MeanAbsoluteChangePoints percentage points, and $\ddagger$ names a post hoc analysis.}
  \label{fig:sets}
\end{figure}

\paragraph{The two observed mean drops lie close.}
A pairing that fitted the photographs used to choose it could lose part of its drop on new photographs. Over the \PairedModels classifiers that pass the new image set's compositing check, a classifier's drop on the first image set less its drop on the new one averages \PairedMean. Its interval over resamples of those classifiers runs from $\PairedModelLo$ to \PairedModelHi. The absolute change of a single classifier's drop averages \AbsChangeNew, and the two largest changes go in opposite directions, as \cref{fig:sets} shows.

\paragraph{Six robustness analyses leave the drop in place.}
In every row of \cref{tab:checks} above the last block, the drop on new images and its lower bound exceed the margin, and every reported interval of a change from the original includes zero. These analyses ask whether the drop depends on which classifiers enter, on which foreground meets which background crop, on which photographs we drew, or on the edge of the pasted object.

The first analysis fixes the group of classifiers before the new images were scored, either as the whole evaluation half or as the \FourthFixedModels that pass the first image set's check. The new image set's own check then excludes none of them. Others give each foreground a new background crop within every label combination, or draw a third image set of \PoolImages photographs that shares no image with any earlier composite. Still others alter the edge of the pasted object: a Gaussian blur of its mask at a standard deviation of \BlurSigma pixels changes the profile of the boundary, and an erosion of \EdgeRadius pixel changes the support of the mask. All \FourthAllModels evaluation classifiers enter each analysis after the first, except the erosion, which uses the \EdgeModels that pass the new image set's check.

\paragraph{The hardest random pairing also exceeds the margin.}
The sixth analysis asks how much the full search adds over a cheap one. It replaces the search over all pairings by the hardest random pairing, the one of the \RandomPairings random pairings on which the selection half scores lowest on the first image set. That pairing's lower bound also exceeds the margin on the new and third image sets. The hardest pairing lowers accuracy by \ComparatorDelta more on the new image set and by \PoolDelta more on the third, each with an interval that includes zero. So neither image set separates the full search from the hardest random pairing.

\paragraph{The third image set also shows the drop.}
On the third image set the hardest pairing again lowers accuracy, with a drop of \PoolHardSampledDrop against the random pairings and a lower bound of \PoolHardSampledLower. These photographs give another comparison because none of them entered an earlier composite, so they repeat the test of the new image set on a second draw of new photographs. The third image set is drawn by the new image set's rules from the licensed training images, excluding the \PoolUsed images that earlier composites used. Each label draws \UnitsPerLabel of its unused candidates by seed \SeedPool, the smallest label having \PoolSmallest candidates, so the set has \PoolForegrounds foregrounds and \PoolWindows background crops from \PoolImages images. It keeps the permutations of both earlier image sets within each label combination, and the number of random pairings with a drop within \WithinPoint of the hardest pairing's is \PoolWithin.

We fixed the hardest pairing and the hardest random pairing before the third image set was drawn, and we took their drops against the mean over all \Cells label combinations. We paired their difference within each of \BootReps resamples of the set's source images, drawn by seed \SeedUnit. On the third image set the hardest pairing has a drop of \PoolHardDrop and the hardest random pairing one of \PoolCompDrop, a difference of \PoolDelta whose interval runs from $\PoolDeltaLo$ to \PoolDeltaHi. That interval includes zero, so the third image set cannot tell whether the full search adds to the drop of the hardest random pairing, and it allows an extra drop as large as \PoolDeltaHi. \Cref{tab:pools} sets these drops beside the new image set and the blurred composites of \cref{sec:supp-edge}.

\begin{table}[!t]
  \caption{\textbf{On the third image set the hardest pairing keeps its drop, and its extra drop over the hardest random pairing has an interval that includes zero.}\aftertest Columns give all \PoolModels evaluation classifiers on three sets of composites, and every drop is taken against all \Cells label combinations. Subscripts give lower bounds or intervals, $\uparrow$ the better direction, tint the hardest pairing and $\ddagger$ a post hoc analysis.}
  \label{tab:pools}
  \centering
  \small
  \begin{tabular}{@{}lccc@{}}
\toprule
\multicolumn{4}{@{}l}{The upper rows count images. Accuracy drops are fractions of answers.} \\
& New image set & Third image set & \shortstack{New image set,\\blurred mask} \\
\midrule
Source images & \ImagesNew & \PoolImages & \ImagesNew \\
Foregrounds and background crops & \Foregrounds and \Windows & \PoolForegrounds and \PoolWindows & \Foregrounds and \Windows \\
\midrule
\multicolumn{4}{@{}l}{\emph{Drop against all \Cells label combinations $\uparrow$}} \\
\rowcolor{MethodTint}Hardest pairing & \CompactAllUniformDrop\iv{\CompactAllUniformLower} & \PoolHardDrop\iv{\PoolHardLower} & \BlurDrop\iv{\BlurLower} \\
Hardest random pairing & \ComparatorDrop\iv{\ComparatorLower} & \PoolCompDrop\iv{\PoolCompLower} & \\
\midrule
Hardest less hardest random & \ComparatorDelta\iv{$[\ComparatorDeltaLo,\ComparatorDeltaHi]$} & \PoolDelta\iv{$[\PoolDeltaLo,\PoolDeltaHi]$} & \\
Blurred less original & & & $\BlurChange$\iv{$[\BlurChangeLo,\BlurChangeHi]$} \\
\bottomrule
\end{tabular}

\end{table}

\paragraph{The drop persists where cards report \INonek alone.}
The evaluation half includes classifiers whose cards report \INonek alone and others whose cards report larger collections, so the drop could come from the extra data alone. Model cards, the documents released with each checkpoint, state what data trained it. For the \GroupInOneK classifiers whose cards state \INonek alone, the drop against all \Cells label combinations is \GroupInOneKDropNew on the new image set and \GroupInOneKDropThird on the third, each with a lower bound above the margin. The \GroupFurther classifiers whose cards report further training data have smaller drops, \GroupFurtherDropNew and \GroupFurtherDropThird, as \cref{tab:groups} shows. The first image set shows the same pattern, and the difference between the two groups has an interval above zero on all three image sets.

\begin{table}[!t]
  \caption{\textbf{The drop persists among the evaluation classifiers whose cards report \INonek alone.}\aftertest Rows group the \HalfSize evaluation classifiers by the training data their model cards state, with drops against the mean over all \Cells label combinations, and the last row is the difference between two groups. Our tests need at least \GroupFloor classifiers in a group, so the table reports the drops of the smaller groups without testing them. Subscripts give lower bounds or intervals, $\uparrow$ the better direction and $\ddagger$ a post hoc analysis.}
  \label{tab:groups}
  \centering
  \small
  \setlength{\tabcolsep}{5pt}
\begin{tabular}{@{}p{0.29\textwidth}rccc@{}}
\toprule
\multicolumn{5}{@{}l}{Image set columns give accuracy drops as fractions of answers.} \\
Training data the card states & Classifiers & \shortstack{First\\image set $\uparrow$} & \shortstack{New\\image set $\uparrow$} & \shortstack{Third\\image set $\uparrow$} \\
\midrule
\INonek alone & \GroupInOneK & \GroupInOneKDropFirst\iv{\GroupInOneKLowerFirst} & \GroupInOneKDropNew\iv{\GroupInOneKLowerNew} & \GroupInOneKDropThird\iv{\GroupInOneKLowerThird} \\
\quad no teacher named & \GroupNoTeacher & \GroupNoTeacherDropFirst\iv{\GroupNoTeacherLowerFirst} & \GroupNoTeacherDropNew\iv{\GroupNoTeacherLowerNew} & \GroupNoTeacherDropThird\iv{\GroupNoTeacherLowerThird} \\
\quad distillation stated & \GroupDistilled & \GroupDistilledDropFirst\iv{\GroupDistilledLowerFirst} & \GroupDistilledDropNew\iv{\GroupDistilledLowerNew} & \GroupDistilledDropThird\iv{\GroupDistilledLowerThird} \\
Further training data & \GroupFurther & \GroupFurtherDropFirst\iv{\GroupFurtherLowerFirst} & \GroupFurtherDropNew\iv{\GroupFurtherLowerNew} & \GroupFurtherDropThird\iv{\GroupFurtherLowerThird} \\
\midrule
\INonek alone less further data & & \GroupGapFirst\iv{$[\GroupGapFirstLo,\GroupGapFirstHi]$} & \GroupGapNew\iv{$[\GroupGapNewLo,\GroupGapNewHi]$} & \GroupGapThird\iv{$[\GroupGapThirdLo,\GroupGapThirdHi]$} \\
\bottomrule
\end{tabular}

\end{table}

\subsection{Labels and backgrounds}
\label{sec:bylabel}

\begin{figure}[!t]
  \centering
  \includegraphics[width=\linewidth]{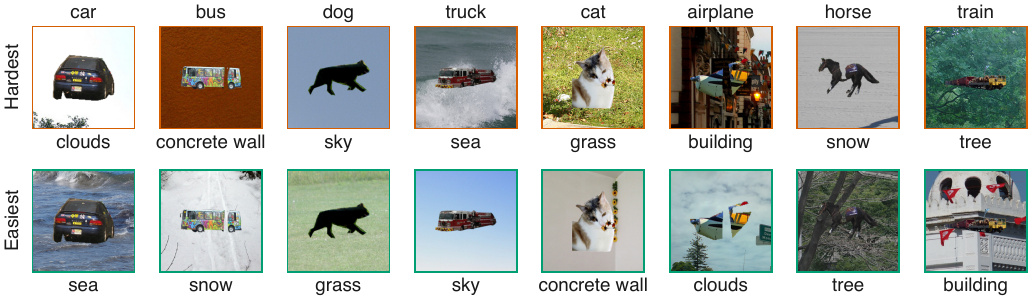}
  \caption{\textbf{The hardest and easiest pairings place the same foregrounds on different backgrounds.} Each column shows one object label's first foreground of the new image set on its background in the hardest pairing, framed in vermillion, above its background in the easiest pairing, framed in green.}
  \label{fig:gallery}
\end{figure}

Four labels account for most of the drop, and two label combinations, airplane on building and truck on sea, account for the largest label drops on both image sets. This analysis asks whether the drop is spread over all labels or comes from a few, which tells a benchmark where its difficulty lies. The hardest pairing puts car on clouds, bus on concrete wall, dog on sky, truck on sea, cat on grass, airplane on building, horse on snow and train on tree, as \cref{fig:gallery} shows for the first foreground of each label in the new image set. Its summed background effect is $\HardSum$, against $\EasySum$ for the easiest pairing.

\begin{table}[t]
  \caption{\textbf{Four object labels account for most of the drop on both image sets.}\aftertest Each row gives an object label, its background in the hardest pairing, the selection half's first image set effect against grey, and the drop against random pairings for the \AdmittedFirst and \AdmittedNew evaluation classifiers passing the respective checks. The overall drop is the mean of the label drops. Arrows mark the direction that the hardest pairing favors, bold the extremes, tint the hardest pairing and $\ddagger$ a post hoc analysis.}
  \label{tab:labels}
  \centering
  \small
  \setlength{\tabcolsep}{4.8pt}
  \begin{tabular}{@{}llrrr@{}}
\toprule
\multicolumn{5}{@{}l}{Background effects and label drops are differences in fraction correct.} \\
Object & Background & Effect $\downarrow$ & \shortstack{Drop, first\\image set $\uparrow$} & \shortstack{Drop, new\\image set $\uparrow$} \\
\midrule
\rowcolor{MethodTint}car & clouds & $\BestNX{\EffAh}$ & $\BestNY{\LabAFirst}$ & $\BestNZ{\LabANew}$ \\
\rowcolor{MethodTint}bus & concrete wall & $\BestOA{\EffBc}$ & $\BestOB{\LabBFirst}$ & $\BestOC{\LabBNew}$ \\
\rowcolor{MethodTint}dog & sky & $\BestOD{\EffCa}$ & $\BestOE{\LabCFirst}$ & $\BestOF{\LabCNew}$ \\
\rowcolor{MethodTint}truck & sea & $\BestOG{\EffDf}$ & $\BestOH{\LabDFirst}$ & $\BestOI{\LabDNew}$ \\
\rowcolor{MethodTint}cat & grass & $\BestOJ{\EffEd}$ & $\BestOK{\LabEFirst}$ & $\BestOL{\LabENew}$ \\
\rowcolor{MethodTint}airplane & building & $\BestOM{\EffFg}$ & $\BestON{\LabFFirst}$ & $\BestOO{\LabFNew}$ \\
\rowcolor{MethodTint}horse & snow & $\BestOP{\EffGe}$ & $\BestOQ{\LabGFirst}$ & $\BestOR{\LabGNew}$ \\
\rowcolor{MethodTint}train & tree & $\BestOS{\EffHb}$ & $\BestOT{\LabHFirst}$ & $\BestOU{\LabHNew}$ \\
\midrule
Mean over labels & inapplicable & inapplicable & \DropFirst & \DropNew \\
\bottomrule
\end{tabular}

\end{table}

\paragraph{Four labels account for most of the drop.}
Airplane, truck, cat and bus have the four largest label drops in \cref{tab:labels}, and airplane on building has the largest on both image sets, \LabFFirst and \LabFNew. The table sets each label's drop beside the background effect of its label combination on the first image set. Every effect in the table is negative, so each label's background in the hardest pairing lowers its accuracy below grey. Car on clouds still has a label drop below zero on both image sets, $\LabAFirst$ and $\LabANew$. The hardest pairing minimizes a sum over all eight labels, which can leave one label better off than on its random backgrounds.

\paragraph{Two label combinations give the largest label drops.}
Airplane on building and truck on sea have the two largest label drops on both image sets, and all \SplitCount splits of the classifiers select both, as \cref{tab:splits} in Appendix~\ref{sec:supp-after} shows. To ask whether the other labels lower accuracy on their own, we removed the object labels airplane and truck and the background labels building and sea. Chosen again on the first image set, the hardest of the \RemovalPairings remaining pairings is the six label combinations that the hardest pairing already gives the remaining labels. On the new and third image sets, this pairing lowers the evaluation half's accuracy by \RemovalDropNew and \RemovalDropThird against the mean of the \RemovalCells remaining label combinations. Both lower bounds lie above zero but below the margin, too low to count as transfer (\cref{tab:removal}). The two removed combinations also differ in how often their labels occur together in COCO, since COCO shows a building in \CoocFg of its images of airplanes and the sea in \CoocDf of its images of trucks.

\begin{table}[!t]
  \caption{\textbf{Without airplane, truck, building and sea, the hardest remaining pairing lowers accuracy by less than the margin of \Margin{} at its lower bound.}\aftertest Columns give all \HalfSize evaluation classifiers on each image set, still choosing among all \LabelsPerSide object labels. The remaining pairing is chosen on the first image set, and its drop is taken against the mean of the \RemovalCells remaining label combinations. Subscripts give lower bounds, $\uparrow$ the better direction and $\ddagger$ a post hoc analysis.}
  \label{tab:removal}
  \centering
  \small
  \begin{tabular}{@{}lccc@{}}
\toprule
\multicolumn{4}{@{}l}{Accuracy and drops are fractions of answers.} \\
& \shortstack{First\\image set} & \shortstack{New\\image set} & \shortstack{Third\\image set} \\
\midrule
Mean accuracy over the \RemovalCells remaining combinations & \RemovalBaseFirst & \RemovalBaseNew & \RemovalBaseThird \\
Accuracy on the hardest remaining pairing & \RemovalAccFirst & \RemovalAccNew & \RemovalAccThird \\
\midrule
Drop $\uparrow$ & \RemovalDropFirst\iv{\RemovalLowerFirst} & \RemovalDropNew\iv{\RemovalLowerNew} & \RemovalDropThird\iv{\RemovalLowerThird} \\
\bottomrule
\end{tabular}

\end{table}

\paragraph{Other splits choose similar pairings.}
Our split of the classifiers is one of many, so we drew \SplitCount splits by the stratified rule of \cref{sec:halves}. The hardest pairing of each selection half lowers the other half's accuracy on the new image set by \SplitDropMin to \SplitDropMax, and every lower bound exceeds the margin. Of those pairings, \SplitSame equal ours.

\paragraph{No background effect's lower bound exceeds \CueMargin.}
We fixed this threshold in advance to detect a background that raises accuracy, and no label combination of the first image set exceeds it at its lower bound. The largest effect is \EffMax, among \EffPositive label combinations above zero at the point estimate, while the largest lower bound reaches only \EffLowMax. So the hardest pairing selects backgrounds with negative point estimates in the selection half.

\paragraph{A pairing of labels transfers more than a matching transferred by image position.}
A pairing chooses at the level of labels, and a search could instead choose at the level of single images. An image matching does so: it assigns each foreground of the first image set to one background crop, using each crop once, so as to minimize the selection half's accuracy. It is an assignment of whole images, not a correspondence between points of two images. Of its \PanelSize matched composites, \MatchInside lie in the hardest pairing's label combinations. On the new image set it lowers accuracy less than the hardest pairing does. For the same \EdgeModels classifiers, the hardest pairing's drop exceeds the matching's by \MatchDeltaNew, with an interval from \MatchDeltaNewLo to \MatchDeltaNewHi.

\subsection{Order of classifiers}
\label{sec:line}

Neither test meets its criterion for a change in classifier order. The two tests ask whether a benchmark built on the hardest pairing would rank classifiers differently from one built on all label combinations, and they ask it in different ways. The correlation test asks whether accuracy on the hardest pairing still follows ImageNet accuracy across qualifying classifiers, and the test for rank reversals asks whether any reliably ordered pair crosses the reversal margin.

On the new image set, the correlation with ImageNet accuracy is \LineHardNew on the hardest pairing against \LineGridNew over all \Cells label combinations. The interval of the decrease runs from $\LineDecreaseNewLow$ to \LineDecreaseNewHigh and includes zero, so the correlation test finds no weakening. On the first image set the correlation over all label combinations, \LineGridFirst, lies below the \LineGate that the test needs, so we do not run it there. The hardest pairing's correlation there, \LineHardFirst, is still higher, and the interval of the decrease excludes zero. The easiest pairing's correlation is lower than that over all label combinations on both image sets, with both intervals above zero, as \cref{tab:results} gives and \cref{fig:line} in Appendix~\ref{sec:supp-after} shows for the new image set.

Some pairs of classifiers do change order at their point estimates, but none of these changes meets the test for rank reversals. Of the \ResolvedFirst pairs with a reliable lead on the first image set and the \ResolvedNew on the new one, the test counts none as reversed. Across both image sets, \PointBeyondBoth of these pairs change the order of their point estimates by more than \ReversalMargin, and one on each image set has an interval wholly below zero. None of them has an upper bound below $-\ReversalMargin$, so none meets the reversal criterion, and \cref{tab:pairs} in Appendix~\ref{sec:supp-after} lists each pair.

\section{Discussion}
\label{sec:discussion}

For one group of \Models classifiers and one way of building composites, the fixed pairing lowers mean accuracy on both image sets. Both halves scored the first image set, but the new image set shares no COCO image identifier with it. It keeps the labels, the halves, the pairings and the way we build composites, so the drop repeats on different source images for the pairing chosen on the first image set. With the split into halves, the new image set tests transfer across classifiers and across source images, while the way we build composites and any shared training data or distillation teachers stay fixed. The drop is present among classifiers whose cards report \INonek alone.

The label analysis and the order of classifiers tell a benchmark how to read the drop of a held-out classifier in this pool. Most of the drop comes from airplane on building and truck on sea. Without those labels, the six remaining label combinations still lower accuracy, but the lower bound of their drop lies below the margin, too low to count as transfer (\cref{sec:bylabel}). So the remaining pairing misses the transfer margin at its lower bound. The drop is largest for the classifiers weakest on ImageNet (\cref{fig:ranking}), and no pair meets the reversal criterion (\cref{sec:line}). So the hardest pairing reverses no reliable lead by that criterion, and what it adds is the size of each classifier's drop.

\section{Conclusion}
\label{sec:conclusion}

The pairing that one half of \Models classifiers found hardest lowers the other half's mean accuracy by \DropFirst on the composites used to choose it and by \DropNew on composites of new images. Each mean covers the evaluation classifiers that pass the compositing check on those composites. The evidence of transfer rests on the second drop, since we searched all pairings on the composites that gave the first, and a bound that allows for that search lies below the margin. A benchmark of composites built this way can therefore score its \Cells label combinations and the same objects alone on grey once, and fix its hardest pairing by \cref{eq:hardest}. It can then report each held-out classifier's drop by \cref{eq:drop} beside its accuracy over random pairings. That use rests on composites, and natural photographs of the same labels lie outside it.

\clearpage
\appendix
\section{The composites in detail}
\label{sec:supp-composites}

We build every composite from stored layers by fixed rules, so that every classifier scores the same images of an image set. Under the foreground rule of \cref{sec:composites}, an instance is a usable foreground when its annotation does not mark a crowd, its box keeps two pixels from every image edge, its longer box side is at least \MinFgLong pixels and its mask fills at least \MinMaskFill of its box. Each image gives at most one foreground, the usable instance with the most pixels in its decoded mask, with ties going to the lower annotation identifier. Under the rule for background crops, a crop's side is at least \BgMinSide pixels, it shrinks in steps of \BgSideStep pixels and its corners lie on a grid of \BgStride pixels, and background crops come only from images that give no foreground. COCO-Stuff names three of the background labels \texttt{sky-other}, \texttt{wall-concrete} and \texttt{building-other}, which we write as sky, concrete wall and building. The labels gather different numbers of ImageNet classes, car \ClassesA, bus \ClassesB, dog \ClassesC, truck \ClassesD and cat \ClassesE, while airplane, horse and train each map to a single class.

The resized foregrounds, masks and background crops are the layers of an image set. We resize these layers once and store them as arrays of bytes, and a composite blends a stored foreground over a stored background crop through its mask in integer arithmetic. Intuitively, each pixel of a composite is a weighted average of foreground and background, with the mask as the weight. Precisely, a composite's pixel takes the integer part of $(a u + (255 - a) v + 127)/255$, with $u$ and $v$ the levels of the foreground and the background crop and $a$ the level of the mask, each from $0$ to $255$. Every classifier scores batches of \ScoreBatch images in deterministic single precision with no reduced precision in its matrix products, in evaluation mode and without gradients.

The rules leave many candidates, from which each label draws its images at random. The foreground rule examined \FgCandidates instances of the mapped labels and included \FgAdmitted instances. The \FgSources images containing an included instance each supplied their largest usable instance as a foreground candidate. The rule for background crops searched the \BgSearched images under the licence that give no foreground, examined \BgCandidates pairs of an image and a background label and found a background crop for \BgAdmitted of them. Each label drew its \UnitsPerLabel images by a seeded permutation of its candidates sorted by image identifier. An image gives at most one background crop, for the selected label of its largest square, ties going to the purer square and then to the lower label.

The masks come from COCO's polygons, and we checked that our stored masks do not depend on the version of the library that drew them. A foreground's mask is rasterized at the photograph's size by the COCO library pycocotools, which converts each polygon to run lengths, merges their union and decodes a binary mask before the bilinear resize. The environment we recorded uses version \RasterVersion, and the version that first rasterized the masks was not recorded. A complete rebuild of the first image set's layers in the recorded environment equals the stored layers level for level, all \RebuildMasks masks among them, so the first version returned the levels the recorded one returns\aftertest.

A few conventions fix every pixel of a composite. Fractional resized dimensions and offsets use Python's rounding to the nearest integer, with ties to the even integer. The crop of the original surroundings uses the same rounding for its corners. The rule for background crops counts every annotated instance as an object, and it decodes a crowd region from its run lengths. A background crop keeps the purest square of its largest side, with ties going to the smaller row and then the smaller column.

The correlation test needs probits of accuracies that can reach zero or one, and its decrease needs an interval. The probit of an accuracy takes the accuracy clipped to the interval from $1/(2n)$ to $1-1/(2n)$, with $n$ the number of composites in the test set, \CompositesAll for all \Cells label combinations and \PanelSize for a pairing, and it takes ImageNet accuracy unclipped. The correlation test of \cref{sec:measures} draws classifiers as well as images. Each of \BootReps source image replicates also draws the classifiers that pass the compositing check with replacement, using the same drawn classifiers and image weights for both correlations. Subtracting the second correlation from the first gives the decrease, whose lower bound is the fifth percentile of these coupled differences and whose upper end is the ninety fifth. An undefined correlation causes the draw of classifiers to be repeated, and the bootstrap that bounds the paired difference of \cref{sec:newset} draws the classifiers that pass the new image set's check with replacement, \BootReps times.

Shuffling the background labels before the search turns the hardest pairing into a random one. Each random pairing shuffles the background columns of the table of background effects by a seeded permutation, takes the first minimizing pairing in lexicographic order on the shuffled table and takes its column positions as true background labels. With an untied minimum, that pairing is the hardest pairing composed with a uniform permutation, so it is itself a uniform draw.

The seeds and orders below let a reader draw the same images again. Every draw comes from the default generator of NumPy. Seed \SeedUnitsSecond samples images within the first image set's initial labels in the order car, bus, dog, truck, cat, airplane, then sky, tree, concrete wall, grass, snow, building. Seed \SeedUnitsAdded retains the images sampled for those labels and samples images within the added labels in the order horse, train, sea, clouds. Seed \SeedCells draws the permutations within label combinations in object label order, visiting every background label in its table order before the next object label. The new image set's images use \SeedNewUnits in the same table order, all object labels before all background labels, and they keep the first image set's permutations within label combinations.

Every other random step has its own seed. The draw of the classifiers takes the seeds \SeedFamilies and \SeedBins, and the halves \SeedSplit. The bootstraps over source images take \SeedEffect for the background effects and \SeedUnit for the drops, the correlation test's draw of classifiers takes \SeedModels, the random pairings take \SeedRandom and the bootstrap over classifiers of the paired difference takes \SeedPaired. The post hoc analyses take \SeedPairing for pairing the two image sets' replicates, \SeedSplits for the other splits of the classifiers, \SeedRerun for the new permutations within label combinations and \SeedPool for the third image set's images.

\section{The label combinations}
\label{sec:supp-cells}

Backgrounds differ from one another for a label even where every background lowers its accuracy. The hardest pairing must give every label a different background, so some of its label combinations in \cref{tab:effects} have the most negative effects of their rows and others do not. The airplane row shows the most negative effects and the rows of dog and train the least negative. \Cref{tab:cooccurrence} counts how often each label's training images also contain each background.

To compare backgrounds within one object label, we subtract the label's mean background effect from the effect of each of its label combinations. We mark a label combination where the lower bound of this difference exceeds \RelMargin, a criterion we fixed with the new image set's tests, after the first image set's scores were seen and before the new image set's were. \Cref{tab:relative} gives every difference with its lower bound and underlines the marked label combinations, \RelNamedFirst on the first image set and \RelNamedNew on the new one. On both image sets \RelNamedBoth are marked: cat on concrete wall, airplane on sky and airplane on clouds. Dog, whose label gathers \ClassesC ImageNet classes, has background effects near zero in every label combination. On the new image set the easiest pairing raises accuracy by \GainNew over the random pairings, with a lower bound of \GainNewLow.

\begin{table}[!t]
  \caption{\textbf{The hardest pairing joins label combinations common in COCO photographs with combinations seldom seen there.} Each entry is the share of COCO 2017 training images showing the row's object that also contain a pixel of the column's background, counted over all training images rather than the licensed ones, beside each row's count of images. Arrows point to larger shares, bold marks each column's largest share and tint the hardest pairing.}
  \label{tab:cooccurrence}
  \centering
  \small
  \begin{tabular}{@{}lrrrrrrrrr@{}}
\toprule
\multicolumn{10}{@{}l}{Background columns give the fraction of images with each background.} \\
& Images & sky $\uparrow$ & tree $\uparrow$ & \shortstack{concrete\\wall} $\uparrow$ & grass $\uparrow$ & snow $\uparrow$ & sea $\uparrow$ & building $\uparrow$ & clouds $\uparrow$ \\
\midrule
car & \CoocTotalA & \BestED{\CoocAa} & \BestEE{\CoocAb} & \BestEF{\CoocAc} & \BestEG{\CoocAd} & \BestEH{\CoocAe} & \BestEI{\CoocAf} & \BestEJ{\CoocAg} & \cellcolor{MethodTint}\BestEK{\CoocAh} \\
bus & \CoocTotalB & \BestEL{\CoocBa} & \BestEM{\CoocBb} & \cellcolor{MethodTint}\BestEN{\CoocBc} & \BestEO{\CoocBd} & \BestEP{\CoocBe} & \BestEQ{\CoocBf} & \BestER{\CoocBg} & \BestES{\CoocBh} \\
dog & \CoocTotalC & \cellcolor{MethodTint}\BestET{\CoocCa} & \BestEU{\CoocCb} & \BestEV{\CoocCc} & \BestEW{\CoocCd} & \BestEX{\CoocCe} & \BestEY{\CoocCf} & \BestEZ{\CoocCg} & \BestFA{\CoocCh} \\
truck & \CoocTotalD & \BestFB{\CoocDa} & \BestFC{\CoocDb} & \BestFD{\CoocDc} & \BestFE{\CoocDd} & \BestFF{\CoocDe} & \cellcolor{MethodTint}\BestFG{\CoocDf} & \BestFH{\CoocDg} & \BestFI{\CoocDh} \\
cat & \CoocTotalE & \BestFJ{\CoocEa} & \BestFK{\CoocEb} & \BestFL{\CoocEc} & \cellcolor{MethodTint}\BestFM{\CoocEd} & \BestFN{\CoocEe} & \BestFO{\CoocEf} & \BestFP{\CoocEg} & \BestFQ{\CoocEh} \\
airplane & \CoocTotalF & \BestFR{\CoocFa} & \BestFS{\CoocFb} & \BestFT{\CoocFc} & \BestFU{\CoocFd} & \BestFV{\CoocFe} & \BestFW{\CoocFf} & \cellcolor{MethodTint}\BestFX{\CoocFg} & \BestFY{\CoocFh} \\
horse & \CoocTotalG & \BestFZ{\CoocGa} & \BestGA{\CoocGb} & \BestGB{\CoocGc} & \BestGC{\CoocGd} & \cellcolor{MethodTint}\BestGD{\CoocGe} & \BestGE{\CoocGf} & \BestGF{\CoocGg} & \BestGG{\CoocGh} \\
train & \CoocTotalH & \BestGH{\CoocHa} & \cellcolor{MethodTint}\BestGI{\CoocHb} & \BestGJ{\CoocHc} & \BestGK{\CoocHd} & \BestGL{\CoocHe} & \BestGM{\CoocHf} & \BestGN{\CoocHg} & \BestGO{\CoocHh} \\
\bottomrule
\end{tabular}

\end{table}

\begin{table}[!t]
  \caption{\textbf{Against the other backgrounds of the same object, cat on concrete wall and airplane on sky and on clouds exceed our threshold on both image sets.} Each entry is the selection half's background effect for \UnitsPerLabel composites less the object's mean effect, with its fifth bootstrap percentile below. Underlining marks a threshold set between the image draws: a lower bound above \RelMargin, arrows point to higher values and bold marks each image set's column maxima.}
  \label{tab:relative}
  \centering
  \small
  \setlength{\tabcolsep}{4pt}
  \setlength{\tabcolsep}{3pt}
  \setlength{\tabcolsep}{2pt}
\begin{tabular}{@{}lrrrrrrrr@{}}
\toprule
\multicolumn{9}{@{}l}{Entries are differences in fraction correct against each object's mean effect.} \\
\multicolumn{9}{@{}l}{\emph{The first image set}} \\
& sky $\uparrow$ & tree $\uparrow$ & \shortstack{concrete\\wall} $\uparrow$ & grass $\uparrow$ & snow $\uparrow$ & sea $\uparrow$ & building $\uparrow$ & clouds $\uparrow$ \\
\midrule
car & \shortstack[r]{$\BestSE{\RelAaFirst}$\\[-2pt]\iv{$\RelLowAaFirst$}} & \shortstack[r]{$\BestSF{\RelAbFirst}$\\[-2pt]\iv{$\RelLowAbFirst$}} & \shortstack[r]{$\BestSG{\RelAcFirst}$\\[-2pt]\iv{$\RelLowAcFirst$}} & \shortstack[r]{$\BestSH{\RelAdFirst}$\\[-2pt]\iv{$\RelLowAdFirst$}} & \shortstack[r]{$\BestSI{\RelAeFirst}$\\[-2pt]\iv{$\RelLowAeFirst$}} & \shortstack[r]{$\BestSJ{\RelAfFirst}$\\[-2pt]\iv{$\RelLowAfFirst$}} & \shortstack[r]{$\BestSK{\RelAgFirst}$\\[-2pt]\iv{$\RelLowAgFirst$}} & \shortstack[r]{$\BestSL{\RelAhFirst}$\\[-2pt]\iv{$\RelLowAhFirst$}} \\
bus & \shortstack[r]{$\BestSM{\RelBaFirst}$\\[-2pt]\iv{$\RelLowBaFirst$}} & \shortstack[r]{$\BestSN{\RelBbFirst}$\\[-2pt]\iv{$\RelLowBbFirst$}} & \shortstack[r]{$\BestSO{\RelBcFirst}$\\[-2pt]\iv{$\RelLowBcFirst$}} & \shortstack[r]{$\BestSP{\RelBdFirst}$\\[-2pt]\iv{$\RelLowBdFirst$}} & \shortstack[r]{$\BestSQ{\RelBeFirst}$\\[-2pt]\iv{$\RelLowBeFirst$}} & \shortstack[r]{$\BestSR{\RelBfFirst}$\\[-2pt]\iv{$\RelLowBfFirst$}} & \shortstack[r]{$\BestSS{\RelBgFirst}$\\[-2pt]\iv{$\RelLowBgFirst$}} & \shortstack[r]{$\BestST{\RelBhFirst}$\\[-2pt]\iv{$\RelLowBhFirst$}} \\
dog & \shortstack[r]{$\BestSU{\RelCaFirst}$\\[-2pt]\iv{$\RelLowCaFirst$}} & \shortstack[r]{$\BestSV{\RelCbFirst}$\\[-2pt]\iv{$\RelLowCbFirst$}} & \shortstack[r]{$\BestSW{\RelCcFirst}$\\[-2pt]\iv{$\RelLowCcFirst$}} & \shortstack[r]{$\BestSX{\RelCdFirst}$\\[-2pt]\iv{$\RelLowCdFirst$}} & \shortstack[r]{$\BestSY{\RelCeFirst}$\\[-2pt]\iv{$\RelLowCeFirst$}} & \shortstack[r]{$\BestSZ{\RelCfFirst}$\\[-2pt]\iv{$\RelLowCfFirst$}} & \shortstack[r]{$\BestTA{\RelCgFirst}$\\[-2pt]\iv{$\RelLowCgFirst$}} & \shortstack[r]{$\BestTB{\RelChFirst}$\\[-2pt]\iv{$\RelLowChFirst$}} \\
truck & \shortstack[r]{$\underline{\BestTC{\RelDaFirst}}$\\[-2pt]\iv{$\RelLowDaFirst$}} & \shortstack[r]{$\BestTD{\RelDbFirst}$\\[-2pt]\iv{$\RelLowDbFirst$}} & \shortstack[r]{$\BestTE{\RelDcFirst}$\\[-2pt]\iv{$\RelLowDcFirst$}} & \shortstack[r]{$\BestTF{\RelDdFirst}$\\[-2pt]\iv{$\RelLowDdFirst$}} & \shortstack[r]{$\BestTG{\RelDeFirst}$\\[-2pt]\iv{$\RelLowDeFirst$}} & \shortstack[r]{$\BestTH{\RelDfFirst}$\\[-2pt]\iv{$\RelLowDfFirst$}} & \shortstack[r]{$\BestTI{\RelDgFirst}$\\[-2pt]\iv{$\RelLowDgFirst$}} & \shortstack[r]{$\BestTJ{\RelDhFirst}$\\[-2pt]\iv{$\RelLowDhFirst$}} \\
cat & \shortstack[r]{$\BestTK{\RelEaFirst}$\\[-2pt]\iv{$\RelLowEaFirst$}} & \shortstack[r]{$\BestTL{\RelEbFirst}$\\[-2pt]\iv{$\RelLowEbFirst$}} & \shortstack[r]{$\underline{\BestTM{\RelEcFirst}}$\\[-2pt]\iv{$\RelLowEcFirst$}} & \shortstack[r]{$\BestTN{\RelEdFirst}$\\[-2pt]\iv{$\RelLowEdFirst$}} & \shortstack[r]{$\BestTO{\RelEeFirst}$\\[-2pt]\iv{$\RelLowEeFirst$}} & \shortstack[r]{$\BestTP{\RelEfFirst}$\\[-2pt]\iv{$\RelLowEfFirst$}} & \shortstack[r]{$\BestTQ{\RelEgFirst}$\\[-2pt]\iv{$\RelLowEgFirst$}} & \shortstack[r]{$\BestTR{\RelEhFirst}$\\[-2pt]\iv{$\RelLowEhFirst$}} \\
airplane & \shortstack[r]{$\underline{\BestTS{\RelFaFirst}}$\\[-2pt]\iv{$\RelLowFaFirst$}} & \shortstack[r]{$\BestTT{\RelFbFirst}$\\[-2pt]\iv{$\RelLowFbFirst$}} & \shortstack[r]{$\BestTU{\RelFcFirst}$\\[-2pt]\iv{$\RelLowFcFirst$}} & \shortstack[r]{$\BestTV{\RelFdFirst}$\\[-2pt]\iv{$\RelLowFdFirst$}} & \shortstack[r]{$\BestTW{\RelFeFirst}$\\[-2pt]\iv{$\RelLowFeFirst$}} & \shortstack[r]{$\BestTX{\RelFfFirst}$\\[-2pt]\iv{$\RelLowFfFirst$}} & \shortstack[r]{$\BestTY{\RelFgFirst}$\\[-2pt]\iv{$\RelLowFgFirst$}} & \shortstack[r]{$\underline{\BestTZ{\RelFhFirst}}$\\[-2pt]\iv{$\RelLowFhFirst$}} \\
horse & \shortstack[r]{$\BestUA{\RelGaFirst}$\\[-2pt]\iv{$\RelLowGaFirst$}} & \shortstack[r]{$\BestUB{\RelGbFirst}$\\[-2pt]\iv{$\RelLowGbFirst$}} & \shortstack[r]{$\BestUC{\RelGcFirst}$\\[-2pt]\iv{$\RelLowGcFirst$}} & \shortstack[r]{$\BestUD{\RelGdFirst}$\\[-2pt]\iv{$\RelLowGdFirst$}} & \shortstack[r]{$\BestUE{\RelGeFirst}$\\[-2pt]\iv{$\RelLowGeFirst$}} & \shortstack[r]{$\BestUF{\RelGfFirst}$\\[-2pt]\iv{$\RelLowGfFirst$}} & \shortstack[r]{$\BestUG{\RelGgFirst}$\\[-2pt]\iv{$\RelLowGgFirst$}} & \shortstack[r]{$\BestUH{\RelGhFirst}$\\[-2pt]\iv{$\RelLowGhFirst$}} \\
train & \shortstack[r]{$\BestUI{\RelHaFirst}$\\[-2pt]\iv{$\RelLowHaFirst$}} & \shortstack[r]{$\BestUJ{\RelHbFirst}$\\[-2pt]\iv{$\RelLowHbFirst$}} & \shortstack[r]{$\BestUK{\RelHcFirst}$\\[-2pt]\iv{$\RelLowHcFirst$}} & \shortstack[r]{$\BestUL{\RelHdFirst}$\\[-2pt]\iv{$\RelLowHdFirst$}} & \shortstack[r]{$\BestUM{\RelHeFirst}$\\[-2pt]\iv{$\RelLowHeFirst$}} & \shortstack[r]{$\BestUN{\RelHfFirst}$\\[-2pt]\iv{$\RelLowHfFirst$}} & \shortstack[r]{$\BestUO{\RelHgFirst}$\\[-2pt]\iv{$\RelLowHgFirst$}} & \shortstack[r]{$\BestUP{\RelHhFirst}$\\[-2pt]\iv{$\RelLowHhFirst$}} \\
\midrule
\multicolumn{9}{@{}l}{\emph{The new image set}} \\
& sky $\uparrow$ & tree $\uparrow$ & \shortstack{concrete\\wall} $\uparrow$ & grass $\uparrow$ & snow $\uparrow$ & sea $\uparrow$ & building $\uparrow$ & clouds $\uparrow$ \\
\midrule
car & \shortstack[r]{$\BestUQ{\RelAaNew}$\\[-2pt]\iv{$\RelLowAaNew$}} & \shortstack[r]{$\BestUR{\RelAbNew}$\\[-2pt]\iv{$\RelLowAbNew$}} & \shortstack[r]{$\BestUS{\RelAcNew}$\\[-2pt]\iv{$\RelLowAcNew$}} & \shortstack[r]{$\BestUT{\RelAdNew}$\\[-2pt]\iv{$\RelLowAdNew$}} & \shortstack[r]{$\BestUU{\RelAeNew}$\\[-2pt]\iv{$\RelLowAeNew$}} & \shortstack[r]{$\BestUV{\RelAfNew}$\\[-2pt]\iv{$\RelLowAfNew$}} & \shortstack[r]{$\BestUW{\RelAgNew}$\\[-2pt]\iv{$\RelLowAgNew$}} & \shortstack[r]{$\BestUX{\RelAhNew}$\\[-2pt]\iv{$\RelLowAhNew$}} \\
bus & \shortstack[r]{$\BestUY{\RelBaNew}$\\[-2pt]\iv{$\RelLowBaNew$}} & \shortstack[r]{$\BestUZ{\RelBbNew}$\\[-2pt]\iv{$\RelLowBbNew$}} & \shortstack[r]{$\BestVA{\RelBcNew}$\\[-2pt]\iv{$\RelLowBcNew$}} & \shortstack[r]{$\BestVB{\RelBdNew}$\\[-2pt]\iv{$\RelLowBdNew$}} & \shortstack[r]{$\BestVC{\RelBeNew}$\\[-2pt]\iv{$\RelLowBeNew$}} & \shortstack[r]{$\BestVD{\RelBfNew}$\\[-2pt]\iv{$\RelLowBfNew$}} & \shortstack[r]{$\BestVE{\RelBgNew}$\\[-2pt]\iv{$\RelLowBgNew$}} & \shortstack[r]{$\BestVF{\RelBhNew}$\\[-2pt]\iv{$\RelLowBhNew$}} \\
dog & \shortstack[r]{$\BestVG{\RelCaNew}$\\[-2pt]\iv{$\RelLowCaNew$}} & \shortstack[r]{$\BestVH{\RelCbNew}$\\[-2pt]\iv{$\RelLowCbNew$}} & \shortstack[r]{$\BestVI{\RelCcNew}$\\[-2pt]\iv{$\RelLowCcNew$}} & \shortstack[r]{$\BestVJ{\RelCdNew}$\\[-2pt]\iv{$\RelLowCdNew$}} & \shortstack[r]{$\BestVK{\RelCeNew}$\\[-2pt]\iv{$\RelLowCeNew$}} & \shortstack[r]{$\BestVL{\RelCfNew}$\\[-2pt]\iv{$\RelLowCfNew$}} & \shortstack[r]{$\BestVM{\RelCgNew}$\\[-2pt]\iv{$\RelLowCgNew$}} & \shortstack[r]{$\BestVN{\RelChNew}$\\[-2pt]\iv{$\RelLowChNew$}} \\
truck & \shortstack[r]{$\BestVO{\RelDaNew}$\\[-2pt]\iv{$\RelLowDaNew$}} & \shortstack[r]{$\BestVP{\RelDbNew}$\\[-2pt]\iv{$\RelLowDbNew$}} & \shortstack[r]{$\BestVQ{\RelDcNew}$\\[-2pt]\iv{$\RelLowDcNew$}} & \shortstack[r]{$\BestVR{\RelDdNew}$\\[-2pt]\iv{$\RelLowDdNew$}} & \shortstack[r]{$\BestVS{\RelDeNew}$\\[-2pt]\iv{$\RelLowDeNew$}} & \shortstack[r]{$\BestVT{\RelDfNew}$\\[-2pt]\iv{$\RelLowDfNew$}} & \shortstack[r]{$\BestVU{\RelDgNew}$\\[-2pt]\iv{$\RelLowDgNew$}} & \shortstack[r]{$\underline{\BestVV{\RelDhNew}}$\\[-2pt]\iv{$\RelLowDhNew$}} \\
cat & \shortstack[r]{$\BestVW{\RelEaNew}$\\[-2pt]\iv{$\RelLowEaNew$}} & \shortstack[r]{$\BestVX{\RelEbNew}$\\[-2pt]\iv{$\RelLowEbNew$}} & \shortstack[r]{$\underline{\BestVY{\RelEcNew}}$\\[-2pt]\iv{$\RelLowEcNew$}} & \shortstack[r]{$\BestVZ{\RelEdNew}$\\[-2pt]\iv{$\RelLowEdNew$}} & \shortstack[r]{$\BestWA{\RelEeNew}$\\[-2pt]\iv{$\RelLowEeNew$}} & \shortstack[r]{$\BestWB{\RelEfNew}$\\[-2pt]\iv{$\RelLowEfNew$}} & \shortstack[r]{$\BestWC{\RelEgNew}$\\[-2pt]\iv{$\RelLowEgNew$}} & \shortstack[r]{$\BestWD{\RelEhNew}$\\[-2pt]\iv{$\RelLowEhNew$}} \\
airplane & \shortstack[r]{$\underline{\BestWE{\RelFaNew}}$\\[-2pt]\iv{$\RelLowFaNew$}} & \shortstack[r]{$\BestWF{\RelFbNew}$\\[-2pt]\iv{$\RelLowFbNew$}} & \shortstack[r]{$\BestWG{\RelFcNew}$\\[-2pt]\iv{$\RelLowFcNew$}} & \shortstack[r]{$\BestWH{\RelFdNew}$\\[-2pt]\iv{$\RelLowFdNew$}} & \shortstack[r]{$\BestWI{\RelFeNew}$\\[-2pt]\iv{$\RelLowFeNew$}} & \shortstack[r]{$\BestWJ{\RelFfNew}$\\[-2pt]\iv{$\RelLowFfNew$}} & \shortstack[r]{$\BestWK{\RelFgNew}$\\[-2pt]\iv{$\RelLowFgNew$}} & \shortstack[r]{$\underline{\BestWL{\RelFhNew}}$\\[-2pt]\iv{$\RelLowFhNew$}} \\
horse & \shortstack[r]{$\BestWM{\RelGaNew}$\\[-2pt]\iv{$\RelLowGaNew$}} & \shortstack[r]{$\BestWN{\RelGbNew}$\\[-2pt]\iv{$\RelLowGbNew$}} & \shortstack[r]{$\BestWO{\RelGcNew}$\\[-2pt]\iv{$\RelLowGcNew$}} & \shortstack[r]{$\BestWP{\RelGdNew}$\\[-2pt]\iv{$\RelLowGdNew$}} & \shortstack[r]{$\BestWQ{\RelGeNew}$\\[-2pt]\iv{$\RelLowGeNew$}} & \shortstack[r]{$\BestWR{\RelGfNew}$\\[-2pt]\iv{$\RelLowGfNew$}} & \shortstack[r]{$\BestWS{\RelGgNew}$\\[-2pt]\iv{$\RelLowGgNew$}} & \shortstack[r]{$\BestWT{\RelGhNew}$\\[-2pt]\iv{$\RelLowGhNew$}} \\
train & \shortstack[r]{$\BestWU{\RelHaNew}$\\[-2pt]\iv{$\RelLowHaNew$}} & \shortstack[r]{$\BestWV{\RelHbNew}$\\[-2pt]\iv{$\RelLowHbNew$}} & \shortstack[r]{$\BestWW{\RelHcNew}$\\[-2pt]\iv{$\RelLowHcNew$}} & \shortstack[r]{$\BestWX{\RelHdNew}$\\[-2pt]\iv{$\RelLowHdNew$}} & \shortstack[r]{$\BestWY{\RelHeNew}$\\[-2pt]\iv{$\RelLowHeNew$}} & \shortstack[r]{$\BestWZ{\RelHfNew}$\\[-2pt]\iv{$\RelLowHfNew$}} & \shortstack[r]{$\BestXA{\RelHgNew}$\\[-2pt]\iv{$\RelLowHgNew$}} & \shortstack[r]{$\BestXB{\RelHhNew}$\\[-2pt]\iv{$\RelLowHhNew$}} \\
\bottomrule
\end{tabular}

\end{table}

\section{The classifiers}
\label{sec:supp-classifiers}

We drew the \Models classifiers from timm's results table by one seeded rule, so that the sample spans ImageNet accuracy and architecture families. The draw kept the rows that take an input of \InputSide by \InputSide pixels and whose model cards give a permissive licence, \ZooEligible rows from \ZooFamilies architecture families. A family groups checkpoints by the initial run of lower case letters in the architecture's name in timm's registry, after a leading \texttt{tf}, \texttt{gluon} or \texttt{legacy} is removed. The family draw chooses one checkpoint per family before binning, so a family whose checkpoints span several bins enters only the bin of its chosen checkpoint.

The bins spread the sample over ImageNet accuracy. The ImageNet accuracy bin edges in percent are \BinEdgeA, \BinEdgeB, \BinEdgeC, \BinEdgeD, \BinEdgeE, \BinEdgeF, \BinEdgeG, \BinEdgeH and \BinEdgeI. Each bin includes its left edge and excludes its right edge, except the last, which includes both. The quotas in that order are \BinQuotaA, \BinQuotaB, \BinQuotaC, \BinQuotaD, \BinQuotaE, \BinQuotaF, \BinQuotaG and \BinQuotaH. The allocation raises a common quota until another full increment would exceed \Models, caps each bin at its available families and assigns the remaining places in ascending bin order. Seed \SeedFamilies permutes the eligible checkpoints within each sorted family, and \SeedBins draws within bins from lowest to highest accuracy.

\Cref{tab:checkpoints} and \cref{tab:training} identify the classifiers. Every checkpoint appears in \cref{tab:checkpoints} under the name its architecture's authors published, with its identifier in timm's registry, its ImageNet accuracy and its drop on each image set. The drops of the selection half on the first image set are those of the half that fixed the pairing on the same composites, and the table gives them for completeness. \Cref{fig:sets} plots each evaluation classifier's drop on the first image set against its drop on the new one. \Cref{tab:training} gives the training data and the teacher that each checkpoint's card states, with the normalization from its timm configuration and the licence its card names.

The drop is present among classifiers that share one training set. \Cref{tab:groups} gives the hardest pairing's drop within groups that the model cards define, which we fixed from the cards before computing any group's drop. The \GroupInOneK evaluation classifiers whose cards state \INonek alone exceed the margin at their lower bound on the new and third image sets, and so do the \GroupNoTeacher of them whose cards name no teacher. The \GroupDistilled whose cards state distillation and the \GroupFurther with further training data are fewer than the \GroupFloor classifiers that our tests need, so we report their drops without testing either group on its own. The classifiers trained on \INonek alone lose more than those with further data, by \GroupGapNew on the new image set and by \GroupGapThird on the third, each with an interval above zero.

\FloatBarrier
\begingroup
\small
\setlength{\tabcolsep}{2pt}
\setlength{\LTcapwidth}{\textwidth}
\begin{longtable}{@{}p{0.16\textwidth}p{0.29\textwidth}rp{0.20\textwidth}p{0.20\textwidth}@{}}
\caption{\textbf{Every evaluation classifier that passes the compositing check has a positive estimated drop on both image sets.}\aftertest Rows give each checkpoint's architecture, identifier in the PyTorch Image Models library (timm), ImageNet accuracy and drop against random pairings on each image set, with the middle nine tenths of \BootReps source image resamples as a subscript. The mark $\dagger$ names exclusion by the compositing check on that image set and \S{} an interval that extends below zero. Arrows give the better direction, bold the extremes within each half and $\ddagger$ a post hoc analysis.}\label{tab:checkpoints}\\

\toprule
\multicolumn{5}{@{}l}{ImageNet accuracy is a percentage. Both drops are fractions of answers.} \\
Architecture & Identifier in timm & ImageNet $\uparrow$ & \shortstack{First\\image set $\downarrow$} & \shortstack{New\\image set $\downarrow$} \\
\midrule
\multicolumn{5}{@{}l}{\emph{Selection half}} \\
\wrapname{\ArchAa}~\cite{cui2021pplcnet} & \ckpt{\CkptAa} & \BestN{\TopAa} & \shortstack[r]{\BestO{\DropAaFirst}\\[-2pt]\iv{\DropAaFirstLo to \DropAaFirstHi}} & \shortstack[r]{\BestP{\DropAaNew}\\[-2pt]\iv{\DropAaNewLo to \DropAaNewHi}} \\
\wrapname{\ArchAh}~\cite{li2019selective} & \ckpt{\CkptAh} & \BestQ{\TopAh} & \shortstack[r]{\BestR{\DropAhFirst}\\[-2pt]\iv{\DropAhFirstLo to \DropAhFirstHi}} & \shortstack[r]{\BestS{\DropAhNew}\\[-2pt]\iv{\DropAhNewLo to \DropAhNewHi}} \\
\wrapname{\ArchAc}~\cite{han2019ghostnet} & \ckpt{\CkptAc} & \BestT{\TopAc} & \shortstack[r]{\BestU{\DropAcFirst}\\[-2pt]\iv{\DropAcFirstLo to \DropAcFirstHi}} & \shortstack[r]{\BestV{\DropAcNew}\\[-2pt]\iv{\DropAcNewLo to \DropAcNewHi}} \\
\wrapname{\ArchAg}~\cite{ma2024rewrite} & \ckpt{\CkptAg} & \BestW{\TopAg} & \shortstack[r]{\BestX{\DropAgFirst}\\[-2pt]\iv{\DropAgFirstLo to \DropAgFirstHi}} & \shortstack[r]{\BestY{\DropAgNew}\\[-2pt]\iv{\DropAgNewLo to \DropAgNewHi}} \\
\wrapname{\ArchAf}~\cite{wu2018fbnet} & \ckpt{\CkptAf} & \BestZ{\TopAf} & \shortstack[r]{\BestAA{\DropAfFirst}\\[-2pt]\iv{\DropAfFirstLo to \DropAfFirstHi}} & \shortstack[r]{\BestAB{\DropAfNew}\\[-2pt]\iv{\DropAfNewLo to \DropAfNewHi}} \\
\wrapname{\ArchAp}~\cite{chen2022repghost} & \ckpt{\CkptAp} & \BestAC{\TopAp} & \shortstack[r]{\BestAD{\DropApFirst}\\[-2pt]\iv{\DropApFirstLo to \DropApFirstHi}} & \shortstack[r]{\BestAE{\DropApNew}\\[-2pt]\iv{\DropApNewLo to \DropApNewHi}} \\
\wrapname{\ArchAm}~\cite{tan2019mixconv} & \ckpt{\CkptAm} & \BestAF{\TopAm} & \shortstack[r]{\BestAG{\DropAmFirst}\\[-2pt]\iv{\DropAmFirstLo to \DropAmFirstHi}} & \shortstack[r]{\BestAH{\DropAmNew}\\[-2pt]\iv{\DropAmNewLo to \DropAmNewHi}} \\
\wrapname{\ArchAo}~\cite{yu2017deep} & \ckpt{\CkptAo} & \BestAI{\TopAo} & \shortstack[r]{\BestAJ{\DropAoFirst}\\[-2pt]\iv{\DropAoFirstLo to \DropAoFirstHi}} & \shortstack[r]{\BestAK{\DropAoNew}\\[-2pt]\iv{\DropAoNewLo to \DropAoNewHi}} \\
\wrapname{\ArchAr}~\cite{chen2023run} & \ckpt{\CkptAr} & \BestAL{\TopAr} & \shortstack[r]{\BestAM{\DropArFirst}\\[-2pt]\iv{\DropArFirstLo to \DropArFirstHi}} & \shortstack[r]{\BestAN{\DropArNew}\\[-2pt]\iv{\DropArNewLo to \DropArNewHi}} \\
\wrapname{\ArchAx}~\cite{han2020rethinking} & \ckpt{\CkptAx} & \BestAO{\TopAx} & \shortstack[r]{\BestAP{\DropAxFirst}\\[-2pt]\iv{\DropAxFirstLo to \DropAxFirstHi}} & \shortstack[r]{\BestAQ{\DropAxNew}\\[-2pt]\iv{\DropAxNewLo to \DropAxNewHi}} \\
\wrapname{\ArchAw}~\cite{yu2022metaformer} & \ckpt{\CkptAw} & \BestAR{\TopAw} & \shortstack[r]{\BestAS{\DropAwFirst}\\[-2pt]\iv{\DropAwFirstLo to \DropAwFirstHi}} & \shortstack[r]{\BestAT{\DropAwNew}\\[-2pt]\iv{\DropAwNewLo to \DropAwNewHi}\refusedmark} \\
\wrapname{\ArchAt}~\cite{touvron2021resmlp} & \ckpt{\CkptAt} & \BestAU{\TopAt} & \shortstack[r]{\BestAV{\DropAtFirst}\\[-2pt]\iv{\DropAtFirstLo to \DropAtFirstHi}} & \shortstack[r]{\BestAW{\DropAtNew}\\[-2pt]\iv{\DropAtNewLo to \DropAtNewHi}} \\
\wrapname{\ArchAy}~\cite{dai2021coatnet} & \ckpt{\CkptAy} & \BestAX{\TopAy} & \shortstack[r]{\BestAY{\DropAyFirst}\\[-2pt]\iv{\DropAyFirstLo to \DropAyFirstHi}} & \shortstack[r]{\BestAZ{\DropAyNew}\\[-2pt]\iv{\DropAyNewLo to \DropAyNewHi}} \\
\wrapname{\ArchBb}~\cite{radosavovic2020designing} & \ckpt{\CkptBb} & \BestBA{\TopBb} & \shortstack[r]{\BestBB{\DropBbFirst}\\[-2pt]\iv{\DropBbFirstLo to \DropBbFirstHi}} & \shortstack[r]{\BestBC{\DropBbNew}\\[-2pt]\iv{\DropBbNewLo to \DropBbNewHi}\refusedmark} \\
\wrapname{\ArchBa}~\cite{yu2023inceptionnext} & \ckpt{\CkptBa} & \BestBD{\TopBa} & \shortstack[r]{\BestBE{\DropBaFirst}\\[-2pt]\iv{$\DropBaFirstLo$ to \DropBaFirstHi}} & \shortstack[r]{\BestBF{\DropBaNew}\\[-2pt]\iv{$\DropBaNewLo$ to \DropBaNewHi}} \\
\wrapname{\ArchBd}~\cite{li2022rethinking} & \ckpt{\CkptBd} & \BestBG{\TopBd} & \shortstack[r]{\BestBH{\DropBdFirst}\\[-2pt]\iv{\DropBdFirstLo to \DropBdFirstHi}} & \shortstack[r]{\BestBI{\DropBdNew}\\[-2pt]\iv{\DropBdNewLo to \DropBdNewHi}} \\
\wrapname{\ArchBc}~\cite{peng2022beit} & \ckpt{\CkptBc} & \BestBJ{\TopBc} & \shortstack[r]{\BestBK{\DropBcFirst}\\[-2pt]\iv{\DropBcFirstLo to \DropBcFirstHi}} & \shortstack[r]{\BestBL{\DropBcNew}\\[-2pt]\iv{\DropBcNewLo to \DropBcNewHi}} \\
\wrapname{\ArchBi}~\cite{yuan2021volo} & \ckpt{\CkptBi} & \BestBM{\TopBi} & \shortstack[r]{\BestBN{\DropBiFirst}\\[-2pt]\iv{\DropBiFirstLo to \DropBiFirstHi}} & \shortstack[r]{\BestBO{\DropBiNew}\\[-2pt]\iv{$\DropBiNewLo$ to \DropBiNewHi}} \\
\wrapname{\ArchBk}~\cite{liu2022convnet} & \ckpt{\CkptBk} & \BestBP{\TopBk} & \shortstack[r]{\BestBQ{\DropBkFirst}\\[-2pt]\iv{\DropBkFirstLo to \DropBkFirstHi}} & \shortstack[r]{\BestBR{\DropBkNew}\\[-2pt]\iv{$\DropBkNewLo$ to \DropBkNewHi}\refusedmark} \\
\wrapname{\ArchBl}~\cite{wang2022image} & \ckpt{\CkptBl} & \BestBS{\TopBl} & \shortstack[r]{\BestBT{\DropBlFirst}\\[-2pt]\iv{$\DropBlFirstLo$ to \DropBlFirstHi}} & \shortstack[r]{\BestBU{\DropBlNew}\\[-2pt]\iv{$\DropBlNewLo$ to \DropBlNewHi}} \\
\midrule
\multicolumn{5}{@{}l}{\emph{Evaluation half}} \\
\wrapname{\ArchAb}~\cite{howard2019searching} & \ckpt{\CkptAb} & \BestBV{\TopAb} & \shortstack[r]{\BestBW{\DropAbFirst}\\[-2pt]\iv{\DropAbFirstLo to \DropAbFirstHi}} & \shortstack[r]{\BestBX{\DropAbNew}\\[-2pt]\iv{\DropAbNewLo to \DropAbNewHi}} \\
\wrapname{\ArchAd}~\cite{simonyan2014very} & \ckpt{\CkptAd} & \BestBY{\TopAd} & \shortstack[r]{\BestBZ{\DropAdFirst}\\[-2pt]\iv{\DropAdFirstLo to \DropAdFirstHi}} & \shortstack[r]{\BestCA{\DropAdNew}\\[-2pt]\iv{\DropAdNewLo to \DropAdNewHi}} \\
\wrapname{\ArchAj}~\cite{stamoulis2019single} & \ckpt{\CkptAj} & \BestCB{\TopAj} & \shortstack[r]{\BestCC{\DropAjFirst}\\[-2pt]\iv{\DropAjFirstLo to \DropAjFirstHi}} & \shortstack[r]{\BestCD{\DropAjNew}\\[-2pt]\iv{\DropAjNewLo to \DropAjNewHi}\refusedmark} \\
\wrapname{\ArchAe}~\cite{tan2018mnasnet} & \ckpt{\CkptAe} & \BestCE{\TopAe} & \shortstack[r]{\BestCF{\DropAeFirst}\\[-2pt]\iv{\DropAeFirstLo to \DropAeFirstHi}} & \shortstack[r]{\BestCG{\DropAeNew}\\[-2pt]\iv{\DropAeNewLo to \DropAeNewHi}} \\
\wrapname{\ArchAi}~\cite{huang2016densely} & \ckpt{\CkptAi} & \BestCH{\TopAi} & \shortstack[r]{\BestCI{\DropAiFirst}\\[-2pt]\iv{\DropAiFirstLo to \DropAiFirstHi}} & \shortstack[r]{\BestCJ{\DropAiNew}\\[-2pt]\iv{\DropAiNewLo to \DropAiNewHi}} \\
\wrapname{\ArchAk}~\cite{graham2021levit} & \ckpt{\CkptAk} & \BestCK{\TopAk} & \shortstack[r]{\BestCL{\DropAkFirst}\\[-2pt]\iv{\DropAkFirstLo to \DropAkFirstHi}} & \shortstack[r]{\BestCM{\DropAkNew}\\[-2pt]\iv{\DropAkNewLo to \DropAkNewHi}} \\
\wrapname{\ArchAn}~\cite{huang2016densely} & \ckpt{\CkptAn} & \BestCN{\TopAn} & \shortstack[r]{\BestCO{\DropAnFirst}\\[-2pt]\iv{\DropAnFirstLo to \DropAnFirstHi}} & \shortstack[r]{\BestCP{\DropAnNew}\\[-2pt]\iv{\DropAnNewLo to \DropAnNewHi}\refusedmark} \\
\wrapname{\ArchAq}~\cite{touvron2021going} & \ckpt{\CkptAq} & \BestCQ{\TopAq} & \shortstack[r]{\BestCR{\DropAqFirst}\\[-2pt]\iv{\DropAqFirstLo to \DropAqFirstHi}} & \shortstack[r]{\BestCS{\DropAqNew}\\[-2pt]\iv{\DropAqNewLo to \DropAqNewHi}\refusedmark} \\
\wrapname{\ArchAl}~\cite{dai2020fbnetv3} & \ckpt{\CkptAl} & \BestCT{\TopAl} & \shortstack[r]{\BestCU{\DropAlFirst}\\[-2pt]\iv{\DropAlFirstLo to \DropAlFirstHi}} & \shortstack[r]{\BestCV{\DropAlNew}\\[-2pt]\iv{\DropAlNewLo to \DropAlNewHi}} \\
\wrapname{\ArchAu}~\cite{yu2021metaformer} & \ckpt{\CkptAu} & \BestCW{\TopAu} & \shortstack[r]{\BestCX{\DropAuFirst}\\[-2pt]\iv{\DropAuFirstLo to \DropAuFirstHi}} & \shortstack[r]{\BestCY{\DropAuNew}\\[-2pt]\iv{\DropAuNewLo to \DropAuNewHi}\refusedmark} \\
\wrapname{\ArchAz}~\cite{he2015deep} & \ckpt{\CkptAz} & \BestCZ{\TopAz} & \shortstack[r]{\BestDA{\DropAzFirst}\\[-2pt]\iv{\DropAzFirstLo to \DropAzFirstHi}} & \shortstack[r]{\BestDB{\DropAzNew}\\[-2pt]\iv{\DropAzNewLo to \DropAzNewHi}} \\
\wrapname{\ArchAv}~\cite{zagoruyko2016wide} & \ckpt{\CkptAv} & \BestDC{\TopAv} & \shortstack[r]{\BestDD{\DropAvFirst}\\[-2pt]\iv{\DropAvFirstLo to \DropAvFirstHi}\refusedmark} & \shortstack[r]{\BestDE{\DropAvNew}\\[-2pt]\iv{\DropAvNewLo to \DropAvNewHi}\refusedmark} \\
\wrapname{\ArchAs}~\cite{dai2021coatnet} & \ckpt{\CkptAs} & \BestDF{\TopAs} & \shortstack[r]{\BestDG{\DropAsFirst}\\[-2pt]\iv{\DropAsFirstLo to \DropAsFirstHi}} & \shortstack[r]{\BestDH{\DropAsNew}\\[-2pt]\iv{\DropAsNewLo to \DropAsNewHi}\refusedmark} \\
\wrapname{\ArchBg}~\cite{elnouby2021xcit} & \ckpt{\CkptBg} & \BestDI{\TopBg} & \shortstack[r]{\BestDJ{\DropBgFirst}\\[-2pt]\iv{\DropBgFirstLo to \DropBgFirstHi}} & \shortstack[r]{\BestDK{\DropBgNew}\textsuperscript{\S}\\[-2pt]\iv{$\DropBgNewLo$ to \DropBgNewHi}} \\
\wrapname{\ArchBf}~\cite{han2020rethinking} & \ckpt{\CkptBf} & \BestDL{\TopBf} & \shortstack[r]{\BestDM{\DropBfFirst}\\[-2pt]\iv{\DropBfFirstLo to \DropBfFirstHi}} & \shortstack[r]{\BestDN{\DropBfNew}\\[-2pt]\iv{\DropBfNewLo to \DropBfNewHi}} \\
\wrapname{\ArchBe}~\cite{yang2022focal} & \ckpt{\CkptBe} & \BestDO{\TopBe} & \shortstack[r]{\BestDP{\DropBeFirst}\\[-2pt]\iv{\DropBeFirstLo to \DropBeFirstHi}} & \shortstack[r]{\BestDQ{\DropBeNew}\textsuperscript{\S}\\[-2pt]\iv{$\DropBeNewLo$ to \DropBeNewHi}} \\
\wrapname{\ArchBm}~\cite{yu2022metaformer} & \ckpt{\CkptBm} & \BestDR{\TopBm} & \shortstack[r]{\BestDS{\DropBmFirst}\\[-2pt]\iv{\DropBmFirstLo to \DropBmFirstHi}\refusedmark} & \shortstack[r]{\BestDT{\DropBmNew}\\[-2pt]\iv{\DropBmNewLo to \DropBmNewHi}} \\
\wrapname{\ArchBh}~\cite{cui2021beyond} & \ckpt{\CkptBh} & \BestDU{\TopBh} & \shortstack[r]{\BestDV{\DropBhFirst}\textsuperscript{\S}\\[-2pt]\iv{$\DropBhFirstLo$ to \DropBhFirstHi}} & \shortstack[r]{\BestDW{\DropBhNew}\\[-2pt]\iv{\DropBhNewLo to \DropBhNewHi}} \\
\wrapname{\ArchBn}~\cite{tu2022maxvit} & \ckpt{\CkptBn} & \BestDX{\TopBn} & \shortstack[r]{\BestDY{\DropBnFirst}\textsuperscript{\S}\\[-2pt]\iv{$\DropBnFirstLo$ to \DropBnFirstHi}} & \shortstack[r]{\BestDZ{\DropBnNew}\textsuperscript{\S}\\[-2pt]\iv{$\DropBnNewLo$ to \DropBnNewHi}} \\
\wrapname{\ArchBj}~\cite{sun2023evaclip} & \ckpt{\CkptBj} & \BestEA{\TopBj} & \shortstack[r]{\BestEB{\DropBjFirst}\textsuperscript{\S}\\[-2pt]\iv{$\DropBjFirstLo$ to \DropBjFirstHi}} & \shortstack[r]{\BestEC{\DropBjNew}\\[-2pt]\iv{\DropBjNewLo to \DropBjNewHi}} \\
\bottomrule

\end{longtable}

\endgroup

\begin{table}[!t]
  \caption{\textbf{The classifiers differ in their training data and in the teacher, if any, from which they were distilled.} Letters $A$ to $F$ index each card's data and teacher and timm's normalization, whose vectors follow red, green, blue order, beside the licence each card names. Data include \DatasetLaion~\cite{schuhmann2021laion}, \DatasetCoyo~\cite{byeon2022coyo}, Visual Genome~\cite{krishna2017visual} and \DatasetCombined, Conceptual Captions with \DatasetConceptualTwelve~\cite{sharma2018conceptual,changpinyo2021conceptual}. CLIP \ClipVariant denotes the base Contrastive Language Image Pretraining (CLIP) vision transformer with sixteen pixel patches~\cite{radford2021learning}, VQ-KD vector quantized knowledge distillation~\cite{peng2022beit} and SSLD simple semi supervised distillation~\cite{cui2021beyond}.}
  \label{tab:training}
  \centering
  \small
  \setlength{\tabcolsep}{3pt}
  \renewcommand{\ckpt}[1]{\texttt{\let\_\checkpointbreak#1}}
  \begin{minipage}[t]{0.495\textwidth}
\vspace{0pt}
\setlength{\tabcolsep}{0.5pt}
\begin{tabular}{@{}p{0.30\linewidth}cccp{0.25\linewidth}@{}}
\toprule
Classifier & Data & Teacher & Norm & Licence \\
\midrule
\wrapname{\ArchAa} & A & A & A & \LicNameAa \\
\wrapname{\ArchAb} & A & A & A & \LicNameAb \\
\wrapname{\ArchAc} & A & A & A & \LicNameAc \\
\wrapname{\ArchAd} & A & A & A & \LicNameAd \\
\wrapname{\ArchAe} & A & A & A & \LicNameAe \\
\wrapname{\ArchAf} & A & A & A & \LicNameAf \\
\wrapname{\ArchAg} & A & A & A & \LicNameAg \\
\wrapname{\ArchAh} & A & A & A & \LicNameAh \\
\wrapname{\ArchAi} & A & A & A & \LicNameAi \\
\wrapname{\ArchAj} & A & A & A & \LicNameAj \\
\wrapname{\ArchAk} & A & B & A & \LicNameAk \\
\wrapname{\ArchAl} & A & A & A & \LicNameAl \\
\wrapname{\ArchAm} & A & A & A & \LicNameAm \\
\wrapname{\ArchAn} & A & A & A & \LicNameAn \\
\wrapname{\ArchAo} & A & A & A & \LicNameAo \\
\wrapname{\ArchAp} & A & A & A & \LicNameAp \\
\wrapname{\ArchAq} & A & B & A & \LicNameAq \\
\wrapname{\ArchAr} & A & A & A & \LicNameAr \\
\wrapname{\ArchAs} & A & A & B & \LicNameAs \\
\wrapname{\ArchAt} & A & B & A & \LicNameAt \\
\bottomrule
\end{tabular}
\end{minipage}
\hfill
\begin{minipage}[t]{0.495\textwidth}
\vspace{0pt}
\setlength{\tabcolsep}{0.5pt}
\begin{tabular}{@{}p{0.30\linewidth}cccp{0.25\linewidth}@{}}
\toprule
Classifier & Data & Teacher & Norm & Licence \\
\midrule
\wrapname{\ArchAu} & A & A & A & \LicNameAu \\
\wrapname{\ArchAv} & A & A & A & \LicNameAv \\
\wrapname{\ArchAw} & A & A & A & \LicNameAw \\
\wrapname{\ArchAx} & A & A & A & \LicNameAx \\
\wrapname{\ArchAy} & A & A & A & \LicNameAy \\
\wrapname{\ArchAz} & A & A & A & \LicNameAz \\
\wrapname{\ArchBa} & A & A & A & \LicNameBa \\
\wrapname{\ArchBb} & A & A & A & \LicNameBb \\
\wrapname{\ArchBc} & A & C & A & \LicNameBc \\
\wrapname{\ArchBd} & A & B & A & \LicNameBd \\
\wrapname{\ArchBe} & A & A & A & \LicNameBe \\
\wrapname{\ArchBf} & B & A & A & \LicNameBf \\
\wrapname{\ArchBg} & A & A & A & \LicNameBg \\
\wrapname{\ArchBh} & C & D & A & \LicNameBh \\
\wrapname{\ArchBi} & A & E & A & \LicNameBi \\
\wrapname{\ArchBj} & D & F & C & \LicNameBj \\
\wrapname{\ArchBk} & E & A & A & \LicNameBk \\
\wrapname{\ArchBl} & F & A & A & \LicNameBl \\
\wrapname{\ArchBm} & E & A & A & \LicNameBm \\
\wrapname{\ArchBn} & B & A & B & \LicNameBn \\
\bottomrule
\end{tabular}
\end{minipage}
\par\medskip
\begin{tabular}{@{}lp{0.45\textwidth}p{0.45\textwidth}@{}}
\toprule
Code & Training data & Teacher \\
\midrule
A & \TrainingAa & \TeacherAa \\
B & \TrainingBf & \TeacherAk \\
C & \TrainingBh & \TeacherBc \\
D & \TrainingBj & \TeacherBh \\
E & \TrainingBk & \TeacherBi \\
F & \TrainingBl & \TeacherBj \\
\bottomrule
\end{tabular}
\par\medskip
\begin{tabular}{@{}lll@{}}
\toprule
Norm & Mean in red, green, blue order & Deviation in the same order \\
\midrule
A & \NormMeanAa & \NormStdAa \\
B & \NormMeanAs & \NormStdAs \\
C & \NormMeanBj & \NormStdBj \\
\bottomrule
\end{tabular}

\end{table}

\section{Data and software}
\label{sec:supp-data}

Every source image is a COCO 2017 training image under the \CCBYtwo licence, which COCO's annotation file gives as its fourth licence. COCO's annotations and COCO-Stuff's maps are released under the \CCBYfour licence. COCO-Stuff releases maps for both the 2017 training and validation splits. All three image sets sample foregrounds and background crops only from the training images. The label mapping follows the noun hierarchy of \WordNetRel. The results table of timm is released under the \Apachetwo licence, and each checkpoint under the licence its model card gives.

\Cref{tab:credits} credits every photograph the figures show by its file name on COCO's host and its address on Flickr, with the panel it appears in and its role. The frames of \cref{fig:overview} show each image set's first car on its background crop in the hardest pairing and in the first random pairing that gives the car another background.

The scoring ran in PyTorch \VerTorch~\cite{paszke2019pytorch} with torchvision \VerTorchvision~\cite{torchvision2026}, timm \VerTimm, NumPy \VerNumpy and Pillow \VerPillow~\cite{clark2025pillow} under Python \VerPython. Each scoring run used one accelerator, an \DeviceSxm or an \DevicePcie. The scoring of the first image set took \SecondsFirst seconds of accelerator time and that of the new image set \SecondsNew seconds. The post hoc analyses ran on a processor in NumPy \VerNumpyHere and SciPy \VerScipy.

\FloatBarrier
\begingroup
\small
\setlength{\tabcolsep}{2pt}
\setlength{\LTcapwidth}{\textwidth}
\begin{longtable}{@{}p{0.22\textwidth}p{0.17\textwidth}p{0.16\textwidth}p{0.38\textwidth}@{}}
\caption{\textbf{The illustrated photographs are COCO 2017 training images under the \CCBYtwo licence.} Each row gives a photograph's file name, figure, role and Flickr address. COCO's host serves each file under \url{http://images.cocodataset.org/train2017/}.}\label{tab:credits}\\

\toprule
File on COCO's host & Where it appears & Role & Flickr address \\
\midrule
\texttt{000000397859.jpg} & \cref{fig:overview}, first image set & car foreground & \url{http://farm8.staticflickr.com/7406/9755146383_1ef2b49925_z.jpg} \\
\texttt{000000166881.jpg} & \cref{fig:overview}, first image set & clouds background crop & \url{http://farm1.staticflickr.com/47/105785354_48bfc195ef_z.jpg} \\
\texttt{000000215024.jpg} & \cref{fig:overview}, first image set & sea background crop & \url{http://farm6.staticflickr.com/5445/9766445991_f3224376dc_z.jpg} \\
\texttt{000000173634.jpg} & \cref{fig:overview}, new image set and \cref{fig:gallery} & car foreground & \url{http://farm9.staticflickr.com/8025/7444957000_32ec757822_z.jpg} \\
\texttt{000000006120.jpg} & \cref{fig:overview}, new image set and \cref{fig:gallery} & clouds background crop & \url{http://farm1.staticflickr.com/120/290916538_d4ad97b0b2_z.jpg} \\
\texttt{000000194067.jpg} & \cref{fig:overview}, new image set and \cref{fig:gallery} & sea background crop & \url{http://farm5.staticflickr.com/4094/4910723132_9c14eb5b47_z.jpg} \\
\texttt{000000288054.jpg} & \cref{fig:gallery} & bus foreground & \url{http://farm8.staticflickr.com/7287/8745059784_5b23e4d3c4_z.jpg} \\
\texttt{000000021475.jpg} & \cref{fig:gallery} & concrete wall background crop & \url{http://farm1.staticflickr.com/232/451749515_5098e87895_z.jpg} \\
\texttt{000000526542.jpg} & \cref{fig:gallery} & snow background crop & \url{http://farm8.staticflickr.com/7032/6772167653_fb40694025_z.jpg} \\
\texttt{000000240537.jpg} & \cref{fig:gallery} & dog foreground & \url{http://farm5.staticflickr.com/4141/4747121249_c0096e4cac_z.jpg} \\
\texttt{000000481723.jpg} & \cref{fig:gallery} & sky background crop & \url{http://farm8.staticflickr.com/7178/6957319661_095caf5ffa_z.jpg} \\
\texttt{000000539510.jpg} & \cref{fig:gallery} & grass background crop & \url{http://farm4.staticflickr.com/3730/9121078232_5c09e0dbc9_z.jpg} \\
\texttt{000000514454.jpg} & \cref{fig:gallery} & truck foreground & \url{http://farm4.staticflickr.com/3798/8916127439_7a07a2ef9f_z.jpg} \\
\texttt{000000460997.jpg} & \cref{fig:gallery} & sea background crop & \url{http://farm5.staticflickr.com/4033/4414594439_a4c0c376b1_z.jpg} \\
\texttt{000000556057.jpg} & \cref{fig:gallery} & sky background crop & \url{http://farm4.staticflickr.com/3475/3290757479_79c53ccbea_z.jpg} \\
\texttt{000000294314.jpg} & \cref{fig:gallery} & cat foreground & \url{http://farm6.staticflickr.com/5052/5520988333_c38f907c06_z.jpg} \\
\texttt{000000228334.jpg} & \cref{fig:gallery} & grass background crop & \url{http://farm1.staticflickr.com/31/64904208_46cba88e1e_z.jpg} \\
\texttt{000000518179.jpg} & \cref{fig:gallery} & concrete wall background crop & \url{http://farm9.staticflickr.com/8164/7654055710_6eab525cca_z.jpg} \\
\texttt{000000514299.jpg} & \cref{fig:gallery} & airplane foreground & \url{http://farm5.staticflickr.com/4133/4968671792_a6fc81c9b7_z.jpg} \\
\texttt{000000320014.jpg} & \cref{fig:gallery} & building background crop & \url{http://farm8.staticflickr.com/7263/7710672596_2e2260fb06_z.jpg} \\
\texttt{000000581887.jpg} & \cref{fig:gallery} & clouds background crop & \url{http://farm3.staticflickr.com/2338/2121474444_4345f35f5a_z.jpg} \\
\texttt{000000173693.jpg} & \cref{fig:gallery} & horse foreground & \url{http://farm3.staticflickr.com/2739/4086983693_5c544bd515_z.jpg} \\
\texttt{000000286645.jpg} & \cref{fig:gallery} & snow background crop & \url{http://farm3.staticflickr.com/2779/4380910985_2921985874_z.jpg} \\
\texttt{000000267840.jpg} & \cref{fig:gallery} & tree background crop & \url{http://farm1.staticflickr.com/167/378891837_49e51a31e3_z.jpg} \\
\texttt{000000032059.jpg} & \cref{fig:gallery} & train foreground & \url{http://farm4.staticflickr.com/3744/9326980433_acc313eef3_z.jpg} \\
\texttt{000000467502.jpg} & \cref{fig:gallery} & tree background crop & \url{http://farm1.staticflickr.com/23/33974966_b5d5078b9e_z.jpg} \\
\texttt{000000516178.jpg} & \cref{fig:gallery} & building background crop & \url{http://farm9.staticflickr.com/8089/8597876080_84e46a181c_z.jpg} \\
\bottomrule

\end{longtable}

\endgroup

\clearpage
\section{The post hoc analyses and the constructed cases}
\label{sec:supp-after}

We specified every analysis that \cref{tab:posthoc} lists after we had seen both image sets' results, and we fixed each computation before it ran. Every rank correlation is Spearman's, with tied values given their average rank.

\begin{table}[!t]
  \caption{\textbf{Each post hoc analysis is reported in one place.} Rows name every analysis specified after both image sets' results were seen, with the section, the table or the figure that reports it, and the mark $\ddagger$ names these analyses elsewhere.}
  \label{tab:posthoc}
  \centering
  \footnotesize
  \begin{tabular}[t]{@{}ll@{}}
\toprule
Post hoc analysis & Where we report it \\
\midrule
\multicolumn{2}{@{}l}{\emph{Single classifiers and labels}} \\
Count and range of the classifiers' drops & \cref{tab:checkpoints} \\
Largest drop of any random pairing & \cref{sec:supp-after,tab:uniform} \\
Rank correlation of drop with ImageNet accuracy & \cref{fig:ranking} \\
Each label's drop & \cref{tab:labels} \\
Change of each classifier's drop between the image sets & \cref{fig:sets} \\
Interval of the mean absolute change & \cref{sec:supp-after} \\
Drops of the excluded checkpoints & \cref{tab:refused} \\
\midrule
\multicolumn{2}{@{}l}{\emph{The choice of the pairing and the baseline}} \\
Simultaneous lower bounds over all pairings & \cref{tab:band} \\
Mean over all label combinations as the baseline & \cref{tab:uniform} \\
Groups of classifiers fixed before the new image set & \cref{tab:transfer} \\
Other splits of the classifiers & \cref{tab:splits} \\
Hardest random pairing & \cref{tab:checks,tab:pools} \\
\midrule
\multicolumn{2}{@{}l}{\emph{Labels and training data}} \\
Hardest pairing without two labels a side & \cref{tab:removal} \\
Groups by the training data the cards state & \cref{tab:groups} \\
\bottomrule
\end{tabular}\qquad
\begin{tabular}[t]{@{}ll@{}}
\toprule
Post hoc analysis & Where we report it \\
\midrule
\multicolumn{2}{@{}l}{\emph{Images and composites}} \\
New matches within label combinations & \cref{tab:checks} \\
Third image set & \cref{tab:pools} \\
Gaussian blur of the mask & \cref{tab:pools} \\
Erosion of the mask & \cref{tab:edge} \\
Image matching & \cref{tab:matching} \\
Answers that compositing alone changes & \cref{tab:transitions} \\
Object on grey accuracy weighted by foregrounds & \cref{sec:supp-after} \\
Complete rebuild of the first image set's layers & \cref{sec:supp-composites} \\
\midrule
\multicolumn{2}{@{}l}{\emph{Order of classifiers}} \\
Upper end of the decrease in correlation & \cref{tab:results} \\
That decrease's interval on the first image set & \cref{tab:results} \\
Easiest pairing's correlation against all label combinations & \cref{tab:results} \\
Changes of order on integer counts & \cref{tab:ranks} \\
Reliable leads that change order & \cref{tab:pairs} \\
\bottomrule
\end{tabular}

\end{table}

\paragraph{The drops of single classifiers.}
The drops vary widely from classifier to classifier. On the first image set the drops of the classifiers that pass the compositing check run from \DropMinFirst to \DropMaxFirst. No random pairing has a drop within \WithinPoint of the hardest pairing's there, and the largest drop of any random pairing is \RandomMaxFirst. On the new image set the classifiers' drops run from \DropMinNew to \DropMaxNew, and the largest drop of any random pairing is \RandomMaxNew. \Cref{tab:checkpoints} marks with \S{} the \BelowZeroFirst classifiers on each image set whose interval includes zero.

\paragraph{The change between the image sets.}
The two classifiers whose drops change most move in opposite directions: \ArchBj from \MovedAFirst to \MovedANew and \ArchBe from \MovedBFirst to \MovedBNew. The signed changes of each classifier's drop are \AbsChangeNew in size on average. Over source image resamples, their signed mean runs from $\ChangeSignedLo$ to \ChangeSignedHi. Each of the first image set's \BootReps replicates is paired with one of the new image set's by a seeded permutation, so each pair joins two independent draws. The resampled mean absolute change runs from \ChangeAbsLo to \ChangeAbsHi, above the observed \AbsChangeNew, because resampling adds noise to each classifier's change and an absolute value counts noise of either sign. For the \BothAdmitted classifiers that pass both checks, the first image set's drop of \CohortFirst less the new image set's \CohortNew is \PairedBoth, with source image interval $[\CohortDiffLo,\CohortDiffHi]$.

\paragraph{The order of classifiers.}
A change of rank concerns point estimates. The test for rank reversals asks instead whether a reliable lead becomes a reliable deficit. \Cref{tab:ranks} ranks the evaluation classifiers that pass each image set's check by their accuracy on all \Cells label combinations and on the hardest pairing, beside their accuracy on the easiest pairing. A pair changes its order where the classifier ahead on all \Cells label combinations drops behind on the hardest pairing at the point estimates. We count such changes on the integer counts of correct answers, so that a tie on the hardest pairing changes no order. Then \OrderChangesFirst pairs on the first image set and \OrderChangesNew on the new one change their order, \OrderChangesResolvedFirst and \OrderChangesResolvedNew of them pairs with a reliable lead, and the two orders correlate at \OrderRhoFirst and \OrderRhoNew.

\Cref{tab:pairs} gives each of those \OrderChangesResolvedBoth pairs with a reliable lead, with the lead on all \Cells label combinations and its lower bound and the lead on the hardest pairing and its interval. On the first image set \PointBeyondFirst of these pairs change their order at the point estimate by more than \ReversalMargin, though none has an upper bound below $-\ReversalMargin$. The largest change there, $\WorstFirst$ on a pair without a reliable lead, has an upper bound of $\WorstUpperFirst$. On the new image set \PointBeyondNew pairs with a reliable lead change by more than \ReversalMargin, and none has an upper bound below $-\ReversalMargin$.

\paragraph{The correlations.}
Each correlation pairs a classifier's probit ImageNet accuracy with its probit accuracy on one test set, over the evaluation classifiers that pass the compositing check. The two ends of a difference of correlations in \cref{tab:results} are the fifth and the ninety fifth percentiles of the coupled differences of \cref{sec:supp-composites}. On the first image set, where the correlation test is not run, the same \BootReps coupled draws give both ends.

\begin{figure}[!t]
  \centering
  \includegraphics{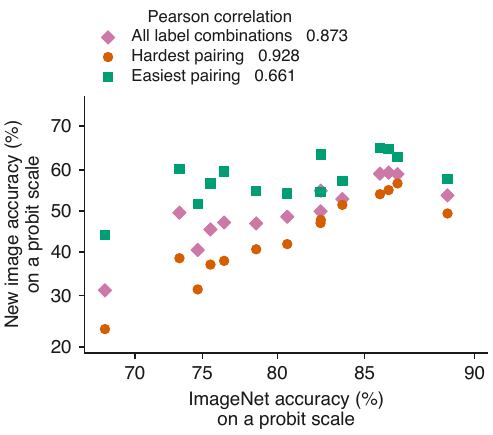}
  \caption{\textbf{The easiest pairing has a lower probit correlation than all label combinations, and the hardest pairing a higher one at its point estimate.}\aftertest Each point is one of the \AdmittedNew evaluation classifiers passing the new image set's check, plotted by its accuracy on a test set against its ImageNet accuracy, both in percent on probit scales. The legend gives Pearson's correlation for each test set; \cref{tab:results} gives intervals for the paired differences, and $\ddagger$ names a post hoc analysis.}
  \label{fig:line}
\end{figure}

There the decrease runs from $\LineDecreaseFirstLow$ to $\LineDecreaseFirstHigh$, wholly below zero, so the hardest pairing's correlation exceeds that over all \Cells label combinations on the first image set alone, and the new image set's higher correlation is a point estimate. The same draws give the difference between the correlation over all label combinations and the correlation on the easiest pairing, which is \LineEasyFirst on the first image set and \LineEasyNew on the new one. That difference is \EasyGapFirst on the first image set, with an interval from \EasyGapFirstLo to \EasyGapFirstHi, and \EasyGapNew on the new one, with an interval from \EasyGapNewLo to \EasyGapNewHi. Both intervals lie above zero, and \cref{fig:line} plots the new image set's three correlations.

\paragraph{The hardest random pairing.}
The hardest random pairing tells how much the full search adds over a cheap search among random pairings. It is the stored random pairing of lowest mean accuracy for the selection half on the first image set, index \ComparatorIndex counted from zero, which we fixed before we looked again at the evaluation scores. On the new image set it has a drop of \ComparatorDrop against the mean over all \Cells label combinations for all \HalfSize evaluation classifiers, with a lower bound of \ComparatorLower. Its difference from the hardest pairing pairs the two within each of the new image set's \BootReps source image resamples.

\paragraph{New matches within label combinations.}
New matches keep every pairing of labels and change only which foreground meets which background crop inside each label combination. They draw one new permutation for each label combination by seed \SeedRerun, in the order of seed \SeedCells, and we score the new image set's \ImagesScored images again with all \RerunModels evaluation classifiers on an \DeviceSxm in the software above. The new permutations change no image other than the composites, and on every one of those other images each classifier's prediction equals its earlier prediction. Against the random pairings the classifiers have a drop of \RerunSampledDrop under the new matches, with a lower bound of \RerunSampledLower. The resamples draw the new image set's source images by seed \SeedUnit, and the change from the original matches pairs the two scorings within each resample.

\paragraph{The third image set.}
The third image set uses the same way of building composites as the new one. All \PoolModels evaluation classifiers score its \ImagesScored images on an \DeviceSxm in the software above, and its own compositing check excludes none of them. Before we built the set's layers, we rebuilt the new image set's layers from its stored images in the recorded environment, and every level equalled the stored one.

\paragraph{The two label combinations removed.}
\Cref{tab:removal} removes airplane, truck, building and sea from the stored scores of the label combinations and keeps every prediction as a choice among all eight object labels. The hardest of the \RemovalPairings remaining pairings on the first image set has a summed background effect of $\RemovalSum$. It is car on clouds, bus on concrete wall, dog on sky, cat on grass, horse on snow and train on tree, the \RemovalShared label combinations that the hardest pairing already gives those labels. Its drop is taken against the mean of the \RemovalCells remaining label combinations, which equals the mean over all \RemovalPairings remaining pairings, for all \HalfSize evaluation classifiers. On the new and third image sets the drop is \RemovalDropNew and \RemovalDropThird, and its lower bounds of \RemovalLowerNew and \RemovalLowerThird lie above zero and below the margin of \Margin.

\begin{table}[!t]
  \caption{\textbf{No reliable lead meets the reversal margin, though \PointBeyondBoth pairs change order at their point estimates by more than \ReversalMargin and one on each image set has an interval wholly below zero.}\aftertest Rows give each pair with a reliable lead whose leader on all \Cells label combinations trails on the hardest pairing at the point estimate. A reversal needs the upper end of the hardest pairing's interval below $-\ReversalMargin$, and $\ddagger$ names a post hoc analysis.}
  \label{tab:pairs}
  \centering
  \small
  \renewcommand{\ckpt}[1]{\texttt{\let\_\checkpointbreak#1}}
  \begin{tabular}{@{}llrr@{}}
\toprule
\multicolumn{4}{@{}l}{A lead is the leader's accuracy less the other classifier's, in fraction correct.} \\
Leader on all combinations & Other classifier & All \Cells combinations & Hardest pairing \\
\midrule
\multicolumn{4}{@{}l}{\textit{First image set}} \\
\ArchAd & \ArchAz & \PairFirstAAll\iv{\PairFirstAAllLow} & $\PairFirstAHard$\iv{[$\PairFirstAHardLo$, \PairFirstAHardHi]} \\
\ArchAd & \ArchBg & \PairFirstBAll\iv{\PairFirstBAllLow} & $\PairFirstBHard$\iv{[$\PairFirstBHardLo$, $\PairFirstBHardHi$]} \\
\ArchAn & \ArchBg & \PairFirstCAll\iv{\PairFirstCAllLow} & $\PairFirstCHard$\iv{[$\PairFirstCHardLo$, \PairFirstCHardHi]} \\
\ArchAq & \ArchBg & \PairFirstDAll\iv{\PairFirstDAllLow} & $\PairFirstDHard$\iv{[$\PairFirstDHardLo$, \PairFirstDHardHi]} \\
\ArchAs & \ArchBg & \PairFirstEAll\iv{\PairFirstEAllLow} & $\PairFirstEHard$\iv{[$\PairFirstEHardLo$, \PairFirstEHardHi]} \\
\ArchAu & \ArchBe & \PairFirstFAll\iv{\PairFirstFAllLow} & $\PairFirstFHard$\iv{[$\PairFirstFHardLo$, \PairFirstFHardHi]} \\
\ArchAu & \ArchBj & \PairFirstGAll\iv{\PairFirstGAllLow} & $\PairFirstGHard$\iv{[$\PairFirstGHardLo$, \PairFirstGHardHi]} \\
\ArchBf & \ArchBj & \PairFirstHAll\iv{\PairFirstHAllLow} & $\PairFirstHHard$\iv{[$\PairFirstHHardLo$, \PairFirstHHardHi]} \\
\midrule
\multicolumn{4}{@{}l}{\textit{New image set}} \\
\ArchAd & \ArchAl & \PairNewAAll\iv{\PairNewAAllLow} & $\PairNewAHard$\iv{[$\PairNewAHardLo$, \PairNewAHardHi]} \\
\ArchBf & \ArchBe & \PairNewBAll\iv{\PairNewBAllLow} & $\PairNewBHard$\iv{[$\PairNewBHardLo$, $\PairNewBHardHi$]} \\
\bottomrule
\end{tabular}

\end{table}

\paragraph{Other splits of the classifiers.}
Other splits ask whether the result depends on which classifiers fell into which half. \Cref{tab:splits} draws \SplitCount splits of the \Models classifiers by our rule, one checkpoint of each of the \Strata strata of two to each half by a seeded coin, our own split first. Each split's selection half finds its hardest pairing on the first image set by the enumeration of \cref{eq:hardest}, and the compositing check includes or excludes each evaluation classifier separately on each image set by our rule. Depending on the split, \SplitAdmittedMinFirst to \SplitAdmittedMaxFirst evaluation classifiers pass the check on the first image set and \SplitAdmittedMinNew to \SplitAdmittedMaxNew on the new one. All \SplitDistinct splits differ, and on the first image set the splits' drops run from \SplitFirstMin to \SplitFirstMax. On the new image set the least lower bound is \SplitBoundMin, and no split has \ShuffleLimit random pairings within \WithinPoint of its drop.

\paragraph{Answers that compositing alone changes.}
Compositing alone changes answers in both directions, so a small net change can hide larger changes each way. It turns \TransRwLoFirst to \TransRwHiFirst right answers of a classifier that passes the check wrong and \TransWrLoFirst to \TransWrHiFirst wrong ones right on the first image set, and \TransRwLoNew to \TransRwHiNew and \TransWrLoNew to \TransWrHiNew on the new one. \Cref{tab:transitions} counts, for each classifier and each image set, the originals answered right whose check images are answered wrong, the reverse and the foregrounds whose answer changes at all. After scoring, weighting each classifier's stored scores on the objects on grey by the foreground frequencies of the first image set's \PanelsCompared test sets gave the same accuracy for every test set, as it must when every test set uses each foreground equally.

\paragraph{The constructed cases.}
Each constructed case of \cref{tab:known} gives our scoring an input whose result is known in advance, and in every case the scoring returned that result. A set of simulated classifiers has accuracies that we build into every label combination. One kind sets its hardest pairing's label combinations low and linear in a shuffle of the ImageNet probit coordinate, which breaks the correlation, and a second sets them low and linear in that coordinate, which keeps it. A third kind sets every label combination at one level; random pairings reproduce its drop, so the correlation test does not apply.

Effect error is the largest absolute difference across \Cells label combinations between the recovered mean effect and the built in probability difference. A drop exceeds the margin when its point estimate and its lower bound exceed \Margin. The correlation weakens where it is at least \LineGate over all label combinations and decreases by more than \LineMargin at both the point estimate and the lower bound. A reversal of a reliable lead uses the complete pairwise rule of \cref{sec:measures}, and the two classifiers that the simulations build in for exclusion each gain \KnownShamShift correct answers on check images against a limit of \ShamLimit.

For the simulated classifiers, let \(x\) be equally spaced over the probits of \KnownCoordinateLo to \KnownCoordinateHi and let \(z\) be a seeded permutation of \(x\). The simulation with six labels a side repeats its first three classifier coordinates after the \Models original coordinates. Its maximum label combinations are the identity assignment and its minimum ones the cyclic assignment shifted by one. Their probabilities are \(\Phi(\KnownHighIntercept+\KnownSlope x+u_{fb})\) and \(\Phi(\KnownLowIntercept+\KnownSlope z)\), where \(\Phi\) is the standard normal distribution function, with \(x\) replacing \(z\) for the case that keeps the correlation. Other label combinations have probability \(\Phi(\KnownMiddleIntercept+\KnownSlope x+u_{fb})\), where the fixed offsets \(u_{fb}\) are uniform on \([\KnownOffsetLo,\KnownOffsetHi]\).

The simulation with eight labels a side uses the same coordinate grid and middle label combinations, with minimum label combinations at background indices \(\KnownHardestCoordinates\) in object label order, indexed from zero. Its minimum uses the same low probability formula. The case with one level gives every label combination probability \(\Phi(\KnownMiddleIntercept+\KnownSlope x)\). The offsets with six labels use seed \KnownSeedSix and those with eight labels seed \KnownSeedEight, with each coordinate permutation drawn from the next integer seed. The separate effect recovery case builds in grey accuracy \KnownGrey, default effect $\KnownDefault$ with uniform jitter from $\KnownJitterLo$ to \KnownJitterHi, identity effect \KnownLiftHigh and minimum effect $\KnownLiftLow$ on the minimum label combinations of the eight label simulation.

\begin{table}[!t]
  \caption{\textbf{Our scoring returned the expected result in every constructed case.} Each case ran before its image set was scored, on scores or tables whose result was fixed by construction. Its row gives the expected and the returned result.}
  \label{tab:known}
  \centering
  \small
  \begin{tabular}{@{}lp{0.27\textwidth}p{0.29\textwidth}p{0.29\textwidth}@{}}
\toprule
\shortstack[l]{Image\\set} & Case & Expected result & Returned result \\
\midrule
First & a pairing built in at six labels a side & found by enumeration, the solver equal & found, the solver equal \\
First & an additive table at six labels a side & every pairing one sum & one sum \\
First & a constructed table of object and background cooccurrence shares at six labels a side & pairings of largest and smallest total cooccurrence & both found \\
First & a set of simulated classifiers built to break the correlation, at six labels a side & the drop exceeds the margin, the correlation weakens, the order reverses & the drop exceeds the margin, the correlation weakens, the order reverses \\
First & a set of simulated classifiers built to keep the correlation, at six labels a side & the drop exceeds the margin, the correlation stays, no order reverses & the drop exceeds the margin, the correlation stays, no order reverses \\
First & a set of simulated classifiers built with one level in every label combination, at six labels a side & the random pairings reproduce the drop, the correlation test does not apply, no order reverses & the random pairings reproduce the drop, the correlation test does not apply, no order reverses \\
First & a constructed table of background effects at eight labels a side & the built in hardest pairing, the identity as the easiest and its eight label combinations marked & all found, maximum absolute effect error \KnownLiftErr against tolerance \KnownLiftTolerance \\
First & a set of simulated classifiers built to break the correlation, at eight labels a side & the drop exceeds the margin, the correlation weakens, the order reverses, both built in classifiers excluded & the drop exceeds the margin, the correlation weakens, the order reverses, both built in classifiers excluded \\
First & a set of simulated classifiers built to keep the correlation, at eight labels a side & the drop exceeds the margin, the correlation stays, no order reverses, both built in classifiers excluded & the drop exceeds the margin, the correlation stays, no order reverses, both built in classifiers excluded \\
First & a set of simulated classifiers built with one level in every label combination, at eight labels a side & the random pairings reproduce the drop, the correlation test does not apply, no order reverses, both built in classifiers excluded & the random pairings reproduce the drop, the correlation test does not apply, no order reverses, both built in classifiers excluded \\
New & the first image set's stored scores & the first image set's drop and lower bound & \KnownReplayDrop and \KnownReplayLow, equal \\
New & constructed scores on the new image set's label combinations & a drop above the margin, one under it and none & \KnownPlantDrop above, \KnownPlantUnder under, and zero drop, reproduced by the random pairings \\
New & our image building for the first image set, on \KnownBuiltImages of its source images & the stored layers and images & equal byte for byte over \KnownBuiltComposites composites and \KnownBuiltControls other images \\
\bottomrule
\end{tabular}

\end{table}

\begin{table}[!t]
  \caption{\textbf{Pasting an object back over its own surroundings both corrects some answers and spoils others, so the net change hides part of each.}\aftertest Columns count, out of \Foregrounds foregrounds, the originals answered right whose check images are answered wrong, the reverse, their net and the foregrounds whose predicted label changes. Arrows give the better direction, bold the extremes within each half and $\ddagger$ a post hoc analysis.}
  \label{tab:transitions}
  \centering
  \small
  \setlength{\tabcolsep}{3pt}
  \renewcommand{\ckpt}[1]{\texttt{\let\_\checkpointbreak#1}}
  \setlength{\tabcolsep}{1pt}
  \begin{tabular}{@{}llrrrrrrrr@{}}
\toprule
\multicolumn{10}{@{}l}{Entries count source images by change in answer.} \\
& & \multicolumn{4}{c}{First image set} & \multicolumn{4}{c}{New image set} \\
\cmidrule(lr){3-6}\cmidrule(l){7-10}
Classifier & Half & \shortstack{Right to\\wrong $\downarrow$} & \shortstack{Wrong to\\right $\uparrow$} & Net & Changed $\downarrow$ & \shortstack{Right to\\wrong $\downarrow$} & \shortstack{Wrong to\\right $\uparrow$} & Net & Changed $\downarrow$ \\
\midrule
\ArchAa & selection & \BestABQ{\RwAaFirst} & \BestABR{\WrAaFirst} & $\ShamAaFirst$ & \BestABS{\ChgAaFirst} & \BestABT{\RwAaNew} & \BestABU{\WrAaNew} & $\ShamAaNew$ & \BestABV{\ChgAaNew} \\
\ArchAh & selection & \BestABW{\RwAhFirst} & \BestABX{\WrAhFirst} & \ShamAhFirst & \BestABY{\ChgAhFirst} & \BestABZ{\RwAhNew} & \BestACA{\WrAhNew} & \ShamAhNew & \BestACB{\ChgAhNew} \\
\ArchAc & selection & \BestACC{\RwAcFirst} & \BestACD{\WrAcFirst} & \ShamAcFirst & \BestACE{\ChgAcFirst} & \BestACF{\RwAcNew} & \BestACG{\WrAcNew} & \ShamAcNew & \BestACH{\ChgAcNew} \\
\ArchAg & selection & \BestACI{\RwAgFirst} & \BestACJ{\WrAgFirst} & $\ShamAgFirst$ & \BestACK{\ChgAgFirst} & \BestACL{\RwAgNew} & \BestACM{\WrAgNew} & \ShamAgNew & \BestACN{\ChgAgNew} \\
\ArchAf & selection & \BestACO{\RwAfFirst} & \BestACP{\WrAfFirst} & \ShamAfFirst & \BestACQ{\ChgAfFirst} & \BestACR{\RwAfNew} & \BestACS{\WrAfNew} & $\ShamAfNew$ & \BestACT{\ChgAfNew} \\
\ArchAp & selection & \BestACU{\RwApFirst} & \BestACV{\WrApFirst} & $\ShamApFirst$ & \BestACW{\ChgApFirst} & \BestACX{\RwApNew} & \BestACY{\WrApNew} & $\ShamApNew$ & \BestACZ{\ChgApNew} \\
\ArchAm & selection & \BestADA{\RwAmFirst} & \BestADB{\WrAmFirst} & \ShamAmFirst & \BestADC{\ChgAmFirst} & \BestADD{\RwAmNew} & \BestADE{\WrAmNew} & \ShamAmNew & \BestADF{\ChgAmNew} \\
\ArchAo & selection & \BestADG{\RwAoFirst} & \BestADH{\WrAoFirst} & \ShamAoFirst & \BestADI{\ChgAoFirst} & \BestADJ{\RwAoNew} & \BestADK{\WrAoNew} & \ShamAoNew & \BestADL{\ChgAoNew} \\
\ArchAr & selection & \BestADM{\RwArFirst} & \BestADN{\WrArFirst} & $\ShamArFirst$ & \BestADO{\ChgArFirst} & \BestADP{\RwArNew} & \BestADQ{\WrArNew} & $\ShamArNew$ & \BestADR{\ChgArNew} \\
\ArchAx & selection & \BestADS{\RwAxFirst} & \BestADT{\WrAxFirst} & \ShamAxFirst & \BestADU{\ChgAxFirst} & \BestADV{\RwAxNew} & \BestADW{\WrAxNew} & \ShamAxNew & \BestADX{\ChgAxNew} \\
\ArchAw & selection & \BestADY{\RwAwFirst} & \BestADZ{\WrAwFirst} & \ShamAwFirst & \BestAEA{\ChgAwFirst} & \BestAEB{\RwAwNew} & \BestAEC{\WrAwNew} & \ShamAwNew & \BestAED{\ChgAwNew} \\
\ArchAt & selection & \BestAEE{\RwAtFirst} & \BestAEF{\WrAtFirst} & \ShamAtFirst & \BestAEG{\ChgAtFirst} & \BestAEH{\RwAtNew} & \BestAEI{\WrAtNew} & \ShamAtNew & \BestAEJ{\ChgAtNew} \\
\ArchAy & selection & \BestAEK{\RwAyFirst} & \BestAEL{\WrAyFirst} & \ShamAyFirst & \BestAEM{\ChgAyFirst} & \BestAEN{\RwAyNew} & \BestAEO{\WrAyNew} & \ShamAyNew & \BestAEP{\ChgAyNew} \\
\ArchBb & selection & \BestAEQ{\RwBbFirst} & \BestAER{\WrBbFirst} & \ShamBbFirst & \BestAES{\ChgBbFirst} & \BestAET{\RwBbNew} & \BestAEU{\WrBbNew} & \ShamBbNew & \BestAEV{\ChgBbNew} \\
\ArchBa & selection & \BestAEW{\RwBaFirst} & \BestAEX{\WrBaFirst} & \ShamBaFirst & \BestAEY{\ChgBaFirst} & \BestAEZ{\RwBaNew} & \BestAFA{\WrBaNew} & \ShamBaNew & \BestAFB{\ChgBaNew} \\
\ArchBd & selection & \BestAFC{\RwBdFirst} & \BestAFD{\WrBdFirst} & \ShamBdFirst & \BestAFE{\ChgBdFirst} & \BestAFF{\RwBdNew} & \BestAFG{\WrBdNew} & \ShamBdNew & \BestAFH{\ChgBdNew} \\
\ArchBc & selection & \BestAFI{\RwBcFirst} & \BestAFJ{\WrBcFirst} & $\ShamBcFirst$ & \BestAFK{\ChgBcFirst} & \BestAFL{\RwBcNew} & \BestAFM{\WrBcNew} & \ShamBcNew & \BestAFN{\ChgBcNew} \\
\ArchBi & selection & \BestAFO{\RwBiFirst} & \BestAFP{\WrBiFirst} & $\ShamBiFirst$ & \BestAFQ{\ChgBiFirst} & \BestAFR{\RwBiNew} & \BestAFS{\WrBiNew} & \ShamBiNew & \BestAFT{\ChgBiNew} \\
\ArchBk & selection & \BestAFU{\RwBkFirst} & \BestAFV{\WrBkFirst} & \ShamBkFirst & \BestAFW{\ChgBkFirst} & \BestAFX{\RwBkNew} & \BestAFY{\WrBkNew} & \ShamBkNew & \BestAFZ{\ChgBkNew} \\
\ArchBl & selection & \BestAGA{\RwBlFirst} & \BestAGB{\WrBlFirst} & \ShamBlFirst & \BestAGC{\ChgBlFirst} & \BestAGD{\RwBlNew} & \BestAGE{\WrBlNew} & \ShamBlNew & \BestAGF{\ChgBlNew} \\
\ArchAb & evaluation & \BestAGG{\RwAbFirst} & \BestAGH{\WrAbFirst} & $\ShamAbFirst$ & \BestAGI{\ChgAbFirst} & \BestAGJ{\RwAbNew} & \BestAGK{\WrAbNew} & $\ShamAbNew$ & \BestAGL{\ChgAbNew} \\
\ArchAd & evaluation & \BestAGM{\RwAdFirst} & \BestAGN{\WrAdFirst} & $\ShamAdFirst$ & \BestAGO{\ChgAdFirst} & \BestAGP{\RwAdNew} & \BestAGQ{\WrAdNew} & \ShamAdNew & \BestAGR{\ChgAdNew} \\
\ArchAj & evaluation & \BestAGS{\RwAjFirst} & \BestAGT{\WrAjFirst} & \ShamAjFirst & \BestAGU{\ChgAjFirst} & \BestAGV{\RwAjNew} & \BestAGW{\WrAjNew} & $\ShamAjNew$ & \BestAGX{\ChgAjNew} \\
\ArchAe & evaluation & \BestAGY{\RwAeFirst} & \BestAGZ{\WrAeFirst} & \ShamAeFirst & \BestAHA{\ChgAeFirst} & \BestAHB{\RwAeNew} & \BestAHC{\WrAeNew} & $\ShamAeNew$ & \BestAHD{\ChgAeNew} \\
\ArchAi & evaluation & \BestAHE{\RwAiFirst} & \BestAHF{\WrAiFirst} & \ShamAiFirst & \BestAHG{\ChgAiFirst} & \BestAHH{\RwAiNew} & \BestAHI{\WrAiNew} & $\ShamAiNew$ & \BestAHJ{\ChgAiNew} \\
\ArchAk & evaluation & \BestAHK{\RwAkFirst} & \BestAHL{\WrAkFirst} & \ShamAkFirst & \BestAHM{\ChgAkFirst} & \BestAHN{\RwAkNew} & \BestAHO{\WrAkNew} & \ShamAkNew & \BestAHP{\ChgAkNew} \\
\ArchAn & evaluation & \BestAHQ{\RwAnFirst} & \BestAHR{\WrAnFirst} & \ShamAnFirst & \BestAHS{\ChgAnFirst} & \BestAHT{\RwAnNew} & \BestAHU{\WrAnNew} & $\ShamAnNew$ & \BestAHV{\ChgAnNew} \\
\ArchAq & evaluation & \BestAHW{\RwAqFirst} & \BestAHX{\WrAqFirst} & \ShamAqFirst & \BestAHY{\ChgAqFirst} & \BestAHZ{\RwAqNew} & \BestAIA{\WrAqNew} & \ShamAqNew & \BestAIB{\ChgAqNew} \\
\ArchAl & evaluation & \BestAIC{\RwAlFirst} & \BestAID{\WrAlFirst} & \ShamAlFirst & \BestAIE{\ChgAlFirst} & \BestAIF{\RwAlNew} & \BestAIG{\WrAlNew} & $\ShamAlNew$ & \BestAIH{\ChgAlNew} \\
\ArchAu & evaluation & \BestAII{\RwAuFirst} & \BestAIJ{\WrAuFirst} & \ShamAuFirst & \BestAIK{\ChgAuFirst} & \BestAIL{\RwAuNew} & \BestAIM{\WrAuNew} & \ShamAuNew & \BestAIN{\ChgAuNew} \\
\ArchAz & evaluation & \BestAIO{\RwAzFirst} & \BestAIP{\WrAzFirst} & \ShamAzFirst & \BestAIQ{\ChgAzFirst} & \BestAIR{\RwAzNew} & \BestAIS{\WrAzNew} & \ShamAzNew & \BestAIT{\ChgAzNew} \\
\ArchAv & evaluation & \BestAIU{\RwAvFirst} & \BestAIV{\WrAvFirst} & \ShamAvFirst & \BestAIW{\ChgAvFirst} & \BestAIX{\RwAvNew} & \BestAIY{\WrAvNew} & \ShamAvNew & \BestAIZ{\ChgAvNew} \\
\ArchAs & evaluation & \BestAJA{\RwAsFirst} & \BestAJB{\WrAsFirst} & $\ShamAsFirst$ & \BestAJC{\ChgAsFirst} & \BestAJD{\RwAsNew} & \BestAJE{\WrAsNew} & \ShamAsNew & \BestAJF{\ChgAsNew} \\
\ArchBg & evaluation & \BestAJG{\RwBgFirst} & \BestAJH{\WrBgFirst} & $\ShamBgFirst$ & \BestAJI{\ChgBgFirst} & \BestAJJ{\RwBgNew} & \BestAJK{\WrBgNew} & \ShamBgNew & \BestAJL{\ChgBgNew} \\
\ArchBf & evaluation & \BestAJM{\RwBfFirst} & \BestAJN{\WrBfFirst} & \ShamBfFirst & \BestAJO{\ChgBfFirst} & \BestAJP{\RwBfNew} & \BestAJQ{\WrBfNew} & $\ShamBfNew$ & \BestAJR{\ChgBfNew} \\
\ArchBe & evaluation & \BestAJS{\RwBeFirst} & \BestAJT{\WrBeFirst} & $\ShamBeFirst$ & \BestAJU{\ChgBeFirst} & \BestAJV{\RwBeNew} & \BestAJW{\WrBeNew} & \ShamBeNew & \BestAJX{\ChgBeNew} \\
\ArchBm & evaluation & \BestAJY{\RwBmFirst} & \BestAJZ{\WrBmFirst} & \ShamBmFirst & \BestAKA{\ChgBmFirst} & \BestAKB{\RwBmNew} & \BestAKC{\WrBmNew} & \ShamBmNew & \BestAKD{\ChgBmNew} \\
\ArchBh & evaluation & \BestAKE{\RwBhFirst} & \BestAKF{\WrBhFirst} & $\ShamBhFirst$ & \BestAKG{\ChgBhFirst} & \BestAKH{\RwBhNew} & \BestAKI{\WrBhNew} & \ShamBhNew & \BestAKJ{\ChgBhNew} \\
\ArchBn & evaluation & \BestAKK{\RwBnFirst} & \BestAKL{\WrBnFirst} & \ShamBnFirst & \BestAKM{\ChgBnFirst} & \BestAKN{\RwBnNew} & \BestAKO{\WrBnNew} & \ShamBnNew & \BestAKP{\ChgBnNew} \\
\ArchBj & evaluation & \BestAKQ{\RwBjFirst} & \BestAKR{\WrBjFirst} & \ShamBjFirst & \BestAKS{\ChgBjFirst} & \BestAKT{\RwBjNew} & \BestAKU{\WrBjNew} & $\ShamBjNew$ & \BestAKV{\ChgBjNew} \\
\bottomrule
\end{tabular}

\end{table}

\begin{table}[!t]
  \captionsetup{skip=18pt}
  \caption{\textbf{On the hardest pairing some classifiers change rank, while the ranks on the new image set still correlate at \OrderCorrNew by Spearman's coefficient.}\aftertest The upper block gives fraction correct for the \AdmittedFirst classifiers passing the first image set's check, and the lower block fraction correct and rank for the \AdmittedNew passing the new image set's check. A change of rank is a change of point estimates, which the test for rank reversals of \cref{tab:pairs} does not count by itself. Arrows give the better direction, bold the extremes and $\ddagger$ a post hoc analysis.}
  \label{tab:ranks}
  \centering
  \small
  \renewcommand{\ckpt}[1]{\texttt{\let\_\checkpointbreak#1}}
  \begin{tabular}{@{}lrrr@{}}
\toprule
First image set classifier & All combinations $\uparrow$ & Hardest $\uparrow$ & Easiest $\uparrow$ \\
\midrule
\ArchAb & \BestGP{\FirstEasyAaGrid} & \BestGQ{\FirstEasyAaHard} & \BestGR{\FirstEasyAaEasy} \\
\ArchAd & \BestGS{\FirstEasyAbGrid} & \BestGT{\FirstEasyAbHard} & \BestGU{\FirstEasyAbEasy} \\
\ArchAe & \BestGV{\FirstEasyAcGrid} & \BestGW{\FirstEasyAcHard} & \BestGX{\FirstEasyAcEasy} \\
\ArchAi & \BestGY{\FirstEasyAdGrid} & \BestGZ{\FirstEasyAdHard} & \BestHA{\FirstEasyAdEasy} \\
\ArchAj & \BestHB{\FirstEasyAeGrid} & \BestHC{\FirstEasyAeHard} & \BestHD{\FirstEasyAeEasy} \\
\ArchAk & \BestHE{\FirstEasyAfGrid} & \BestHF{\FirstEasyAfHard} & \BestHG{\FirstEasyAfEasy} \\
\ArchAl & \BestHH{\FirstEasyAgGrid} & \BestHI{\FirstEasyAgHard} & \BestHJ{\FirstEasyAgEasy} \\
\ArchAn & \BestHK{\FirstEasyAhGrid} & \BestHL{\FirstEasyAhHard} & \BestHM{\FirstEasyAhEasy} \\
\ArchAq & \BestHN{\FirstEasyAiGrid} & \BestHO{\FirstEasyAiHard} & \BestHP{\FirstEasyAiEasy} \\
\ArchAs & \BestHQ{\FirstEasyAjGrid} & \BestHR{\FirstEasyAjHard} & \BestHS{\FirstEasyAjEasy} \\
\ArchAu & \BestHT{\FirstEasyAkGrid} & \BestHU{\FirstEasyAkHard} & \BestHV{\FirstEasyAkEasy} \\
\ArchAz & \BestHW{\FirstEasyAlGrid} & \BestHX{\FirstEasyAlHard} & \BestHY{\FirstEasyAlEasy} \\
\ArchBe & \BestHZ{\FirstEasyAmGrid} & \BestIA{\FirstEasyAmHard} & \BestIB{\FirstEasyAmEasy} \\
\ArchBf & \BestIC{\FirstEasyAnGrid} & \BestID{\FirstEasyAnHard} & \BestIE{\FirstEasyAnEasy} \\
\ArchBg & \BestIF{\FirstEasyAoGrid} & \BestIG{\FirstEasyAoHard} & \BestIH{\FirstEasyAoEasy} \\
\ArchBh & \BestII{\FirstEasyApGrid} & \BestIJ{\FirstEasyApHard} & \BestIK{\FirstEasyApEasy} \\
\ArchBj & \BestIL{\FirstEasyAqGrid} & \BestIM{\FirstEasyAqHard} & \BestIN{\FirstEasyAqEasy} \\
\ArchBn & \BestIO{\FirstEasyArGrid} & \BestIP{\FirstEasyArHard} & \BestIQ{\FirstEasyArEasy} \\
\bottomrule
\end{tabular}
\par\medskip
  \begin{tabular}{@{}lrrrrrr@{}}
\toprule
& & \multicolumn{2}{c}{All \BestOZ{\Cells} combinations} & \multicolumn{2}{c}{Hardest pairing} & Easiest \\
\cmidrule(lr){3-4}\cmidrule(lr){5-6}\cmidrule(l){7-7}
Classifier & \shortstack{ImageNet\\percent $\uparrow$} & \shortstack{Fraction\\correct $\uparrow$} & Rank $\downarrow$ & \shortstack{Fraction\\correct $\uparrow$} & Rank $\downarrow$ & \shortstack{Fraction\\correct $\uparrow$} \\
\midrule
\ArchAb & \BestPA{\TopAb} & \BestPB{\GridAbNew} & \BestPC{\RankGridAb} & \BestPD{\HardAbNew} & \BestPE{\RankHardAb} & \EasyAbNew \\
\ArchAd & \BestPF{\TopAd} & \BestPG{\GridAdNew} & \BestPH{\RankGridAd} & \BestPI{\HardAdNew} & \BestPJ{\RankHardAd} & \EasyAdNew \\
\ArchAe & \BestPK{\TopAe} & \BestPL{\GridAeNew} & \BestPM{\RankGridAe} & \BestPN{\HardAeNew} & \BestPO{\RankHardAe} & \EasyAeNew \\
\ArchAi & \BestPP{\TopAi} & \BestPQ{\GridAiNew} & \BestPR{\RankGridAi} & \BestPS{\HardAiNew} & \BestPT{\RankHardAi} & \EasyAiNew \\
\ArchAk & \BestPU{\TopAk} & \BestPV{\GridAkNew} & \BestPW{\RankGridAk} & \BestPX{\HardAkNew} & \BestPY{\RankHardAk} & \EasyAkNew \\
\ArchAl & \BestPZ{\TopAl} & \BestQA{\GridAlNew} & \BestQB{\RankGridAl} & \BestQC{\HardAlNew} & \BestQD{\RankHardAl} & \EasyAlNew \\
\ArchAz & \BestQE{\TopAz} & \BestQF{\GridAzNew} & \BestQG{\RankGridAz} & \BestQH{\HardAzNew} & \BestQI{\RankHardAz} & \EasyAzNew \\
\ArchBg & \BestQJ{\TopBg} & \BestQK{\GridBgNew} & \BestQL{\RankGridBg} & \BestQM{\HardBgNew} & \BestQN{\RankHardBg} & \EasyBgNew \\
\ArchBf & \BestQO{\TopBf} & \BestQP{\GridBfNew} & \BestQQ{\RankGridBf} & \BestQR{\HardBfNew} & \BestQS{\RankHardBf} & \EasyBfNew \\
\ArchBe & \BestQT{\TopBe} & \BestQU{\GridBeNew} & \BestQV{\RankGridBe} & \BestQW{\HardBeNew} & \BestQX{\RankHardBe} & \EasyBeNew \\
\ArchBm & \BestQY{\TopBm} & \BestQZ{\GridBmNew} & \BestRA{\RankGridBm} & \BestRB{\HardBmNew} & \BestRC{\RankHardBm} & \textbf{\EasyBmNew} \\
\ArchBh & \BestRD{\TopBh} & \BestRE{\GridBhNew} & \BestRF{\RankGridBh} & \BestRG{\HardBhNew} & \BestRH{\RankHardBh} & \EasyBhNew \\
\ArchBn & \BestRI{\TopBn} & \BestRJ{\GridBnNew} & \BestRK{\RankGridBn} & \BestRL{\HardBnNew} & \BestRM{\RankHardBn} & \EasyBnNew \\
\ArchBj & \BestRN{\TopBj} & \BestRO{\GridBjNew} & \BestRP{\RankGridBj} & \BestRQ{\HardBjNew} & \BestRR{\RankHardBj} & \EasyBjNew \\
\bottomrule
\end{tabular}

\end{table}

\begin{table}[!t]
  \caption{\textbf{Every one of \SplitCount splits of the classifiers gives a drop on the new image set whose lower bound exceeds \Margin.}\aftertest The upper table gives quantiles over splits of the drop against \RandomPairings random pairings and of its lower bound, which on the first image set holds each split's pairing fixed. The lower table counts the splits whose hardest pairing contains each label combination, ours in tint. Arrows give the better direction, bold the largest value of each quantity and $\ddagger$ a post hoc analysis.}
  \label{tab:splits}
  \centering
  \small
  \begin{tabular}{@{}llrrrrrrr@{}}
\toprule
\multicolumn{9}{@{}l}{Drop quantiles are differences in fraction correct.} \\
Image set & Quantity & Least $\uparrow$ & Fifth $\uparrow$ & Quarter $\uparrow$ & Median $\uparrow$ & \shortstack{Three\\quarters $\uparrow$} & \shortstack{Ninety\\fifth $\uparrow$} & Greatest $\uparrow$ \\
\midrule
First & Drop & \BestXO{\SplitFirstDropA} & \BestXP{\SplitFirstDropB} & \BestXQ{\SplitFirstDropC} & \BestXR{\SplitFirstDropD} & \BestXS{\SplitFirstDropE} & \BestXT{\SplitFirstDropF} & \BestXU{\SplitFirstDropG} \\
First & Lower bound & \BestXV{\SplitFirstBoundA} & \BestXW{\SplitFirstBoundB} & \BestXX{\SplitFirstBoundC} & \BestXY{\SplitFirstBoundD} & \BestXZ{\SplitFirstBoundE} & \BestYA{\SplitFirstBoundF} & \BestYB{\SplitFirstBoundG} \\
New & Drop & \BestYC{\SplitNewDropA} & \BestYD{\SplitNewDropB} & \BestYE{\SplitNewDropC} & \BestYF{\SplitNewDropD} & \BestYG{\SplitNewDropE} & \BestYH{\SplitNewDropF} & \BestYI{\SplitNewDropG} \\
New & Lower bound & \BestYJ{\SplitNewBoundA} & \BestYK{\SplitNewBoundB} & \BestYL{\SplitNewBoundC} & \BestYM{\SplitNewBoundD} & \BestYN{\SplitNewBoundE} & \BestYO{\SplitNewBoundF} & \BestYP{\SplitNewBoundG} \\
\bottomrule
\end{tabular}
\medskip

\begin{tabular}{@{}lrrrrrrrr@{}}
\toprule
\multicolumn{9}{@{}l}{Entries count splits that select each label combination.} \\
& sky $\uparrow$ & tree $\uparrow$ & \shortstack{concrete\\wall} $\uparrow$ & grass $\uparrow$ & snow $\uparrow$ & sea $\uparrow$ & building $\uparrow$ & clouds $\uparrow$ \\
\midrule
car & \BestYQ{\SplitCellAa} & \BestYR{\SplitCellAb} & \BestYS{\SplitCellAc} & \BestYT{\SplitCellAd} & \BestYU{\SplitCellAe} & \BestYV{\SplitCellAf} & \BestYW{\SplitCellAg} & \cellcolor{MethodTint}\BestYX{\SplitCellAh} \\
bus & \BestYY{\SplitCellBa} & \BestYZ{\SplitCellBb} & \cellcolor{MethodTint}\BestZA{\SplitCellBc} & \BestZB{\SplitCellBd} & \BestZC{\SplitCellBe} & \BestZD{\SplitCellBf} & \BestZE{\SplitCellBg} & \BestZF{\SplitCellBh} \\
dog & \cellcolor{MethodTint}\BestZG{\SplitCellCa} & \BestZH{\SplitCellCb} & \BestZI{\SplitCellCc} & \BestZJ{\SplitCellCd} & \BestZK{\SplitCellCe} & \BestZL{\SplitCellCf} & \BestZM{\SplitCellCg} & \BestZN{\SplitCellCh} \\
truck & \BestZO{\SplitCellDa} & \BestZP{\SplitCellDb} & \BestZQ{\SplitCellDc} & \BestZR{\SplitCellDd} & \BestZS{\SplitCellDe} & \cellcolor{MethodTint}\BestZT{\SplitCellDf} & \BestZU{\SplitCellDg} & \BestZV{\SplitCellDh} \\
cat & \BestZW{\SplitCellEa} & \BestZX{\SplitCellEb} & \BestZY{\SplitCellEc} & \cellcolor{MethodTint}\BestZZ{\SplitCellEd} & \BestAAA{\SplitCellEe} & \BestAAB{\SplitCellEf} & \BestAAC{\SplitCellEg} & \BestAAD{\SplitCellEh} \\
airplane & \BestAAE{\SplitCellFa} & \BestAAF{\SplitCellFb} & \BestAAG{\SplitCellFc} & \BestAAH{\SplitCellFd} & \BestAAI{\SplitCellFe} & \BestAAJ{\SplitCellFf} & \cellcolor{MethodTint}\BestAAK{\SplitCellFg} & \BestAAL{\SplitCellFh} \\
horse & \BestAAM{\SplitCellGa} & \BestAAN{\SplitCellGb} & \BestAAO{\SplitCellGc} & \BestAAP{\SplitCellGd} & \cellcolor{MethodTint}\BestAAQ{\SplitCellGe} & \BestAAR{\SplitCellGf} & \BestAAS{\SplitCellGg} & \BestAAT{\SplitCellGh} \\
train & \BestAAU{\SplitCellHa} & \cellcolor{MethodTint}\BestAAV{\SplitCellHb} & \BestAAW{\SplitCellHc} & \BestAAX{\SplitCellHd} & \BestAAY{\SplitCellHe} & \BestAAZ{\SplitCellHf} & \BestABA{\SplitCellHg} & \BestABB{\SplitCellHh} \\
\bottomrule
\end{tabular}

\end{table}

\clearpage
\section{The mask edits and the image matching}
\label{sec:supp-edge}

Two edits of the mask leave the drop in place, and the image matching transferred by position, transfers less than the pairing of labels. Intuitively, the erosion pulls the edge of the object inward by \EdgeRadius pixel. Precisely, it takes the minimum alpha in a square neighbourhood of radius \EdgeRadius pixel at the final resolution, with zero outside the canvas. Foreground colours and positions stay fixed, as do the background crops, and the operation applies no blur. Pixels whose alpha is unchanged retain their composite colour exactly.

The same operation builds the objects on grey and the check images. The \EdgeModels classifiers that pass the original new image set's check stay the group, and the hardest pairing and the \RandomPairings random pairings remain fixed. \Cref{tab:edge} reports the eroded check images beside that group.

Before any eroded scoring, we checked that the erosion and our scoring change nothing else. An identity operation reproduced the original composites byte for byte, and a constructed square mask moved inward by the radius fixed in advance. Our scoring also reproduced the stored predictions and probabilities on \UnitsPerLabel original composites chosen in advance with a largest absolute error of \ReplayError, under absolute tolerance \ReplayAbsolute plus relative tolerance \ReplayRelative times the stored probability. The chosen composites are those numbered $\ReplayStride i$ for $i=0,\ldots,\ReplayLast$, counting from zero in the order object label, background label, foreground. The last batch of that comparison was padded to \ScoreBatch images to keep the original batch shape. The source image bootstrap couples both kinds of composite within each replicate and takes their paired difference, with the classifiers fixed.

Intuitively, the blur softens the mask inside the boundary and spreads low alpha a few pixels outside it, where the foreground layer contains the object's own surroundings. Precisely, the blurred composites of \cref{sec:new} filter each foreground's mask alpha, at the final resolution, by a Gaussian of standard deviation \BlurSigma pixels in each direction. The kernel is cut at \BlurTruncate standard deviations with zero taken outside the canvas. Each filtered level is then set to the nearest whole level, a half going up.

Foreground colours and positions stay fixed, as do the background crops and the permutations within label combinations. The same blurred alpha builds the objects on grey and the check images. All \BlurModels evaluation classifiers score the blurred composites, and on the \BlurControlImages originals, which pass through no mask, every classifier's prediction equals its earlier prediction. Its change from the original composites pairs the two within each resample of the new image set's source images.

Intuitively, the image matching is the hardest pairing's counterpart at the level of single images. Precisely, it chooses among the \CompositesAll existing composites of the first image set one composite for each foreground and each background crop, so that the selection half's sum of correct predictions is smallest. An exact assignment solver uses each of the \Foregrounds foregrounds and \Windows background crops once. We fixed these matches before we looked again at the evaluation scores or the new image set's scores, and the same matches, by position, select the composites of the new image set.

Of its \PanelSize matched composites, \MatchInside lie in the hardest pairing's label combinations. \Cref{tab:matching} compares both assignments against the same random pairings on the \EdgeModels classifiers that pass the original new image set's check. Each source image resample uses identical weights for both assignments and the random baseline, normalizes within each object label and averages the labels equally.

\begin{table}[!t]
  \caption{\textbf{The drop persists after a one pixel erosion of the mask alpha.}\aftertest The upper table compares the original and eroded mean drops against \RandomPairings random pairings for the \AdmittedNew classifiers passing the original check, and the lower table gives each classifier's drops beside its counts on the eroded check images, out of \Foregrounds foregrounds. Subscripts give intervals, arrows the better direction, bold the extremes within each block and $\ddagger$ a post hoc analysis.}
  \label{tab:edge}
  \centering
  \small
  \setlength{\tabcolsep}{3pt}
  \renewcommand{\ckpt}[1]{\texttt{\let\_\checkpointbreak#1}}
  \begin{tabular}{@{}lc@{}}
\toprule
\multicolumn{2}{@{}l}{Drops and their differences are changes in fraction correct.} \\
Composites & Mean drop $\uparrow$ with interval \\
\midrule
Original & $\BestIR{\EdgeOld}$\iv{$\EdgeOldLo$ to $\EdgeOldHi$} \\
Eroded alpha & $\BestIS{\EdgeTreated}$\iv{$\EdgeTreatedLo$ to $\EdgeTreatedHi$} \\
Eroded less original & $\EdgeDifference$\iv{$\EdgeDifferenceLo$ to $\EdgeDifferenceHi$} \\
\bottomrule
\end{tabular}
\medskip
\begin{tabular}{@{}lrrrrrrr@{}}
\toprule
\multicolumn{8}{@{}l}{Drops are fractions of answers, and the check image columns count foregrounds.} \\
& \multicolumn{3}{c}{Drop in accuracy} & \multicolumn{4}{c}{Eroded check images} \\
\cmidrule(lr){2-4}\cmidrule(l){5-8}
Classifier & Original $\downarrow$ & Eroded $\downarrow$ & Difference & \shortstack{Right to\\wrong $\downarrow$} & \shortstack{Wrong to\\right $\uparrow$} & Net & Changed $\downarrow$ \\
\midrule
\ArchAb & $\BestIT{\EdgeOldAb}$ & $\BestIU{\EdgeNewAb}$ & $\EdgeDeltaAb$ & $\BestIV{\EdgeRwAb}$ & $\BestIW{\EdgeWrAb}$ & $\EdgeNetAb$ & $\BestIX{\EdgeChangedAb}$ \\
\ArchAd & $\BestIY{\EdgeOldAd}$ & $\BestIZ{\EdgeNewAd}$ & $\EdgeDeltaAd$ & $\BestJA{\EdgeRwAd}$ & $\BestJB{\EdgeWrAd}$ & $\EdgeNetAd$ & $\BestJC{\EdgeChangedAd}$ \\
\ArchAe & $\BestJD{\EdgeOldAe}$ & $\BestJE{\EdgeNewAe}$ & $\EdgeDeltaAe$ & $\BestJF{\EdgeRwAe}$ & $\BestJG{\EdgeWrAe}$ & $\EdgeNetAe$ & $\BestJH{\EdgeChangedAe}$ \\
\ArchAi & $\BestJI{\EdgeOldAi}$ & $\BestJJ{\EdgeNewAi}$ & $\EdgeDeltaAi$ & $\BestJK{\EdgeRwAi}$ & $\BestJL{\EdgeWrAi}$ & $\EdgeNetAi$ & $\BestJM{\EdgeChangedAi}$ \\
\ArchAk & $\BestJN{\EdgeOldAk}$ & $\BestJO{\EdgeNewAk}$ & $\EdgeDeltaAk$ & $\BestJP{\EdgeRwAk}$ & $\BestJQ{\EdgeWrAk}$ & $\EdgeNetAk$ & $\BestJR{\EdgeChangedAk}$ \\
\ArchAl & $\BestJS{\EdgeOldAl}$ & $\BestJT{\EdgeNewAl}$ & $\EdgeDeltaAl$ & $\BestJU{\EdgeRwAl}$ & $\BestJV{\EdgeWrAl}$ & $\EdgeNetAl$ & $\BestJW{\EdgeChangedAl}$ \\
\ArchAz & $\BestJX{\EdgeOldAz}$ & $\BestJY{\EdgeNewAz}$ & $\EdgeDeltaAz$ & $\BestJZ{\EdgeRwAz}$ & $\BestKA{\EdgeWrAz}$ & $\EdgeNetAz$ & $\BestKB{\EdgeChangedAz}$ \\
\ArchBe & $\BestKC{\EdgeOldBe}$ & $\BestKD{\EdgeNewBe}$ & $\EdgeDeltaBe$ & $\BestKE{\EdgeRwBe}$ & $\BestKF{\EdgeWrBe}$ & $\EdgeNetBe$ & $\BestKG{\EdgeChangedBe}$ \\
\ArchBf & $\BestKH{\EdgeOldBf}$ & $\BestKI{\EdgeNewBf}$ & $\EdgeDeltaBf$ & $\BestKJ{\EdgeRwBf}$ & $\BestKK{\EdgeWrBf}$ & $\EdgeNetBf$ & $\BestKL{\EdgeChangedBf}$ \\
\ArchBg & $\BestKM{\EdgeOldBg}$ & $\BestKN{\EdgeNewBg}$ & $\EdgeDeltaBg$ & $\BestKO{\EdgeRwBg}$ & $\BestKP{\EdgeWrBg}$ & $\EdgeNetBg$ & $\BestKQ{\EdgeChangedBg}$ \\
\ArchBh & $\BestKR{\EdgeOldBh}$ & $\BestKS{\EdgeNewBh}$ & $\EdgeDeltaBh$ & $\BestKT{\EdgeRwBh}$ & $\BestKU{\EdgeWrBh}$ & $\EdgeNetBh$ & $\BestKV{\EdgeChangedBh}$ \\
\ArchBj & $\BestKW{\EdgeOldBj}$ & $\BestKX{\EdgeNewBj}$ & $\EdgeDeltaBj$ & $\BestKY{\EdgeRwBj}$ & $\BestKZ{\EdgeWrBj}$ & $\EdgeNetBj$ & $\BestLA{\EdgeChangedBj}$ \\
\ArchBm & $\BestLB{\EdgeOldBm}$ & $\BestLC{\EdgeNewBm}$ & $\EdgeDeltaBm$ & $\BestLD{\EdgeRwBm}$ & $\BestLE{\EdgeWrBm}$ & $\EdgeNetBm$ & $\BestLF{\EdgeChangedBm}$ \\
\ArchBn & $\BestLG{\EdgeOldBn}$ & $\BestLH{\EdgeNewBn}$ & $\EdgeDeltaBn$ & $\BestLI{\EdgeRwBn}$ & $\BestLJ{\EdgeWrBn}$ & $\EdgeNetBn$ & $\BestLK{\EdgeChangedBn}$ \\
\bottomrule
\end{tabular}

\end{table}

\begin{table}[!t]
  \caption{\textbf{The pairing of labels keeps a larger drop on new images than the image matching does.}\aftertest Both assignments use the same foregrounds and background crops, and both drops are taken against \RandomPairings random pairings. The upper block uses the \EdgeModels classifiers that pass the new image set's compositing check and the lower block the \BothAdmitted that pass both checks. Subscripts give intervals, $\uparrow$ the better direction, bold the larger drop, tint the pairing of labels and $\ddagger$ a post hoc analysis.}
  \label{tab:matching}
  \centering
  \small
  \begin{tabular}{@{}lcc@{}}
\toprule
\multicolumn{3}{@{}l}{Entries give accuracy drops as fractions of answers.} \\
Assignment & First image set $\uparrow$ & New image set $\uparrow$ \\
\midrule
\rowcolor{MethodTint}Pairing of labels & $\BestOV{\MatchLabelFirst}$\iv{$\MatchLabelFirstLo$ to $\MatchLabelFirstHi$} & $\BestOW{\MatchLabelNew}$\iv{$\MatchLabelNewLo$ to $\MatchLabelNewHi$} \\
Image matching & $\BestOX{\MatchImageFirst}$\iv{$\MatchImageFirstLo$ to $\MatchImageFirstHi$} & $\BestOY{\MatchImageNew}$\iv{$\MatchImageNewLo$ to $\MatchImageNewHi$} \\
Pairing less matching & $\MatchDeltaFirst$\iv{$\MatchDeltaFirstLo$ to $\MatchDeltaFirstHi$} & $\MatchDeltaNew$\iv{$\MatchDeltaNewLo$ to $\MatchDeltaNewHi$} \\
\bottomrule
\end{tabular}
\par\medskip
\begin{tabular}{@{}lccc@{}}
\toprule
Classifiers & First image set & New image set & First less new \\
\midrule
That pass both checks & \CohortFirst & \CohortNew & $\PairedBoth$\iv{$\CohortDiffLo$ to $\CohortDiffHi$} \\
\bottomrule
\end{tabular}

\end{table}

\section{Selection and the comparison against all label combinations}
\label{sec:supp-selection}

Allowing for the choice of the pairing lowers the first image set's bound below the margin, and groups of classifiers fixed in advance exceed it on the new image set. On the new image set the pairing was already fixed, so its bounds need no such allowance.

The simultaneous band accounts for selecting a pairing from the same images of the first image set. Intuitively, it asks how far the luckiest of all pairings could exceed its true drop by chance, and it lowers every bound by that amount. For each candidate pairing \(\pi\), let \(\widehat d(\pi)\) be the mean accuracy of the \AdmittedFirst classifiers that pass the first image set's check over the stored \RandomPairings random pairings less their accuracy on \(\pi\). Each of the original \BootReps source image resamples gives \(\widehat d^{*}_{j}(\pi)\) under the same weights for every pairing. The band uses the ninety fifth percentile \(c\) of \(\max_{\pi}\{\widehat d(\pi)-\widehat d^{*}_{j}(\pi)\}\) and lower bound \(\widehat d(\pi)-c\) over all \CandidatePairings pairings. The computed constant is \BandConstant, as \cref{tab:band} reports, with the classifiers fixed; using this sign for a lower band assumes joint symmetry of the errors across pairings.

We checked by simulation, with a set of simulated classifiers, how often the selected pairing's bound covers its true drop when the errors of the two halves are linked through shared effects of the source images. Coverage is the share of simulations in which a lower bound lies below the true drop. We ran this simulation before reading the stored scores for this post hoc analysis. In each of \CoverageWorlds simulations, \HalfSize selection classifiers chose a pairing and \AdmittedFirst evaluation classifiers measured it over \Cells label combinations, with \Foregrounds foregrounds and \Windows background crops. For zero based labels \(f,b\), the accuracy that the simulation builds into each label combination is
\begin{equation}
 \theta_{fb}=\SyntheticCenter+\SyntheticSlope(f-\SyntheticMidpoint)
 +\SyntheticSlope(b-\SyntheticMidpoint)-\SyntheticPenalty\,\mathbf{1}_{f=b}.
\end{equation}
Here \(\mathbf{1}_{f=b}\) equals one where \(f=b\) and zero otherwise, so the identity pairing uniquely minimizes the sum, and each classifier adds an offset equally spaced from \(-\SyntheticOffset\) to \(\SyntheticOffset\) within its half.

Intuitively, the simulation links the errors of the two halves, both through shared photographs and through shared outcomes. Each simulation redraws independent foreground and background crop effects taking values minus one and one with equal probability, each multiplied by \SyntheticSource. The foreground loading is positive when \(f+b\) is even and negative otherwise. The background crop loading is positive when \(f+2b\) is divisible by three and negative otherwise. One independent uniform variable per composite is shared by every classifier, and a classifier is correct when that variable is below its probability.

The seeds below let a reader repeat the simulation. Seed \CoverageLayoutSeed fixes the permutations within label combinations. Seed \CoverageSeed spawns one seed per simulation, each spawning separate source, outcome, random pairing and bootstrap streams in that order. Every simulation draws \RandomPairings random pairings by the shuffle of background labels of \cref{sec:supp-composites} and uses \CoverageReps source image resamples for the band. Coverage compares the selected pairing's lower bound with its exact drop under \(\theta\) against those same random pairings. The criterion accepts coverage of at least \BandLevel, or below it by at most the half width of its Wilson binomial interval at level \BandLevel.

The selected pairing's bound covers in all \CoverageBandCount simulations, with interval \([\CoverageBandLo,\CoverageBandHi]\), and the bound with the pairing fixed covers in \CoverageOrdinaryCount, a share of \CoverageOrdinary with interval \([\CoverageOrdinaryLo,\CoverageOrdinaryHi]\). The band therefore meets the coverage criterion in this simulation.

Three constructed cases check the computation of the band against results known in advance. With the same replicate shortfall for every candidate pairing, the computation returns constant \FourthKnownCommonConstant and a bound with the pairing fixed and a simultaneous bound of \FourthKnownCommonOrdinary and \FourthKnownCommonSimultaneous, matching the expected result. When the outcomes of the label combinations are constant, it returns constant \FourthKnownCellConstant and both bounds \FourthKnownCellOrdinary, equal to the expected point drop within numerical precision. When outcomes depend only on foregrounds and the weights of the background crops are balanced, it returns maximum shortfall \FourthKnownFgConstant and bound difference \FourthKnownFgBoundError, matching zero within numerical precision. In all three, the computation returned the expected result.

\begin{table}[!t]
 \caption{\textbf{A lower bound that allows for the search over pairings on the first image set falls below the margin.}\aftertest The upper table gives each pairing's drop against random pairings, its lower bound with the pairing held fixed and its simultaneous lower bound. The lower table gives the coverage of each bound over \CoverageWorlds simulations, with Wilson intervals at level \BandLevel as subscripts. Arrows give the better direction, bold the column maxima, tint the hardest pairing and $\ddagger$ a post hoc analysis.}
 \label{tab:band}
 \centering
 \small
 \begin{tabular}{@{}lrrr@{}}
\toprule
\multicolumn{4}{@{}l}{The drop and its bounds are differences in fraction correct.} \\
Pairing & Drop $\uparrow$ & Lower, pairing fixed $\uparrow$ & Simultaneous lower $\uparrow$ \\
\midrule
\rowcolor{MethodTint}Hardest & \BestA{\BandPointHardest} & \BestB{\BandOrdinaryHardest} & \BestC{\BandSimultaneousHardest} \\
Easiest & $\BestD{\BandPointEasiest}$ & $\BestE{\BandOrdinaryEasiest}$ & $\BestF{\BandSimultaneousEasiest}$ \\
Random pairing with largest drop & \BestG{\BandPointRandom} & \BestH{\BandOrdinaryRandom} & $\BestI{\BandSimultaneousRandom}$ \\
\bottomrule
\end{tabular}
\par\medskip
\begin{tabular}{@{}lrr@{}}
\toprule
Bound for simulated classifiers & Simulations covered $\uparrow$ & Share $\uparrow$ \\
\midrule
Simultaneous & \BestJ{\CoverageBandCount}/\CoverageWorlds & \BestK{\CoverageBand}\iv{$[\CoverageBandLo,\CoverageBandHi]$} \\
Pairing fixed & \BestL{\CoverageOrdinaryCount}/\CoverageWorlds & \BestM{\CoverageOrdinary}\iv{$[\CoverageOrdinaryLo,\CoverageOrdinaryHi]$} \\
\bottomrule
\end{tabular}

\end{table}

A second baseline, the mean over all label combinations, needs no random draw. The mean accuracy over all \Cells label combinations in \cref{tab:uniform} equals the mean over all pairings, since every label combination appears in the same fraction of all pairings. Its drop is that mean less accuracy on the hardest pairing, computed with identical source image weights for both terms. On each image set the random pairing with the largest drop is the stored random pairing with the largest observed drop, with ties going to the earliest stored index, counted from zero, and that pairing stays fixed within resamples.

The paired difference is the hardest pairing's drop less that random pairing's drop in each resample. These intervals condition on a pairing chosen from that image set's outcomes. Applying the new image set's choice to the first image set gives difference \CarriedDifference with interval $[\CarriedDifferenceLo,\CarriedDifferenceHi]$.

When one image set determines both which classifiers pass and the drop, fixing those classifiers in every resample leaves that uncertainty outside the interval. The further comparisons on the new image set in \cref{tab:transfer} use groups determined without that set's checks, so each replicate averages the same checkpoints regardless of the new check counts. One group is every checkpoint assigned to evaluation, and the other the checkpoints that pass the first image set's compositing check. Both keep the pairing fixed by the first image set and measure variation across source images for their checkpoints, and group averages give every checkpoint equal weight. The \AdmittedNew classifiers that pass the new image set's check condition on that check, and the first image set's simultaneous bound covers the choice of pairing on its fixed group. The excluded checkpoints' drops on the new image set in \cref{tab:refused} keep the same paired resampling and both baselines.

\begin{table}[!t]
  \caption{\textbf{On the new image set every group's lower bound exceeds the margin of \Margin, while the first image set's simultaneous bound lies below it.}\aftertest Rows give groups of evaluation classifiers, and columns give the drop against random pairings, the drop against all \Cells label combinations, and the hardest pairing's drop less that of the random pairing with the largest drop, chosen on the outcomes it is compared with. Subscripts give lower bounds or intervals, arrows the better direction, bold the column maxima within each image set, tint the hardest pairing and $\ddagger$ a post hoc analysis.}
  \label{tab:transfer}
  \centering
  \small
  \setlength{\tabcolsep}{2.5pt}
  \begin{tabular}{@{}lccc@{}}
\toprule
\multicolumn{4}{@{}l}{Entries give differences in fraction correct.} \\
& \multicolumn{2}{c}{Drop against} & \\
\cmidrule(lr){2-3}
Classifiers & \shortstack{Random\\pairings $\uparrow$} & \shortstack{All \Cells label\\combinations $\uparrow$} & \shortstack{Hardest less largest\\random drop $\uparrow$} \\
\midrule
\multicolumn{4}{@{}l}{\emph{First image set}} \\
\rowcolor{MethodTint}\FourthFirstModels that pass the check & \BestABC{\CompactFirstSampledDrop}\iv{\CompactFirstSampledLower} & \BestABD{\CompactFirstUniformDrop}\iv{\CompactFirstUniformLower} & \BestABE{\CompactFirstDelta}\iv{$[\CompactFirstDeltaLo,\CompactFirstDeltaHi]$} \\
\rowcolor{MethodTint}\FourthFirstModels with the search counted & \BestABF{\CompactFirstSampledDrop}\iv{\CompactFirstSampledSimultaneous} & \BestABG{\CompactFirstUniformDrop}\iv{\CompactFirstUniformSimultaneous} & \\
\midrule
\multicolumn{4}{@{}l}{\emph{New image set}} \\
\rowcolor{MethodTint}\FourthConditionalModels that pass the check & \BestABH{\CompactConditionalSampledDrop}\iv{\CompactConditionalSampledLower} & \BestABI{\CompactConditionalUniformDrop}\iv{\CompactConditionalUniformLower} & \BestABJ{\CompactConditionalDelta}\iv{$[\CompactConditionalDeltaLo,\CompactConditionalDeltaHi]$} \\
\rowcolor{MethodTint}All \FourthAllModels & \BestABK{\CompactAllSampledDrop}\iv{\CompactAllSampledLower} & \BestABL{\CompactAllUniformDrop}\iv{\CompactAllUniformLower} & \BestABM{\CompactAllDelta}\iv{$[\CompactAllDeltaLo,\CompactAllDeltaHi]$} \\
\rowcolor{MethodTint}\FourthFixedModels that pass the first image set's check & \BestABN{\CompactFixedSampledDrop}\iv{\CompactFixedSampledLower} & \BestABO{\CompactFixedUniformDrop}\iv{\CompactFixedUniformLower} & \BestABP{\CompactFixedDelta}\iv{$[\CompactFixedDeltaLo,\CompactFixedDeltaHi]$} \\
\bottomrule
\end{tabular}

\end{table}

\begin{table}[!t]
 \caption{\textbf{The groups fixed in advance exceed the margin on the new image set, while the first image set's simultaneous bounds lie below it.}\aftertest Each block names its group of classifiers, and columns compare drops against \RandomPairings random pairings and against all \Cells label combinations. The random pairing with the largest drop is chosen on the outcomes it is compared with. Subscripts give intervals, arrows the better direction, bold the greater drop, tint the hardest pairing and $\ddagger$ a post hoc analysis.}
 \label{tab:uniform}
 \centering
 \small
 \setlength{\tabcolsep}{3pt}
 \begin{tabular}{@{}lrr@{}}
\toprule
\multicolumn{3}{@{}l}{Accuracy and accuracy drops are fractions of answers.} \\
Quantity & Random pairings $\uparrow$ & All \Cells label combinations $\uparrow$ \\
\midrule
\multicolumn{3}{l}{First image set, \FourthFirstModels classifiers that pass the first image set's check} \\
Baseline accuracy & \FourthFirstSampledAccuracy & \FourthFirstUniformAccuracy \\
\rowcolor{MethodTint}Hardest pairing drop & \BestAKW{\FourthFirstSampledDrop} & \BestAKX{\FourthFirstUniformDrop} \\
Random pairing with largest drop, index \FourthFirstIndex & \BestAKY{\FourthFirstSampledStrong} & \BestAKZ{\FourthFirstUniformStrong} \\
Lower bound, pairing fixed & \FourthFirstSampledLower & \FourthFirstUniformLower \\
Simultaneous lower bound & \FourthFirstSampledSimultaneous & \FourthFirstUniformSimultaneous \\
Hardest less largest random & \FourthFirstDelta\iv{$[\FourthFirstDeltaLo,\FourthFirstDeltaHi]$} & \FourthFirstDelta\iv{$[\FourthFirstDeltaLo,\FourthFirstDeltaHi]$} \\
\midrule
\multicolumn{3}{l}{New image set, \FourthConditionalModels classifiers that pass the new image set's check} \\
Baseline accuracy & \FourthConditionalSampledAccuracy & \FourthConditionalUniformAccuracy \\
\rowcolor{MethodTint}Hardest pairing drop & \BestALA{\FourthConditionalSampledDrop} & \BestALB{\FourthConditionalUniformDrop} \\
Random pairing with largest drop, index \FourthConditionalIndex & \BestALC{\FourthConditionalSampledStrong} & \BestALD{\FourthConditionalUniformStrong} \\
Lower bound, pairing fixed & \FourthConditionalSampledLower & \FourthConditionalUniformLower \\
Hardest less largest random & \FourthConditionalDelta\iv{$[\FourthConditionalDeltaLo,\FourthConditionalDeltaHi]$} & \FourthConditionalDelta\iv{$[\FourthConditionalDeltaLo,\FourthConditionalDeltaHi]$} \\
\midrule
\multicolumn{3}{l}{New image set, all \FourthAllModels evaluation classifiers} \\
Baseline accuracy & \FourthAllSampledAccuracy & \FourthAllUniformAccuracy \\
\rowcolor{MethodTint}Hardest pairing drop & \BestALE{\FourthAllSampledDrop} & \BestALF{\FourthAllUniformDrop} \\
Random pairing with largest drop, index \FourthAllIndex & \BestALG{\FourthAllSampledStrong} & \BestALH{\FourthAllUniformStrong} \\
Lower bound, pairing fixed & \FourthAllSampledLower & \FourthAllUniformLower \\
Hardest less largest random & \FourthAllDelta\iv{$[\FourthAllDeltaLo,\FourthAllDeltaHi]$} & \FourthAllDelta\iv{$[\FourthAllDeltaLo,\FourthAllDeltaHi]$} \\
\midrule
\multicolumn{3}{l}{New image set, \FourthFixedModels classifiers that pass the first image set's check} \\
Baseline accuracy & \FourthFixedSampledAccuracy & \FourthFixedUniformAccuracy \\
\rowcolor{MethodTint}Hardest pairing drop & \BestALI{\FourthFixedSampledDrop} & \BestALJ{\FourthFixedUniformDrop} \\
Random pairing with largest drop, index \FourthFixedIndex & \BestALK{\FourthFixedSampledStrong} & \BestALL{\FourthFixedUniformStrong} \\
Lower bound, pairing fixed & \FourthFixedSampledLower & \FourthFixedUniformLower \\
Hardest less largest random & \FourthFixedDelta\iv{$[\FourthFixedDeltaLo,\FourthFixedDeltaHi]$} & \FourthFixedDelta\iv{$[\FourthFixedDeltaLo,\FourthFixedDeltaHi]$} \\
\bottomrule
\end{tabular}

\end{table}

\begin{table}[!t]
 \caption{\textbf{The classifiers that the new image set's compositing check excludes also have positive drops at their point estimates.}\aftertest Rows give their drops on the new image set against \RandomPairings random pairings and against all \Cells label combinations. Subscripts give intervals, $\downarrow$ the better direction, bold the column minima and $\ddagger$ a post hoc analysis.}
 \label{tab:refused}
 \centering
 \small
 \setlength{\tabcolsep}{4pt}
 \begin{tabular}{@{}lrr@{}}
\toprule
\multicolumn{3}{@{}l}{Both columns give accuracy drops as fractions of answers.} \\
Classifier excluded on the new image set & Random pairings $\downarrow$ & All \Cells label combinations $\downarrow$ \\
\midrule
\ArchAj & \BestRS{\NewRefusedAaSampledDrop}\iv{[\NewRefusedAaSampledLo,\NewRefusedAaSampledHi]} & \BestRT{\NewRefusedAaUniformDrop}\iv{[\NewRefusedAaUniformLo,\NewRefusedAaUniformHi]} \\
\ArchAn & \BestRU{\NewRefusedAbSampledDrop}\iv{[\NewRefusedAbSampledLo,\NewRefusedAbSampledHi]} & \BestRV{\NewRefusedAbUniformDrop}\iv{[\NewRefusedAbUniformLo,\NewRefusedAbUniformHi]} \\
\ArchAq & \BestRW{\NewRefusedAcSampledDrop}\iv{[\NewRefusedAcSampledLo,\NewRefusedAcSampledHi]} & \BestRX{\NewRefusedAcUniformDrop}\iv{[\NewRefusedAcUniformLo,\NewRefusedAcUniformHi]} \\
\ArchAs & \BestRY{\NewRefusedAdSampledDrop}\iv{[\NewRefusedAdSampledLo,\NewRefusedAdSampledHi]} & \BestRZ{\NewRefusedAdUniformDrop}\iv{[\NewRefusedAdUniformLo,\NewRefusedAdUniformHi]} \\
\ArchAu & \BestSA{\NewRefusedAeSampledDrop}\iv{[\NewRefusedAeSampledLo,\NewRefusedAeSampledHi]} & \BestSB{\NewRefusedAeUniformDrop}\iv{[\NewRefusedAeUniformLo,\NewRefusedAeUniformHi]} \\
\ArchAv & \BestSC{\NewRefusedAfSampledDrop}\iv{[\NewRefusedAfSampledLo,\NewRefusedAfSampledHi]} & \BestSD{\NewRefusedAfUniformDrop}\iv{[\NewRefusedAfUniformLo,\NewRefusedAfUniformHi]} \\
\bottomrule
\end{tabular}

\end{table}

\end{document}